\documentclass[10pt,twocolumn,letterpaper]{article}

\usepackage{cvpr}              

\usepackage{graphicx}
\usepackage{array}
\usepackage{multirow}
\usepackage[margin=1in]{geometry}
\usepackage{xcolor}
\usepackage{caption}
\usepackage{longtable}
\usepackage{algorithm}
\usepackage{algpseudocode}
\usepackage{float}

\usepackage[breakable]{tcolorbox}
\usepackage{listings}
\lstdefinelanguage{markdown}{
  basicstyle=\ttfamily\scriptsize,
  breaklines=true,
  breakatwhitespace=true,
  columns=fullflexible,
}
\definecolor{cvprblue}{rgb}{0.21,0.49,0.74}
\usepackage[pagebackref,breaklinks,colorlinks,allcolors=cvprblue]{hyperref}
\definecolor{tabblue}{RGB}{31,119,180}
\definecolor{taborange}{RGB}{255, 127, 14}
\definecolor{tabgreen}{RGB}{44, 160, 44}
\definecolor{tabred}{RGB}{214, 39, 40}

\def\paperID{***} 
\def\confName{CVPR}
\def\confYear{2026}

\title{Understanding Reward Hacking in Text-to-Image Reinforcement Learning}

\author{
Yunqi Hong$^{1}$,\quad Kuei-Chun Kao$^{1}$,\quad Hengguang Zhou$^{1}$,\quad Cho-Jui Hsieh$^{1,2}$\\
$^{1}$University of California, Los Angeles\\
$^{2}$Arena \\
{\tt\small \{yunqihong, chohsieh\}@cs.ucla.edu}
}

\begin{document}
\maketitle
\begin{abstract}
Reinforcement learning (RL) has become a standard approach for post-training large language models and, more recently, for improving image generation models, by optimizing generation quality and human preference alignment with reward signals. However, existing rewards are often imperfect proxies for true human judgment, making models prone to reward hacking and producing unrealistic or low-quality images that nevertheless achieve high reward scores.
In this work, we systematically analyze reward hacking behaviors in text-to-image (T2I) RL. We investigate how aesthetic/human preference rewards and prompt-image consistency rewards individually contribute to reward hacking. We further show that multiple reward ensemble can only partially mitigate this issue.
Across diverse reward models, we identify a common failure mode: the generation of artifact-prone images.
To address this, we propose ArtifactReward, a lightweight adaptive reward model trained on a small curated dataset to penalize generated artifacts.
Experiments demonstrate that incorporating our ArtifactReward into existing RL pipelines significantly improves visual realism and reduces reward hacking across multiple T2I RL setups, validating the effectiveness of our lightweight complementary reward serving as a safeguard against reward hacking.

Our code is available at: \url{https://github.com/yq-hong/ArtifactReward}.
\end{abstract}    
\section{Introduction}
\label{sec:intro}

The use of reinforcement learning (RL) has proven highly effective for post-training large language models (LLMs), enabling stronger alignment with human preferences through methods such as reinforcement learning from human feedback (RLHF)~\cite{ouyang2022training}, or more recently improving prediction accuracy via reinforcement learning with verifiable feedback (RLVF)~\cite{shao2024deepseekmath, guo2025deepseek, he2025skywork}. 
This success has motivated a growing interest in exploring RL-based fine-tuning for image generation tasks \cite{jiang2025t2i, duan2025got, yuan2025ar, chen2025iris, wei2025skywork, xue2025dancegrpo}.

To guide the optimization of generative models, various types of rewards have been proposed. Architecturally, many of them are built on vision-language encoders such as CLIP or BLIP~\cite{wu2023human, kirstain2023pick, xu2023imagereward}; some others leverage multimodal LLMs, either through direct scoring~\cite{ku2023viescore, wang2025unified, yuan2025ar}, additional prediction heads~\cite{zhou2025multimodal}, or token-level probability signals~\cite{lin2024evaluating, wu2023q, you2025teaching, guo2025can, wu2025rewarddance, li2024removing}); another category relies on external perception tools such as object detectors~\cite{liu2024grounding, cheng2024yolo, zhou2022simple}.
Functionally, reward models are typically designed to measure aesthetic or perceptual quality \cite{schuhmann2022laion, you2025teaching, yang2022maniqa}, human preference alignment \cite{wu2023human, kirstain2023pick, xu2023imagereward, wang2025unified}, and prompt-image consistency \cite{guo2025can, li2024removing, liu2024grounding, lin2024evaluating}.

However, unlike textual domains where verifiable signals (e.g., correctness or factuality) can be obtained, rewards for visual generation are often non-verifiable imperfect proxies of human preferences \cite{zhang2024large, liu2025flow}. Such imperfect reward design can inadvertently encourage models to exploit weaknesses in the reward signal to increase rewards without genuinely improving visual generation quality, which is a phenomenon known as reward hacking \cite{amodei2016concrete, gao2023scaling, skalse2022defining, clark2023directly}. As a result, fine-tuned models may produce visually implausible or low-quality images that nonetheless receive high reward scores.

In this work, we conduct the first systematic analysis of reward hacking behaviors in text-to-image (T2I) RL. We investigate how different types of reward functions individually contribute to reward hacking, including those designed for aesthetics or human preference and those enforcing prompt-image consistency. Although ensembling multiple rewards partially mitigate this issue, it cannot fully prevent reward exploitation. Across all settings, we identify a universal pattern of reward hacking: the generation of unrealistic visual artifacts that nevertheless receive high reward scores. While prior studies have occasionally reported instances of reward exploitation or proposed regularization techniques~\cite{clark2023directly, zhang2024large, shekhar2025rocm, liu2025flow}, ours is the first to systematically characterize how different reward types induce specific hacking behaviors and to identify their common pattern of artifact generation.

To address this issue, we propose ArtifactReward, a lightweight adaptive complementary reward that can be easily integrated into existing T2I RL pipelines. Our approach requires only a small dataset \textbf{(approximately 200 samples)} of model-generated images consisting of both artifact-free and artifact-containing examples. Using an LLM-based backbone, we automatically optimize a textual prompt that discriminates between these two categories, so as to assign higher scores to artifact-free images while lower scores to those containing artifacts.

Extensive experiments demonstrate that using our ArtifactReward as a complementary reward consistently improves visual realism and reduces reward hacking across multiple T2I RL setups.
Evaluations on two benchmarks, WISE~\cite{niu2025wise} and LLM4LLM~\cite{wang2025lmm4lmm}, show our method reliably boosts realism, consistency, and semantic alignment across all training configurations.

\section{Related Work}
\label{sec:related}

\subsection{Evaluation Metrics for T2I Generation}
\label{sec:evaluation_metric}
Unlike domains where ground-truth answers can serve as objective reward signals, such as solving math problems or programming, evaluating the quality of generated images is inherently more complex and multi-dimensional. It depends on various factors such as aesthetic appeal, semantic alignment with the prompt, and overall visual realism. This leads to a diverse landscape of reward design and evaluation metrics.

A major class of reward models is built upon vision-language encoders such as CLIP~\cite{radford2021learning} and BLIP~\cite{li2022blip}. HPSv2~\cite{wu2023human} and PickScore~\cite{kirstain2023pick} fine-tune CLIP on human-preference datasets to predict visual quality based on human aesthetic appeal and preference.
Aesthetic Score~\cite{schuhmann2022laion} predicts image aesthetics using a feedforward neural network over CLIP features. ImageReward~\cite{xu2023imagereward} is built on BLIP to estimate preference scores via cross-attention between image and text features.

Many works have also explored using MLLMs as image evaluators.
Davidsonian Scene Graph~\cite{cho2023davidsonian} evaluates image-prompt consistency by answering questions related to the prompt based on the generated image. VIEScore~\cite{ku2023viescore} utilizes the general knowledge of MLLMs to evaluate conditional image generation without additional finetuning. EvalAlign~\cite{tan2024evalalign} finetunes MLLMs to evaluate image faithfulness and text-image alignment.
UnifiedReward~\cite{wang2025unified} finetunes MLLMs to directly output scalar image scores through comprehensive analyses. LLaVA-Reward~\cite{zhou2025multimodal} employs the Bradley-Terry objective to train an MLLM evaluator using pairwise preference data.
Other works, such as Q-Align~\cite{wu2023q}, VQAScore~\cite{lin2024evaluating}, DeQA~\cite{you2025teaching}, finetuned ORM~\cite{guo2025can}, and RewardDance~\cite{wu2025rewarddance}, estimate image quality based on the token probabilities of specific answer tokens.

\subsection{Existing Rewards for T2I Post-Training}

\paragraph{Gradient-based finetuning methods.}
ImageReward~\cite{xu2023imagereward} is proposed with Reward Feedback Learning (ReFL), a direct optimization approach using gradients from reward models, to finetune diffusion models.
DRaFT~\cite{clark2023directly} further analyzes how rewards such as HPSv2~\cite{wu2023human} and PickScore~\cite{kirstain2023pick} influence model behavior when applied through differentiable optimization.
ROCM~\cite{shekhar2025rocm} compared training outcomes using different reward functions, including PickScore~\cite{kirstain2023pick}, HPSv2~\cite{wu2023human}, CLIPScore~\cite{hessel2021clipscore}, and Aesthetic Score~\cite{schuhmann2022laion}, to study their individual effects.

\vspace{-8pt}
\paragraph{RL-based optimization.}

RL Diffusion~\cite{zhang2024large} adopts REINFORCE-style policy gradients to train diffusion models with various reward functions, including ImageReward~\cite{xu2023imagereward} for human preference, UniDet~\cite{zhou2022simple} for compositionality, and a distribution-based statistical parity reward for diversity and fairness.
DDPO~\cite{black2023training} employs Aesthetic Score~\cite{schuhmann2022laion} and BERTScore~\cite{zhang2019bertscore} (measuring the semantic similarity between the prompt and VLM-generated image captions) as rewards.
DanceGRPO~\cite{xue2025dancegrpo} finetunes diffusion models with HPS-v2.1~\cite{wu2023human} and CLIP~\cite{radford2021learning}. Flow-GRPO~\cite{liu2025flow} improves the flow-matching models by using a combination of compositionality (GenEval~\cite{ghosh2023geneval}) and human preference (PickScore~\cite{kirstain2023pick}) rewards.
T2I-R1~\cite{jiang2025t2i} ensembles multiple rewards, including HPS~\cite{wu2023human}, Grounding DINO~\cite{liu2024grounding}, a VQA model~\cite{wang2022git}, and the finetuned ORM~\cite{guo2025can}.
AR-GRPO~\cite{yuan2025ar} incorporates HPS-v2.1~\cite{wu2023human} and CLIP~\cite{radford2021learning} for conditional reward, MANIQA~\cite{yang2022maniqa} for image quality, and prompting Qwen2.5-VL-3B-Instruct~\cite{bai2025qwen2} for realism.

Overall, existing T2I reward functions can be broadly categorized into three groups:
\begin{itemize}
    \item Human preference rewards such as HPS~\cite{wu2023human}, PickScore~\cite{kirstain2023pick}, and ImageReward~\cite{xu2023imagereward};
    \item Compositionality rewards such as Grounding DINO~\cite{liu2024grounding};
    \item MLLM-based rewards derived from visual question answering or image scoring models~\cite{lin2024evaluating, guo2025can, you2025teaching, wang2025unified}.
\end{itemize}
Our work focuses on analyzing model behaviors when trained with rewards from these categories.

\subsection{Reward Hacking in T2I RL}
Although none of the existing work provides a detailed discussion and systematic analysis of reward hacking in T2I RL, several studies have briefly discussed instances of reward hacking in their training experiments.  
DRaFT~\cite{clark2023directly} reports that finetuned models sometimes lose diversity and produce reward-specific visual patterns.
RL Diffusion~\cite{zhang2024large} hypothesizes that gradient-based optimization methods (i.e. ReFL~\cite{xu2023imagereward} and DRaFT~\cite{clark2023directly}) are more susceptible to reward hacking because they directly access reward gradients, and thus proposes an RL-based alternative to improve robustness.
ROCM~\cite{shekhar2025rocm} introduces distributional regularization to stabilize training and mitigate reward exploitation.
Flow-GRPO~\cite{liu2025flow} uses KL regularization for similar purposes.

\section{Reward hacking analysis}
\label{sec:hacking}
In this section, we characterize the reward hacking behaviors observed in T2I RL. We first describe the definition of reward hacking and discuss its relevance to T2I generation (Section~\ref{sec:hacking_definition}). Then we describe our experimental setup and rewards examined (Section ~\ref{sec:reward}). Finally, we analyze the training behaviors under different rewards and reveal that the model consistently exploits these rewards to obtain high scores while producing visually unrealistic, artifact-heavy images (Section ~\ref{sec:hacking_dynamics}). Notably, this behavior persists across a wide range of reward settings.

\subsection{Definition of Reward Hacking}
\label{sec:hacking_definition}
Reward hacking refers to the phenomenon where optimizing an \textit{imperfect proxy reward} results in degraded performance with respect to the \textit{true underlying objective} \cite{skalse2022defining}. In other words, reward hacking occurs when a model learns to exploit weaknesses in the reward function and achieve high scores under the designed proxy while performing poorly according to the true goal.

In the context of text-to-image generation, numerous reward functions have been proposed to guide model training. Each reward is typically designed to capture a specific aspect of generation quality, such as aesthetic appeal, human preference, or prompt-image alignment. Since these rewards are predefined and inherently partial, they are vulnerable to reward hacking \cite{liu2025flow}.

\subsection{Reward Choices and T2I RL Evaluation}
\label{sec:reward}

As discussed in Section~\ref{sec:evaluation_metric}, one major class of reward models for T2I RL is built upon vision-language encoders, primarily aiming to capture aesthetic appeal and human preference~\cite{wu2023human, schuhmann2022laion}. To analyze how such type of rewards affect model behavior, we select HPS~\cite{wu2023human} as a representative. HPS correlates strongly with human preference and aesthetic judgments, making it an ideal candidate to study potential reward hacking phenomena driven by subjective visual quality signals.

Another important aspect of T2I evaluation is image-prompt alignment, which measures how well the generated image reflects the semantics described in the text prompt. A common approach to enforcing this alignment is using an object detector as a compositional reward~\cite{liu2024grounding, cheng2024yolo, zhou2022simple}. Alternatively, recent works fine-tune MLLMs to directly assess whether an image matches the given prompt semantically~\cite{guo2025can, lin2024evaluating, tan2024evalalign, wang2025unified}. In our study, we adopt Grounding DINO~\cite{liu2024grounding} as a representative object detection-based reward, and finetuned ORM~\cite{guo2025can} as a representative MLLM-based prompt-image alignment reward.

We train Janus-Pro~\cite{chen2025janus} with these rewards individually and monitor their behaviors. The training dynamics are evaluated using various kinds of metrics:
\begin{itemize}
    \item Aesthetic/human preference-oriented: HPS~\cite{wu2023human}, Aesthetic Score~\cite{schuhmann2022laion}, PickScore~\cite{kirstain2023pick}, and DeQA~\cite{you2025teaching};
    \item Text-image alignment-based: GDino~\cite{liu2024grounding}, ORM~\cite{guo2025can}, and VQAScore~\cite{lin2024evaluating};
    \item Unified quality and alignment: UnifiedReward~\cite{wang2025unified}.
\end{itemize}

\subsection{Training Dynamics and Reward Hacking Behaviors}
\label{sec:hacking_dynamics}

\begin{figure*}[htbp]
    \centering
    
    \begin{subfigure}[b]{0.24\textwidth}
        \centering
        \includegraphics[width=\textwidth]{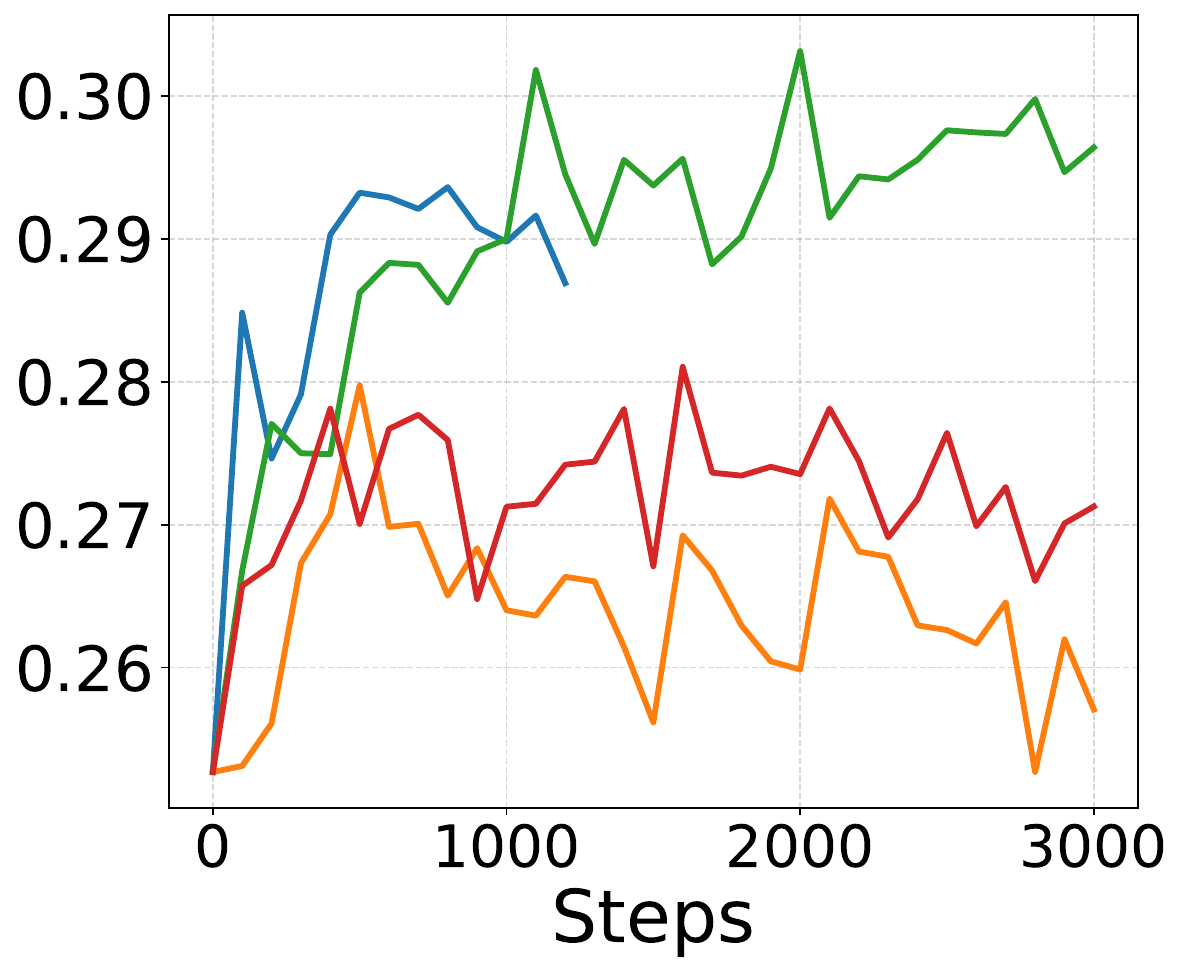}
        \caption{HPS - Object}
        \label{fig:dynamic:hps}
    \end{subfigure}
    \hfill
    \begin{subfigure}[b]{0.24\textwidth}
        \centering
        \includegraphics[width=\textwidth]{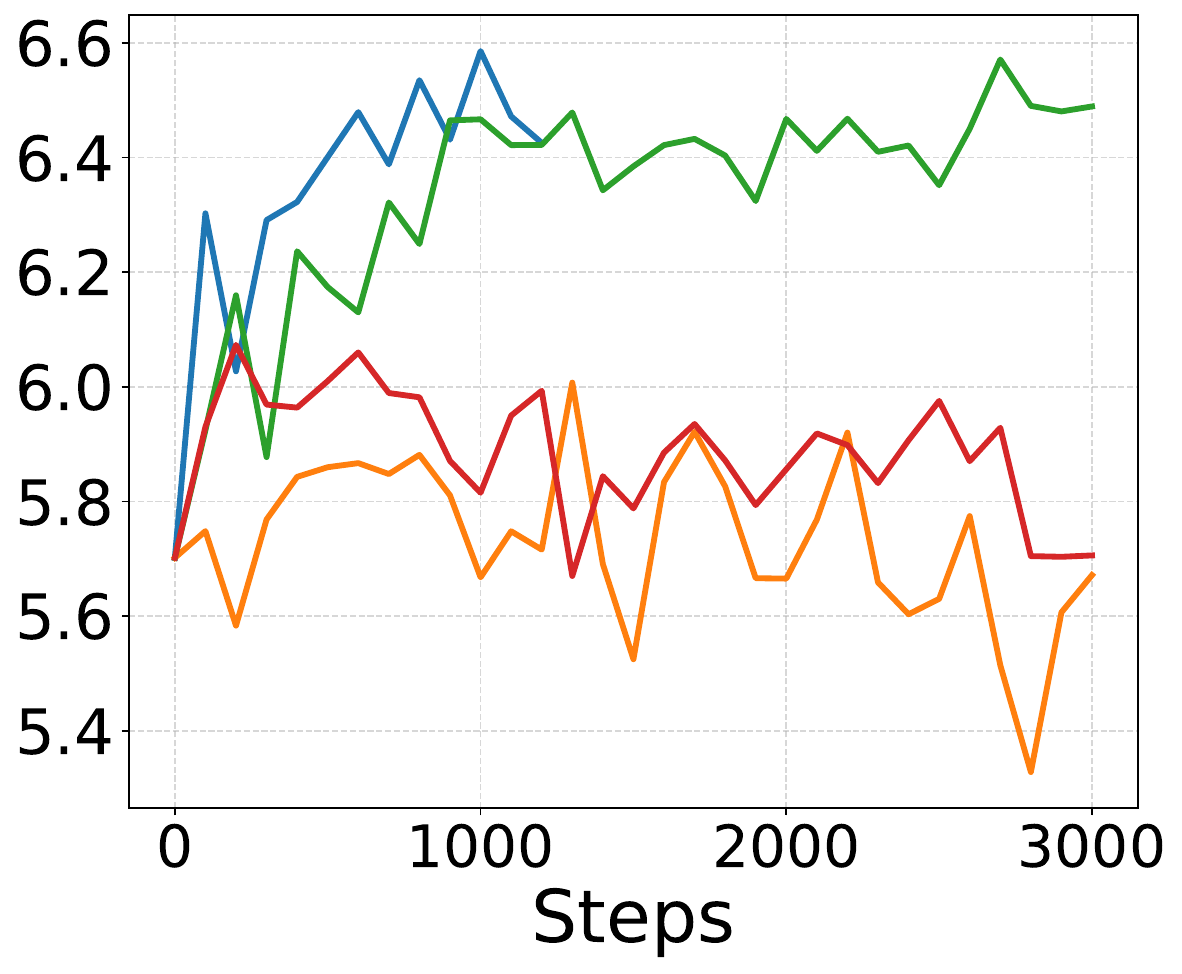}
        \caption{Aesthetic Score - Object}
        \label{fig:dynamic:aes}
    \end{subfigure}
    \hfill
    \begin{subfigure}[b]{0.24\textwidth}
        \centering
        \includegraphics[width=\textwidth]{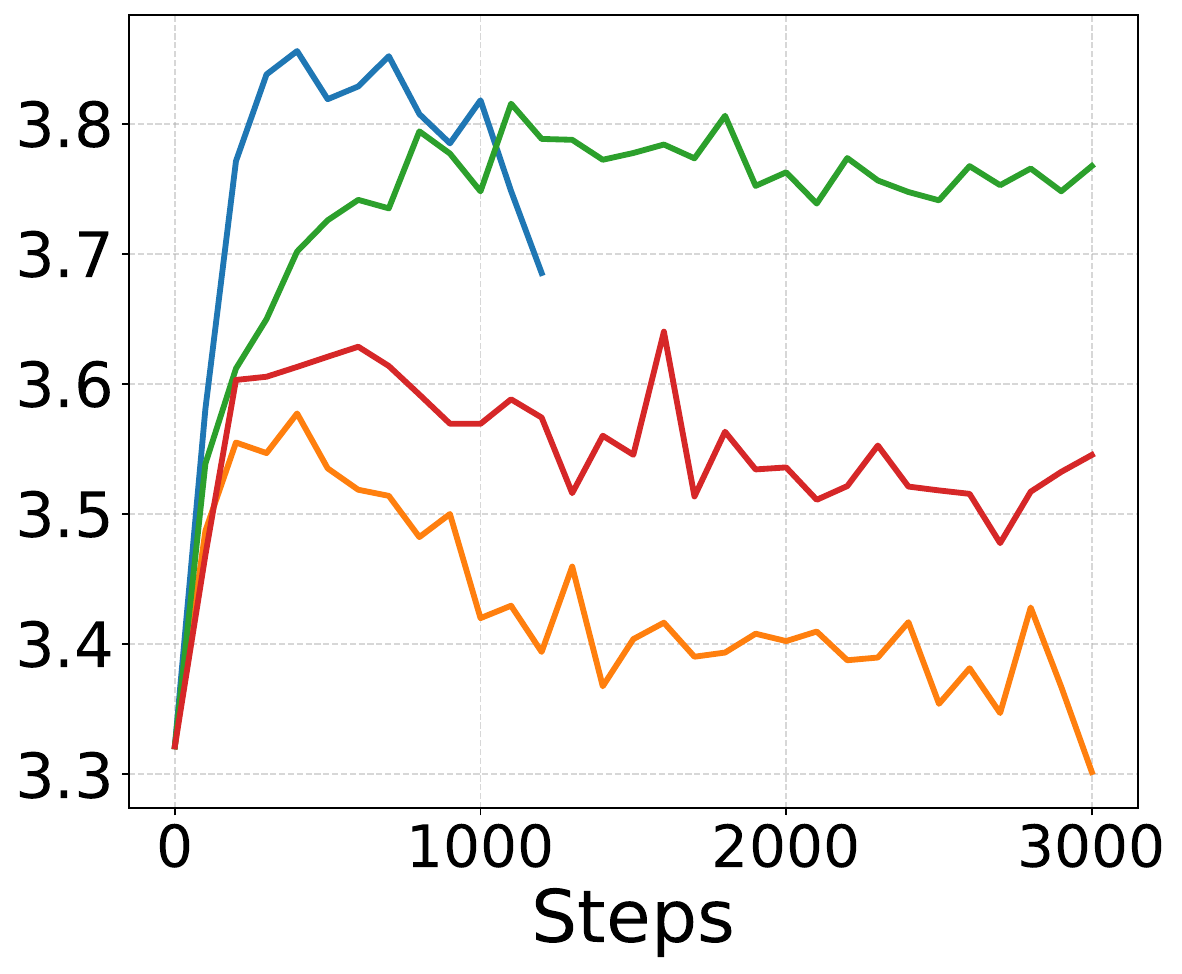}
        \caption{DeQA - Spatial}
        \label{fig:dynamic:deqa}
    \end{subfigure}
    \hfill
    \begin{subfigure}[b]{0.24\textwidth}
        \centering
        \includegraphics[width=\textwidth]{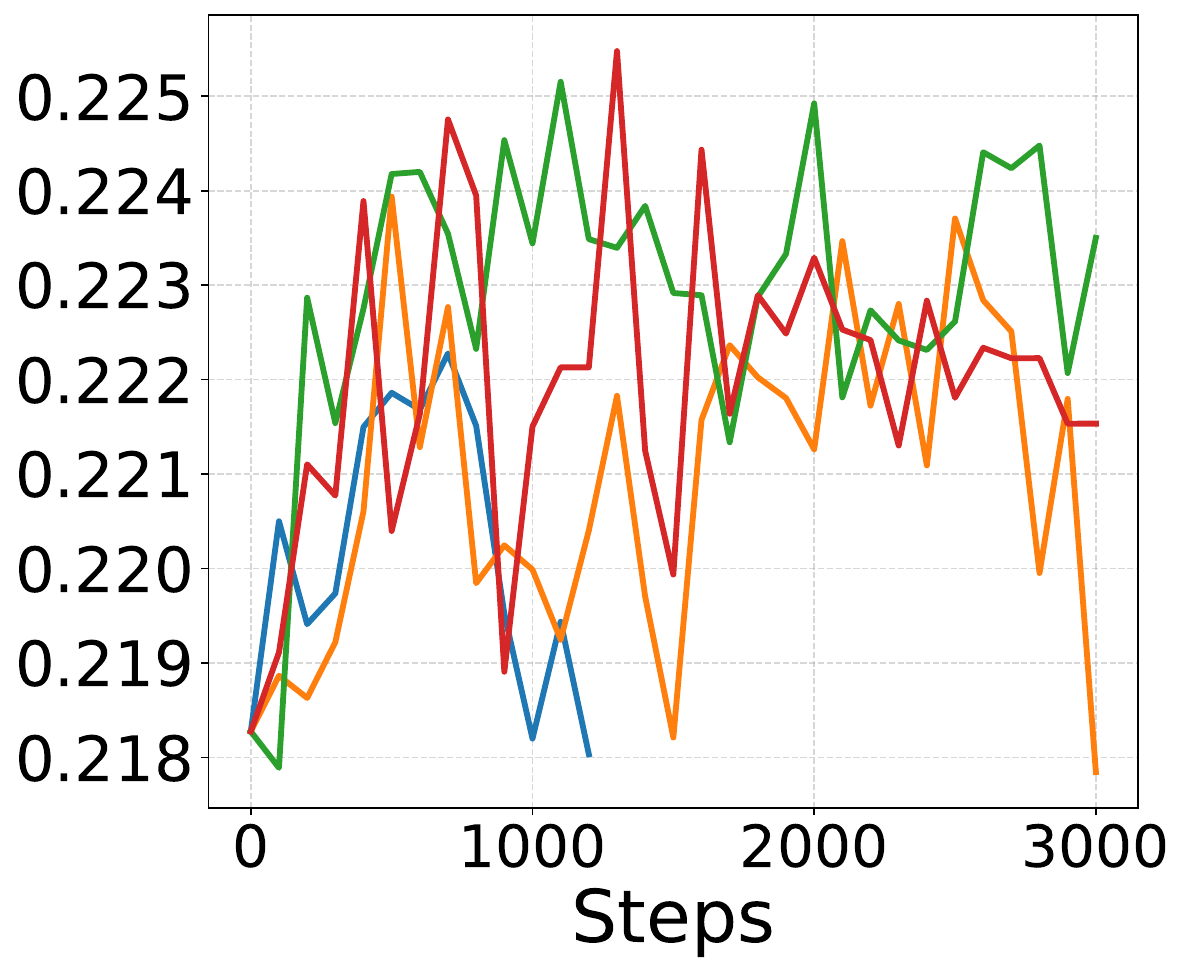}
        \caption{PickScore - Object}
        \label{fig:dynamic:pick}
    \end{subfigure}


    \vspace{0.3cm} 

    \begin{subfigure}[b]{0.24\textwidth}
        \centering
        \includegraphics[width=\textwidth]{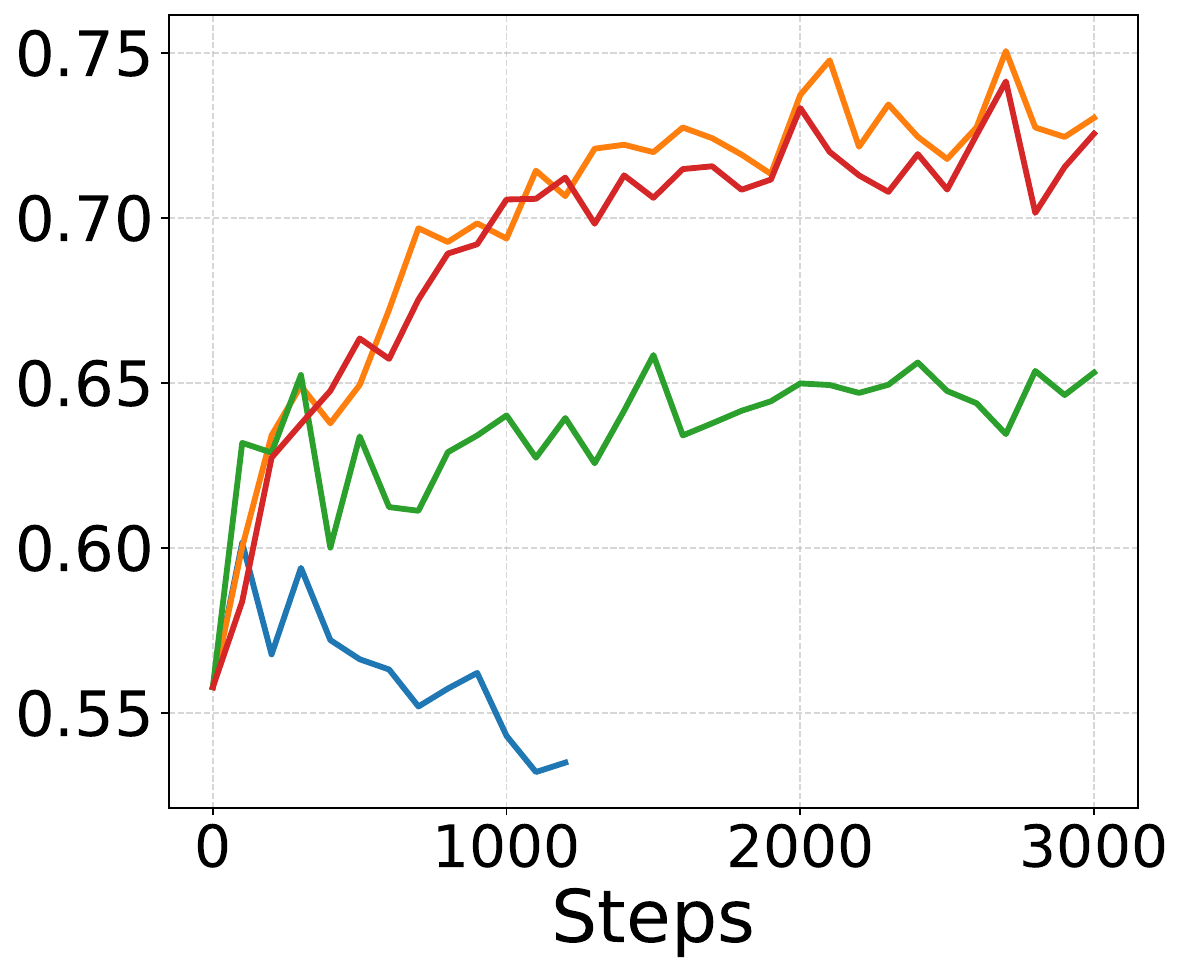}
        \caption{GDino - Spatial}
        \label{fig:dynamic:gdino}
    \end{subfigure}
    \hfill
    \begin{subfigure}[b]{0.24\textwidth}
        \centering
        \includegraphics[width=\textwidth]{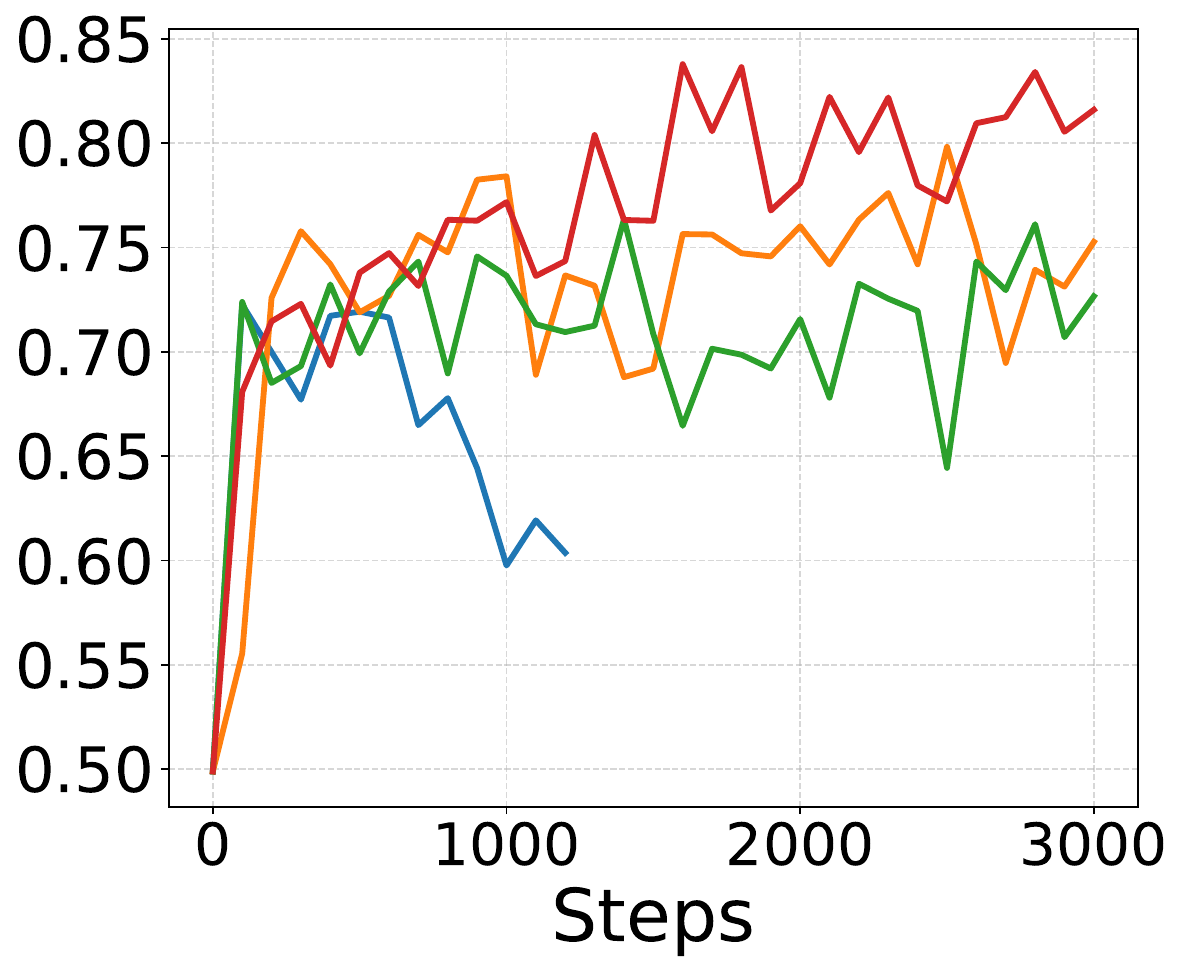}
        \caption{ORM - Color}
        \label{fig:dynamic:orm}
    \end{subfigure}
    \hfill
    \begin{subfigure}[b]{0.24\textwidth}
        \centering
        \includegraphics[width=\textwidth]{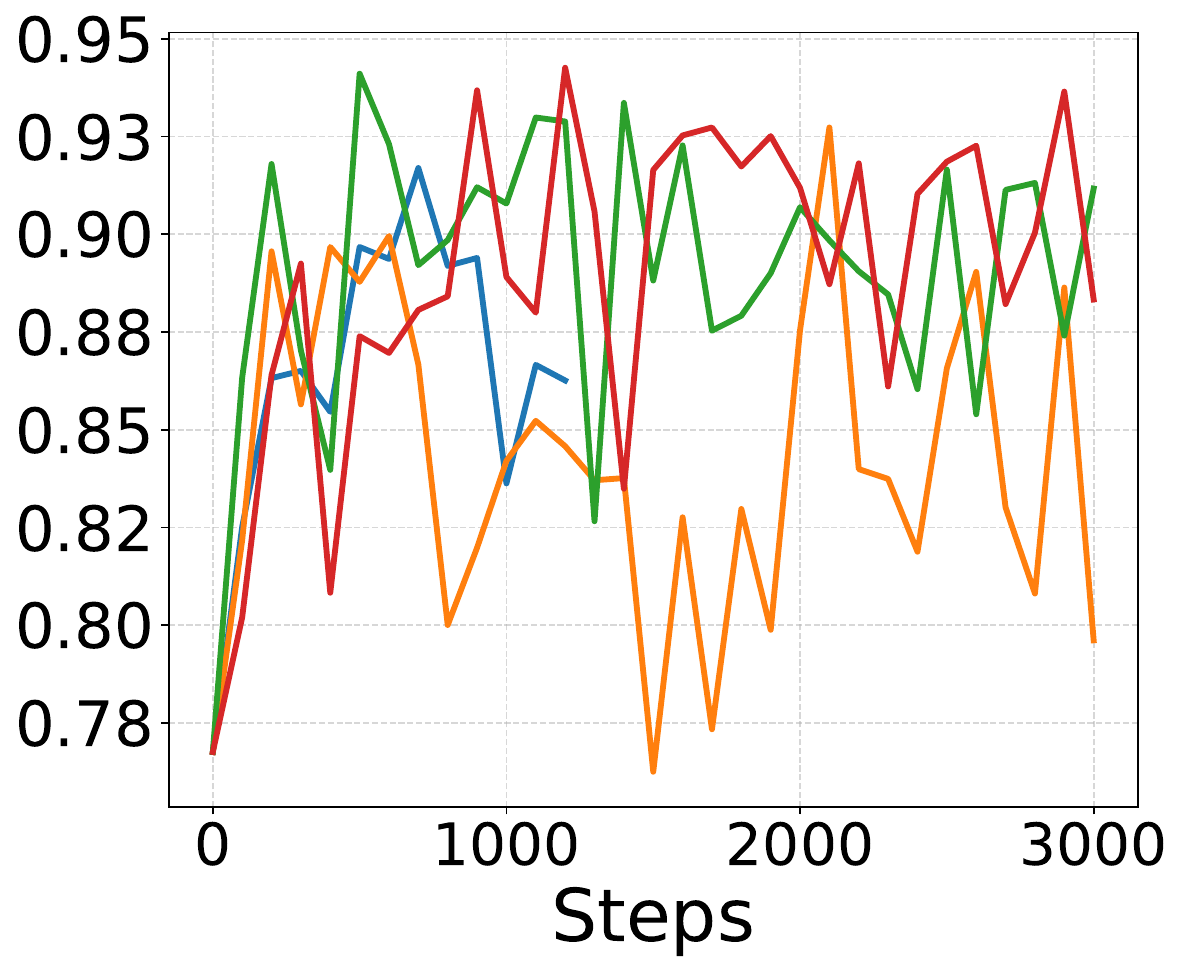}
        \caption{VQAScore - Object}
        \label{fig:dynamic:vqa}
    \end{subfigure}
    \hfill
    \begin{subfigure}[b]{0.24\textwidth}
        \centering
        \includegraphics[width=\textwidth]{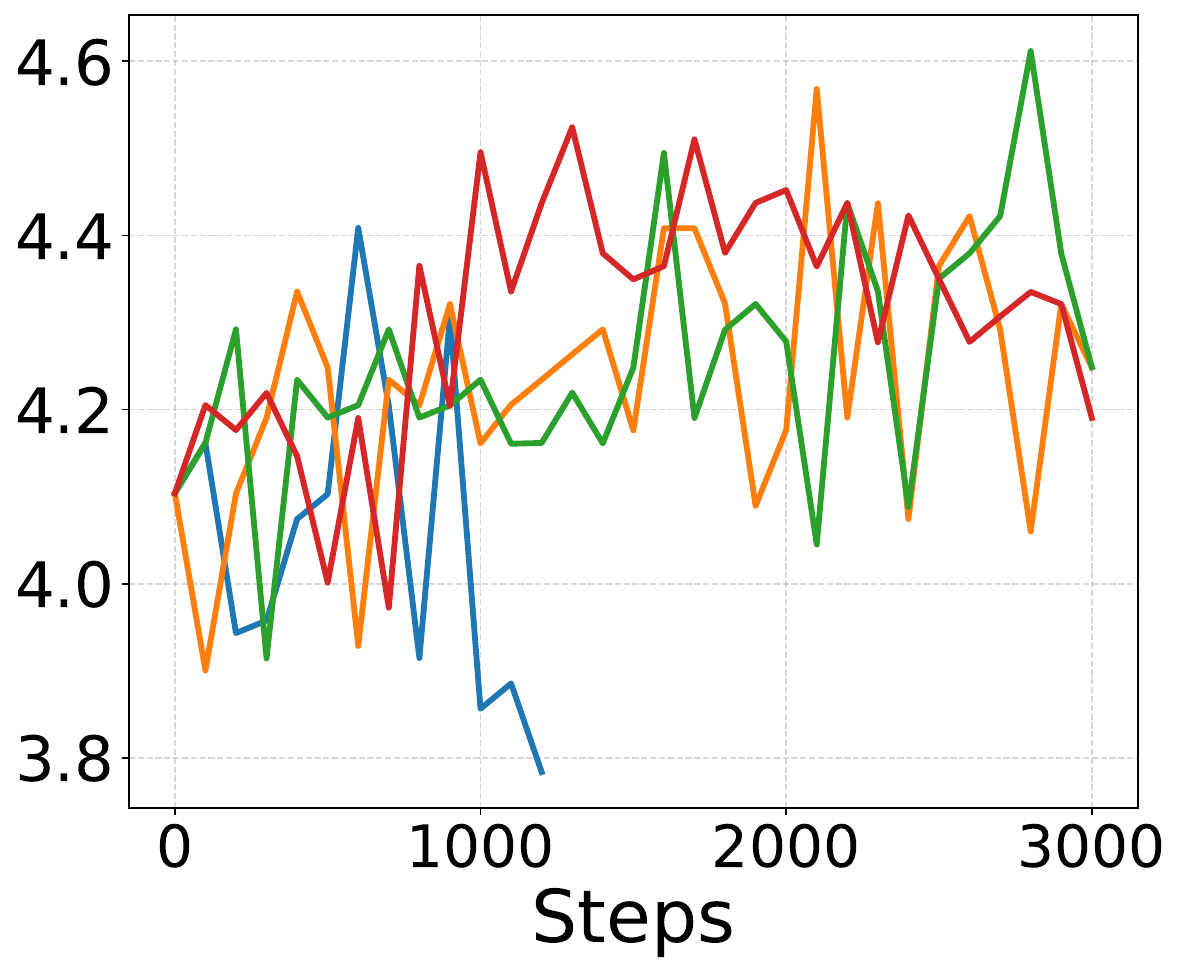}
        \caption{UnifiedReward - Object}
        \label{fig:dynamic:uni}
    \end{subfigure}

    \vspace{0.3cm} 

    \begin{subfigure}[b]{0.24\textwidth}
        \centering
        \includegraphics[width=\textwidth]{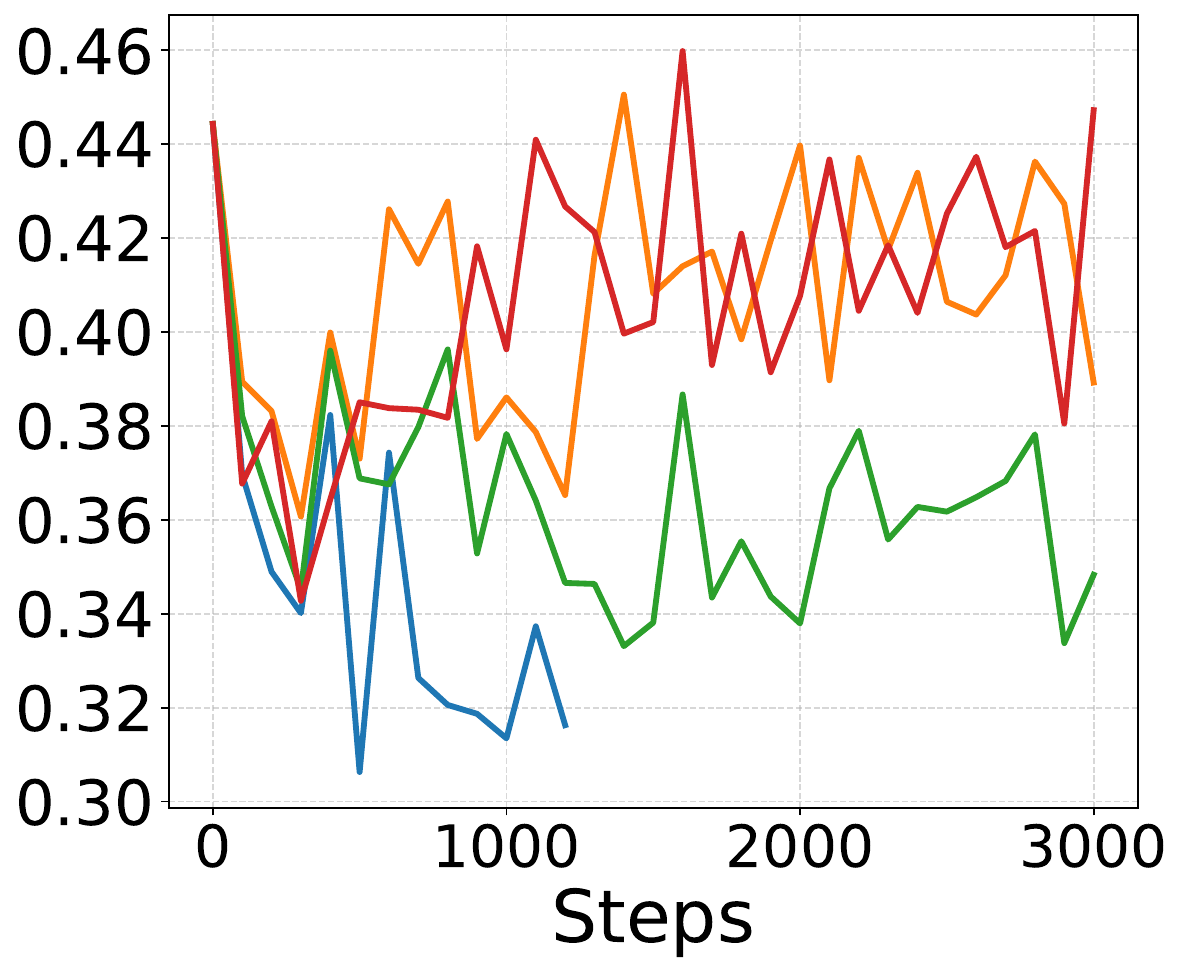}
        \caption{ArtifactR - Object}
        \label{fig:dynamic:arti_a}
    \end{subfigure}
    \hfill
    \begin{subfigure}[b]{0.24\textwidth}
        \centering
        \includegraphics[width=\textwidth]{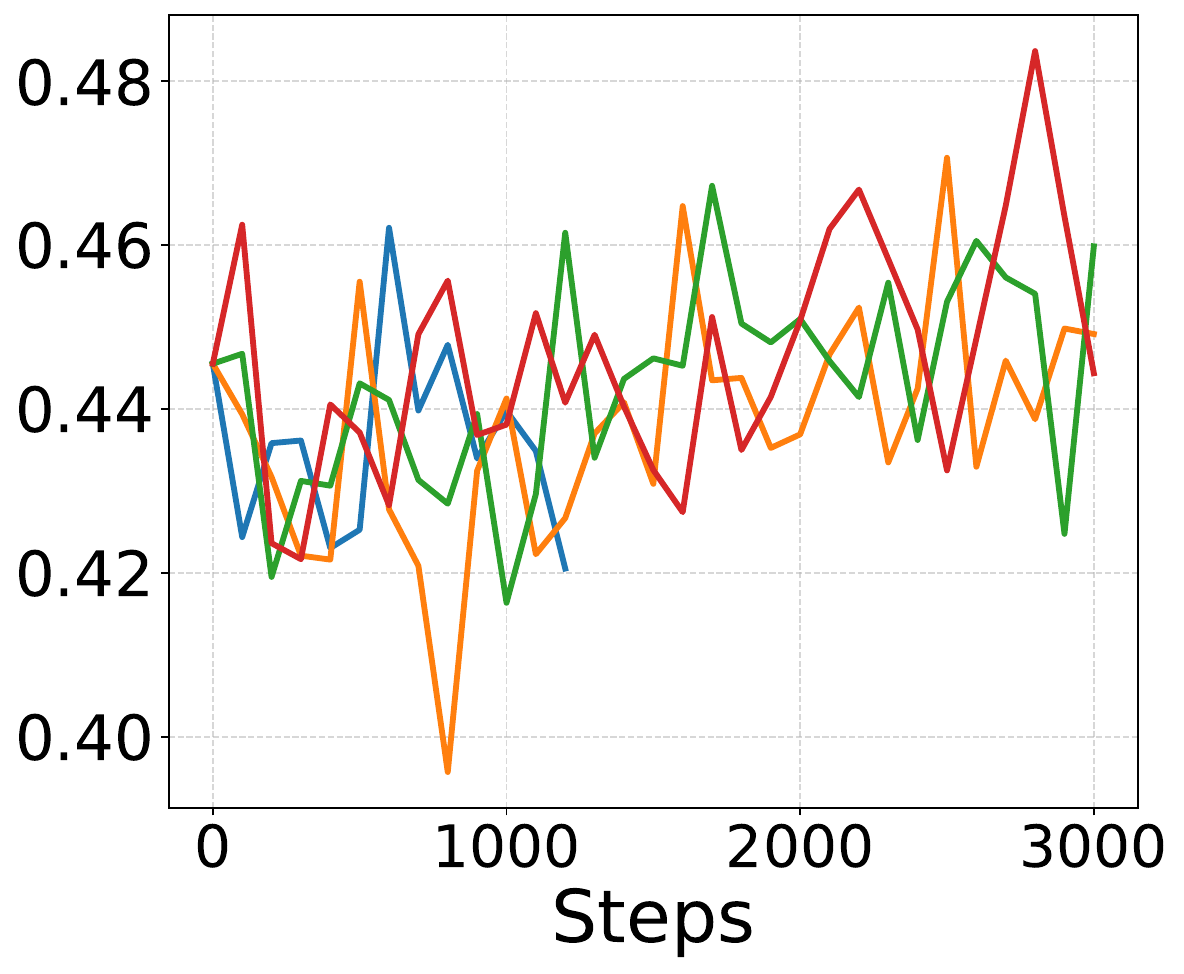}
        \caption{ArtifactR - Shape}
        \label{fig:dynamic:arti_b}
    \end{subfigure}
    \hfill
    \begin{subfigure}[b]{0.24\textwidth}
        \centering
        \includegraphics[width=\textwidth]{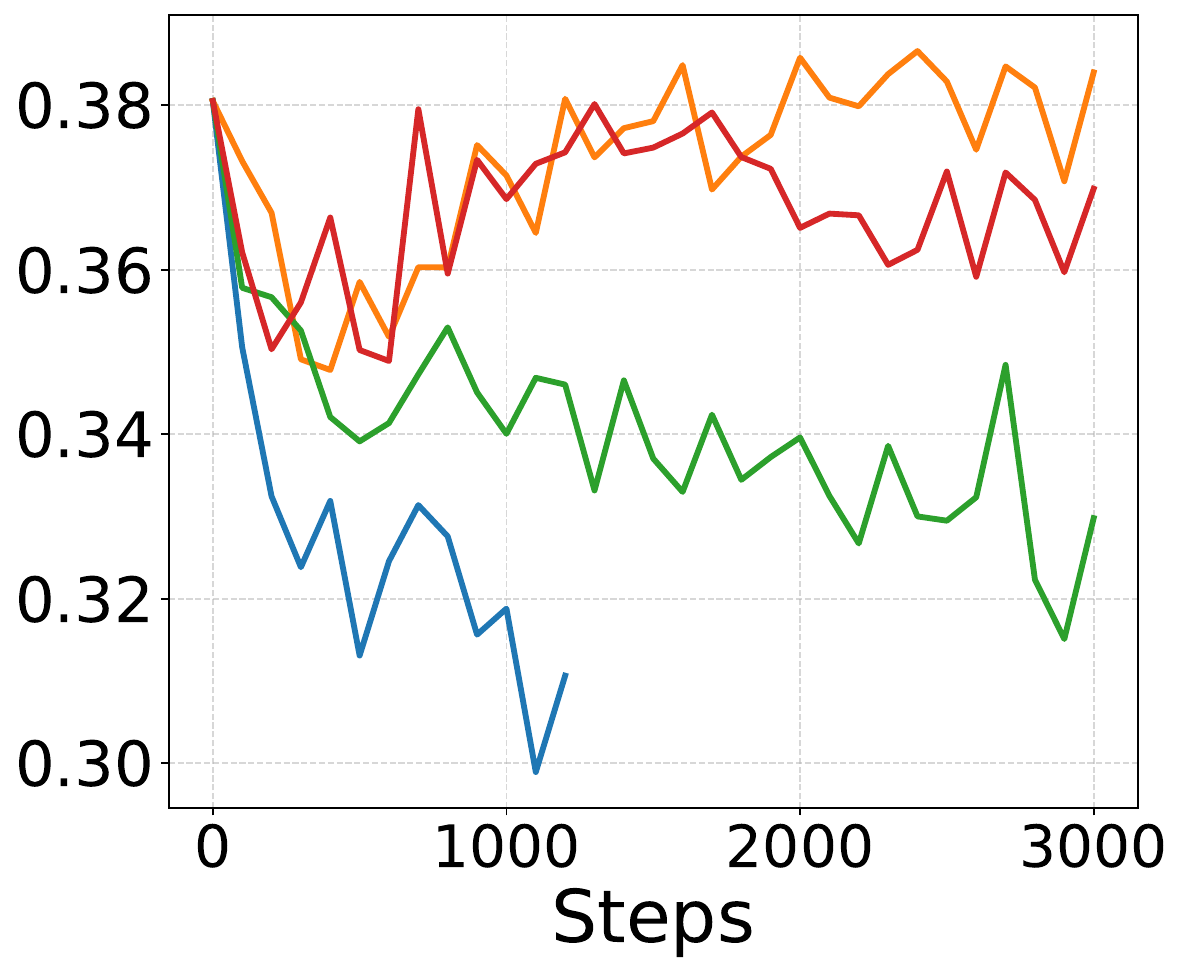}
        \caption{ArtifactR - Spatial}
        \label{fig:dynamic:arti_c}
    \end{subfigure}
    \hfill
    \begin{subfigure}[b]{0.24\textwidth}
        \centering
        \includegraphics[width=\textwidth]{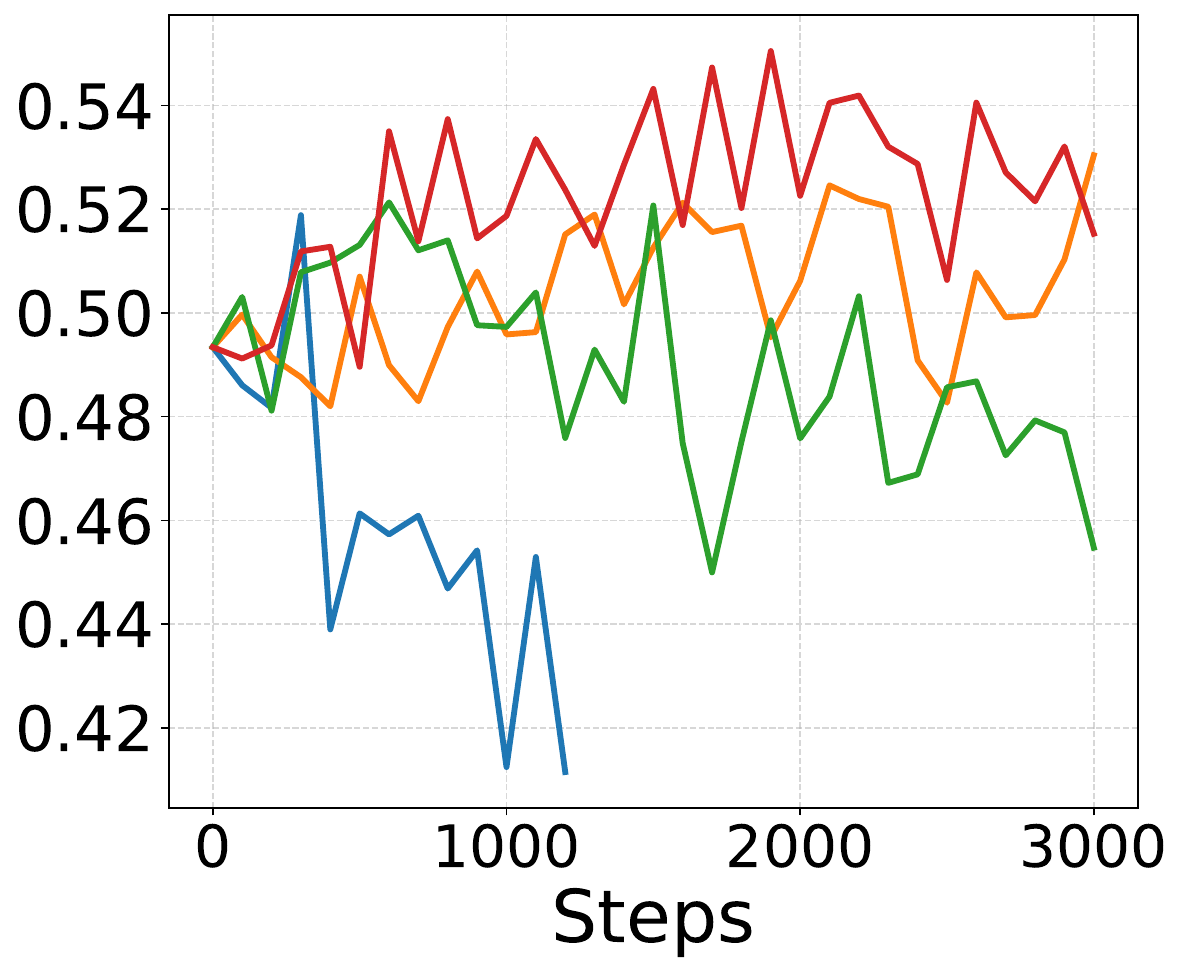}
        \caption{ArtifactR - Texture}
        \label{fig:dynamic:arti_d}
    \end{subfigure}

    \caption{\small 
    Evolution of metrics over training steps trained on Janus-Pro-1B~\cite{chen2025janus}. {\color{tabblue}Blue} color denotes the model trained with HPS~\cite{wu2023human}; {\color{taborange}Orange} color denotes the model trained with GDino~\cite{liu2024grounding}; {\color{tabgreen}Green} color denotes the model trained with HPS and GDino; {\color{tabred}Red} color denotes the model trained with finetuned ORM~\cite{guo2025can}.
    }
    \label{fig:dynamic}

\end{figure*}

We finetune the Janus-Pro~\cite{chen2025janus} on the same training set used in T2I-R1~\cite{jiang2025t2i}, with three epochs of RL training (approximately 3000 steps) under different reward configurations.
Figure~\ref{fig:dynamic} (and Supplementary Figure~\ref{fig:dyn_supp}) show the evolution of various evaluation metrics throughout training. We only visualize the first 1200 steps for the HPS~\cite{wu2023human} reward, as its metric values begin to collapse shortly after this point.

\vspace{-8pt}
\paragraph{Reward-specific improvements and cross-metric degradation.}
Across most metrics, the model primarily improves on the metric family aligned with its training reward, but fails to generalize to others. For instance, models trained with GDino~\cite{liu2024grounding} or ORM~\cite{guo2025can} show steady gains on their own reward values, yet exhibit clear degrade in human preference, aesthetic quality, and DeQA scores (Figure~\ref{fig:dynamic:hps}-\ref{fig:dynamic:deqa}).
Conversely, the model trained with HPS~\cite{wu2023human} shows rising HPS reward and aesthetic scores but experiences clear drops on compositional correctness and semantic consistency (Figure~\ref{fig:dynamic:gdino}-\ref{fig:dynamic:orm}). Other evaluation metrics, such as PickScore (Figure~\ref{fig:dynamic:pick}), VQAScore (Figure~\ref{fig:dynamic:vqa}), and UnifiedReward (Figure~\ref{fig:dynamic:uni}), show no consistent upward trend across training steps, regardless of reward types.
Training with combined rewards (HPS + GDino) yields more stable and moderate improvements, suggesting partial complementary effects between the two reward signals. However, this ensemble only partially mitigates the issue, as reward-specific biases still remain.
Overall, this cross-metric mismatch reveals that the model is optimizing narrowly for the specific reward signals rather than true improvements in generation quality, which is a direct manifestation of reward hacking.

\begin{table*}[h!]
\small
  \caption{\small Image Illustration. The first column is the prompt. The second and fourth columns are from models trained with existing rewards. The third and fifth columns are from models trained with the help of our ArtifactReward. The label above each image is the reward used in RL training.}
\label{tab:image}
\centering
\renewcommand{\arraystretch}{1.25} 

\begin{tabular}{|m{0.1\textwidth}|c|c|c|c|}
\toprule 

&
\textbf{\textcolor{red}{HPS}} &
\textbf{\textcolor{green!50!black}{HPS + ArtifactR}} &
\textbf{\textcolor{red}{GDino}} &
\textbf{\textcolor{green!50!black}{GDino + ArtifactR}} \\

\multirow{2}{=}{A photo of a cup above a tennis racket} &
\includegraphics[width=0.17\textwidth]{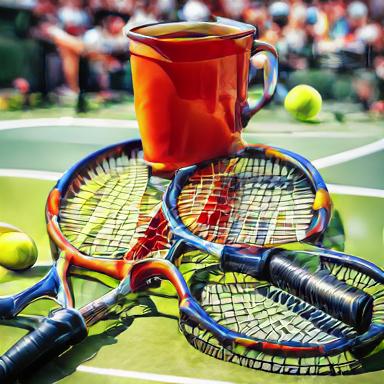} &
\includegraphics[width=0.17\textwidth]{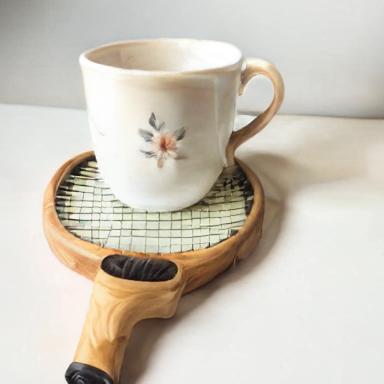} &
\includegraphics[width=0.17\textwidth]{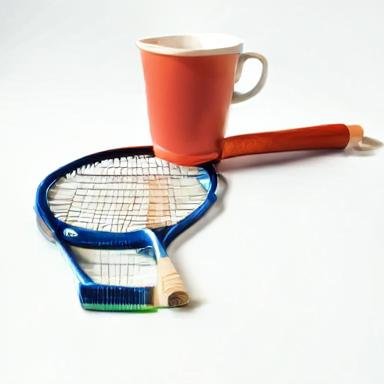} &
\includegraphics[width=0.17\textwidth]{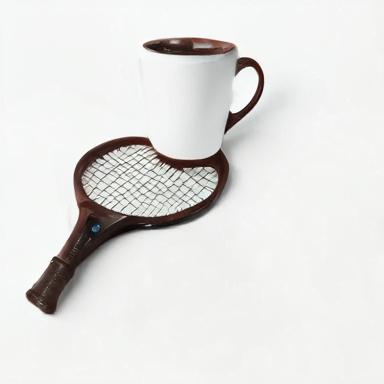} \\
\cline{2-5}

&
\textbf{\textcolor{red}{HPS + GDino}} &
\textbf{\textcolor{green!50!black}{HPS + GDino + ArtifactR}} &
\textbf{\textcolor{red}{ORM}} &
\textbf{\textcolor{green!50!black}{ORM + ArtifactR}} \\

&
\includegraphics[width=0.17\textwidth]{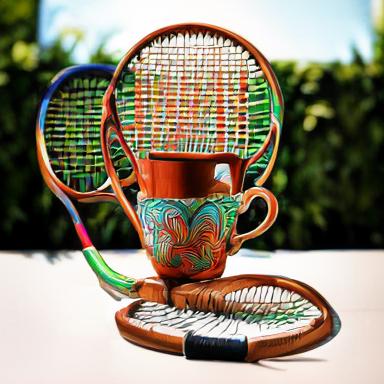} &
\includegraphics[width=0.17\textwidth]{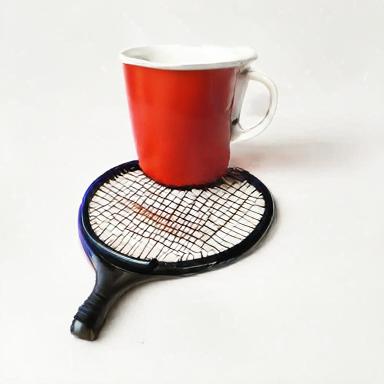} &
\includegraphics[width=0.17\textwidth]{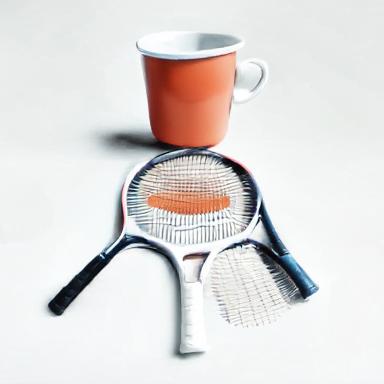} &
\includegraphics[width=0.17\textwidth]{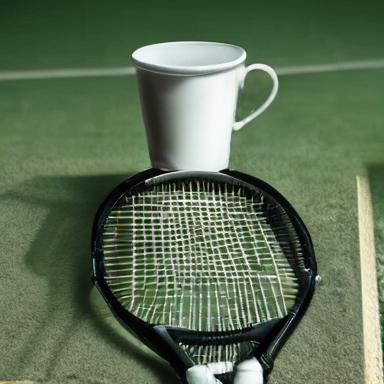} \\

\midrule

&
\textbf{\textcolor{red}{HPS}} &
\textbf{\textcolor{green!50!black}{HPS + ArtifactR}} &
\textbf{\textcolor{red}{GDino}} &
\textbf{\textcolor{green!50!black}{GDino + ArtifactR}} \\

\multirow{2}{=}{A photo of a person right of a bear} &
\includegraphics[width=0.17\textwidth]{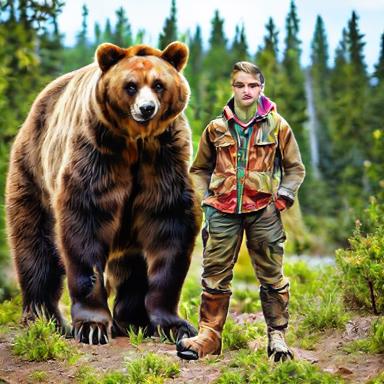} &
\includegraphics[width=0.17\textwidth]{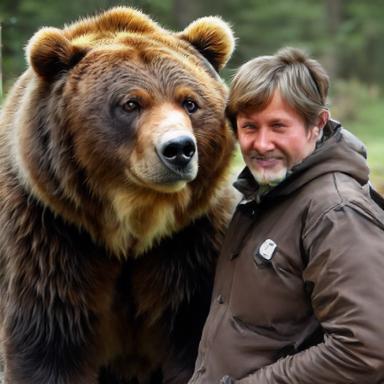} &
\includegraphics[width=0.17\textwidth]{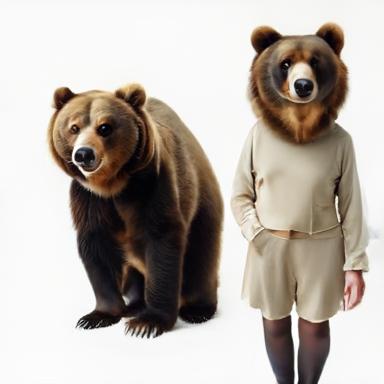} &
\includegraphics[width=0.17\textwidth]{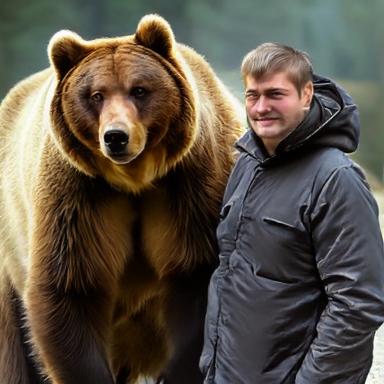} \\
\cline{2-5}

&
\textbf{\textcolor{red}{HPS + GDino}} &
\textbf{\textcolor{green!50!black}{HPS + GDino + ArtifactR}} &
\textbf{\textcolor{red}{ORM}} &
\textbf{\textcolor{green!50!black}{ORM + ArtifactR}} \\

&
\includegraphics[width=0.17\textwidth]{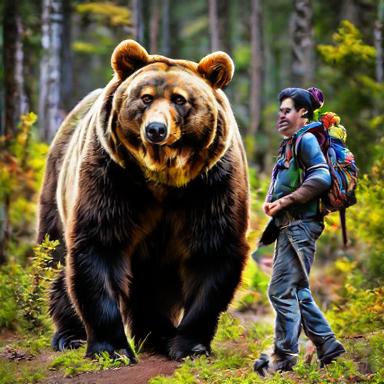} &
\includegraphics[width=0.17\textwidth]{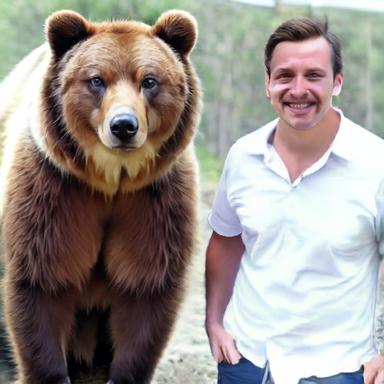} &
\includegraphics[width=0.17\textwidth]{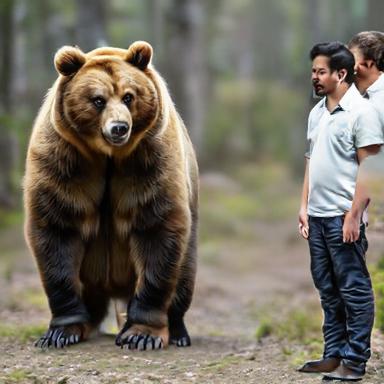} &
\includegraphics[width=0.17\textwidth]{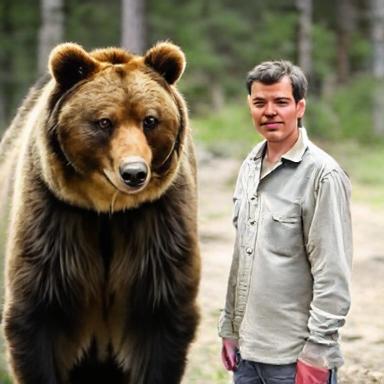} \\

\midrule

&
\textbf{\textcolor{red}{HPS}} &
\textbf{\textcolor{green!50!black}{HPS + ArtifactR}} &
\textbf{\textcolor{red}{GDino}} &
\textbf{\textcolor{green!50!black}{GDino + ArtifactR}} \\

\multirow{3}{=}{A person is looking at a map and planning a hiking trail.} &
\includegraphics[width=0.17\textwidth]{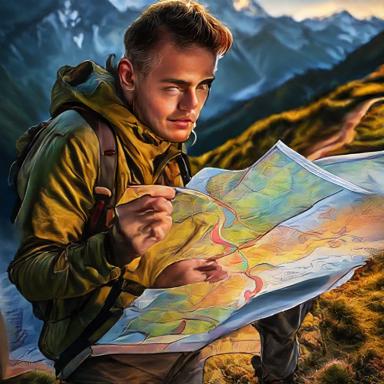} &
\includegraphics[width=0.17\textwidth]{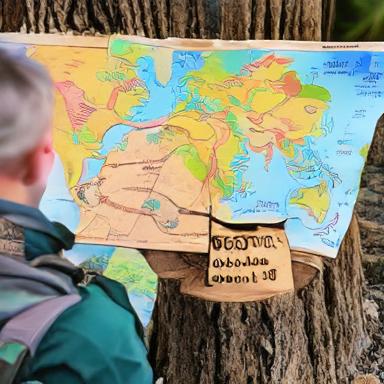} &
\includegraphics[width=0.17\textwidth]{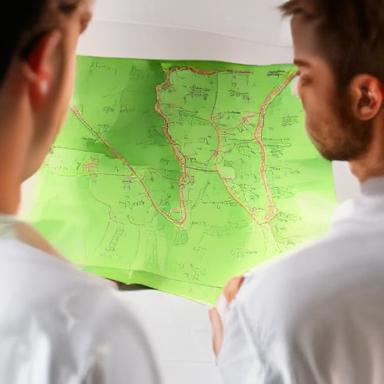} &
\includegraphics[width=0.17\textwidth]{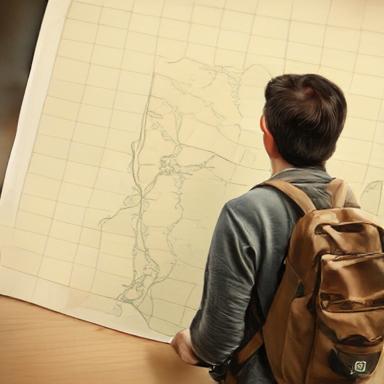} \\
\cline{2-5}

&
\textbf{\textcolor{red}{HPS + GDino}} &
\textbf{\textcolor{green!50!black}{HPS + GDino + ArtifactR}} &
\textbf{\textcolor{red}{ORM}} &
\textbf{\textcolor{green!50!black}{ORM + ArtifactR}} \\

&
\includegraphics[width=0.17\textwidth]{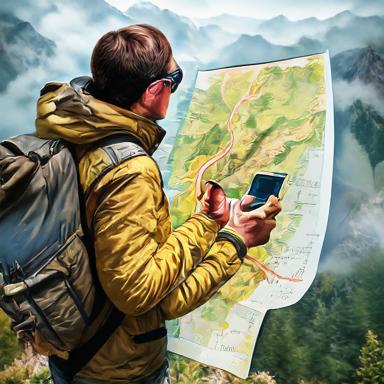} &
\includegraphics[width=0.17\textwidth]{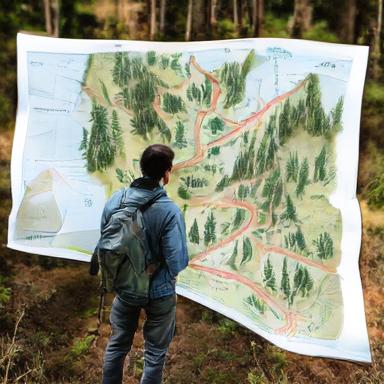} &
\includegraphics[width=0.17\textwidth]{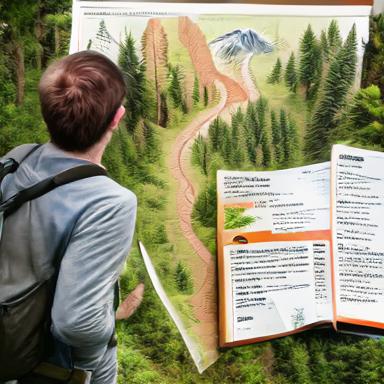} &
\includegraphics[width=0.17\textwidth]{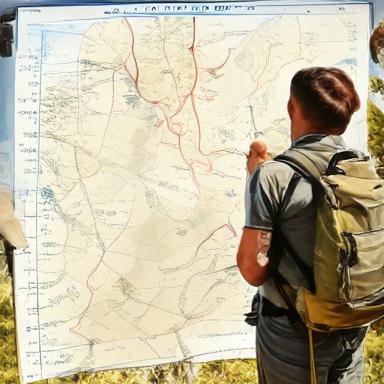} \\


\bottomrule
\end{tabular}
\end{table*}

\vspace{-8pt}
\paragraph{Visual inspection of reward-driven biases.}
By examining the generated images, distinct reward-induced tendencies become apparent (see Table~\ref{tab:image}).
The HPS-trained model tends to produce vibrant backgrounds and visually striking colors, since HPS implicitly favors aesthetic appeal. However, over-emphasizing colorfulness often leads to unrealistic and over-saturated images, sacrificing spatial grounding and object fidelity. Even when combined with GDino, this issue persists, as training remains biased toward HPS gains (Figure~\ref{fig:dynamic:hps} and \ref{fig:dynamic:gdino}), and thus the images still look unrealistic and show saturation. 
In contrast, GDino or ORM-trained models generate simpler scenes and object-centered compositions. This behavior naturally reflects their design focusing on object detectability and entity match. Thus, these models achieve stronger text-image grounding but reduced diversity and realism.

\vspace{-8pt}
\paragraph{Universal reward failure: structural artifacts.}
While the above analysis highlights reward-specific biases and cross-metric degradation, we further identify a more pervasive and universal reward failure pattern in T2I RL: structural artifacts.
Across the qualitative results in Table~\ref{tab:image}, generated images frequently exhibit geometric distortions (e.g., physically implausible or distorted tennis rackets), object duplication or fragmentation (e.g., duplicated limbs or overlapping object outlines), and even blending between different entities (e.g., a human body merged with a bear head).
Remarkably, these structural failures arise consistently under every reward configuration (HPS, GDino, ORM, and HPS + GDino), revealing a shared \textbf{systematic blind spot} in current reward designs. None of them reliably enforce local structure or fine-grained object correctness. As a result, the policy model can exploit this loophole by optimizing rewards without genuinely learning to produce geometrically coherent images. This behavior illustrates a deeper form of reward hacking, where models achieve higher scores by bypassing true structural correctness.

\vspace{-8pt}
\paragraph{Summary.}
Together, these findings reveal a clear and pervasive reward hacking in T2I RL training. Each reward guides the model toward narrow patterns that maximize its own metric while degrading other aspects of performance. The model learns to exploit superficial cues favored by the reward, such as over-saturated colors for HPS or object-centric simplifications for GDino and ORM. Even joint-optimization with multiple rewards (e.g., HPS + GDino) provides only partial mitigation, as the underlying biases remain.
More fundamentally, current reward formulations lack a mechanism that discourages artifact-driven shortcuts. Without constraints enforcing structural plausibility or fine-grained correctness, the model is free to achieve higher scores through geometrically distorted and implausible generations.

\section{Mitigating Reward Hacking via Artifact-Aware Reward Design}
\label{sec:method}

\begin{table}
    \small
  \caption{\small Accuracy of different reward models in assigning higher scores to artifact-free images compared to artifact-containing ones. Each image pair is scored as 1 if the reward assigns a higher score to the artifact-free image, 0.5 if the scores are tied, and 0 if the artifact-containing image receives a higher score. The accuracy is averaged across all pairs.}
  \label{tab:artifacts}
  \centering
  \resizebox{1.0\linewidth}{!}
  {
  \begin{tabular}{l|c|c|c|c}
    \toprule
    \textbf{Rewards}  &\textbf{Correct} &\textbf{Tie} &\textbf{Wrong} &\textbf{Accuracy}\\
    \midrule
    HPS~\cite{wu2023human}  &108 &0 &94 &0.53 \\
    Aesthetic Score~\cite{schuhmann2022laion}  &84 &0 &118 &0.42 \\
    ImageReward~\cite{xu2023imagereward}  &133 &7 &62 &0.68 \\
    MANIQA~\cite{yang2022maniqa}  &80 &0 &122 &0.39 \\
    DeQA~\cite{you2025teaching}  &104 &0 &98 &0.51 \\
    GDino~\cite{liu2024grounding}  &46 &143 &13 &0.58 \\
    ORM~\cite{guo2025can}  &39 &149 &14 &0.56 \\
    UnifiedReward~\cite{wang2025unified} &86 &92 &24 &0.65 \\
    \midrule
    ArtifactReward (Ours) &161 &2 &39 &0.80 \\
    \bottomrule
  \end{tabular}
  }
\end{table}

To quantitatively verify the reward model reliability of penalizing artifact generation, we curated a small diagnostic dataset annotated by human judgment. For each prompt, we formed pairs of one artifact-free image and one artifact-containing image, which allows a controlled comparison of reward model behaviors. Representative examples are provided in Table~\ref{tab:artifact_img}.

\begin{table}[h!]
\small
  \caption{\small Examples of curated artifact diagnostic dataset. The upper images show no artifacts, while the lower images contain artifacts.}
\label{tab:artifact_img}
\centering

\resizebox{1.0\linewidth}{!}{
\begin{tabular}{cl|cl}
\toprule 

\multicolumn{2}{l|}{rubber sole shoes and a wooden chair} &\multicolumn{2}{l}{a wooden table and a fabric hat} \\
\midrule
 \multirow{7}{*}{\includegraphics[width=0.16\textwidth]{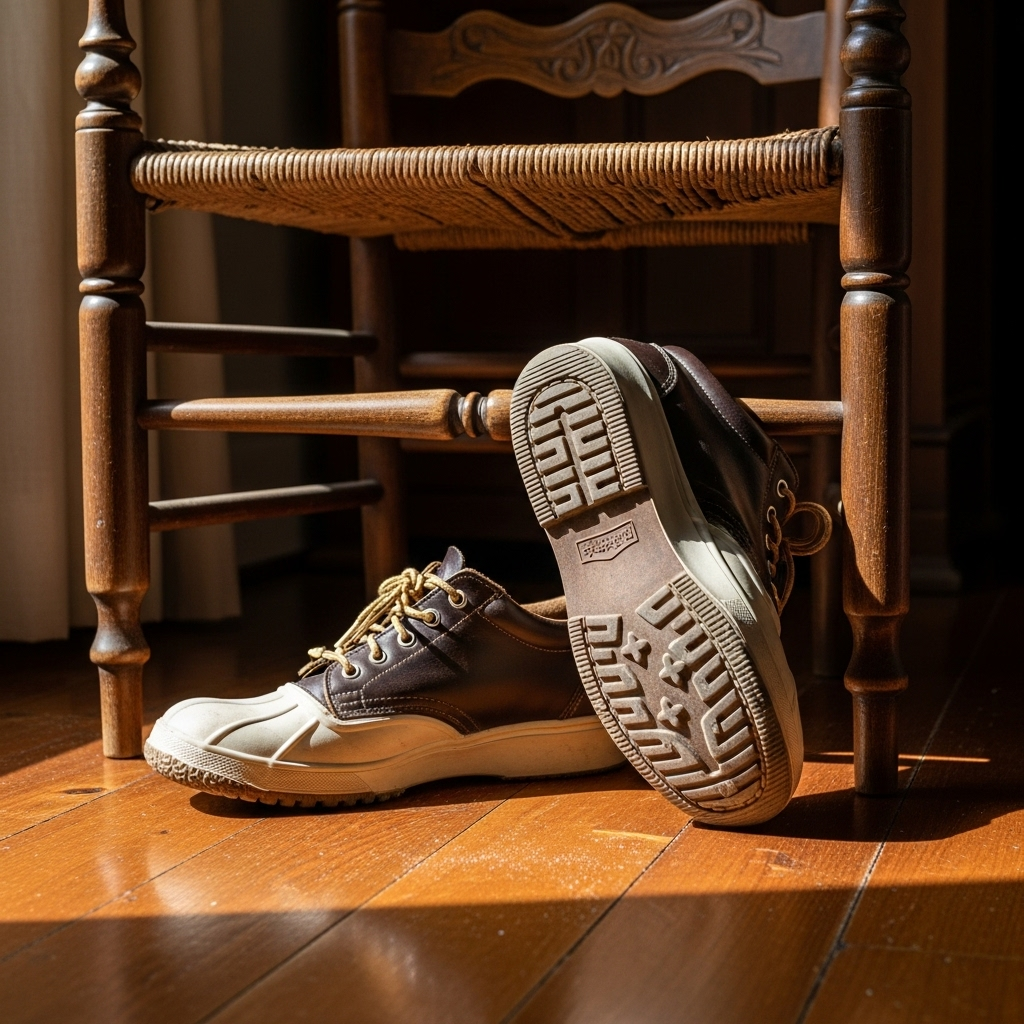}} & & \multirow{7}{*}{\includegraphics[width=0.16\textwidth]{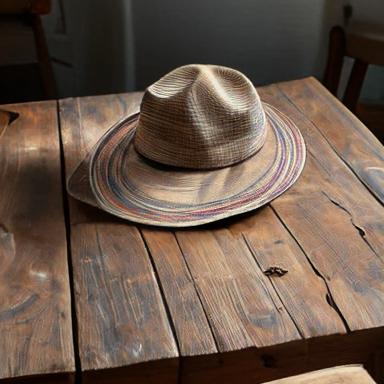}} &\\
 & & &\\
 &HPS: 0.3293 & &HPS: 0.2878\\
 &GDino: 0.4 & &GDino: 1.0\\
 &ORM: 1.0  & &ORM: 1.0\\
 &ArtifactR: 0.7549  & &ArtifactR: 0.9149\\
 & & &\\

 \midrule
  \multirow{7}{*}{\includegraphics[width=0.16\textwidth]{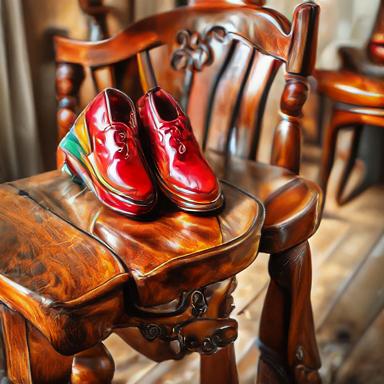}} & & \multirow{7}{*}{\includegraphics[width=0.16\textwidth]{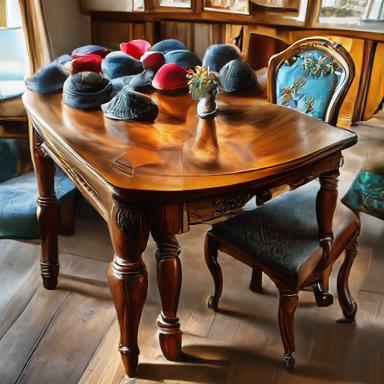}} &\\
 & & &\\
 &HPS: 0.3362 & &HPS: 0.2893\\
 &GDino: 0.4 & &GDino: 1.0\\
 &ORM: 1.0  & &ORM: 1.0\\
 &ArtifactR: 0.2689 & &ArtifactR: 0.3486 \\
 & & &\\

 \midrule
 \midrule

 \multicolumn{2}{l|}{a photo of a bird above a scissors} &\multicolumn{2}{l}{a photo of a handbag left of a toaster} \\
\midrule
 \multirow{7}{*}{\includegraphics[width=0.16\textwidth]{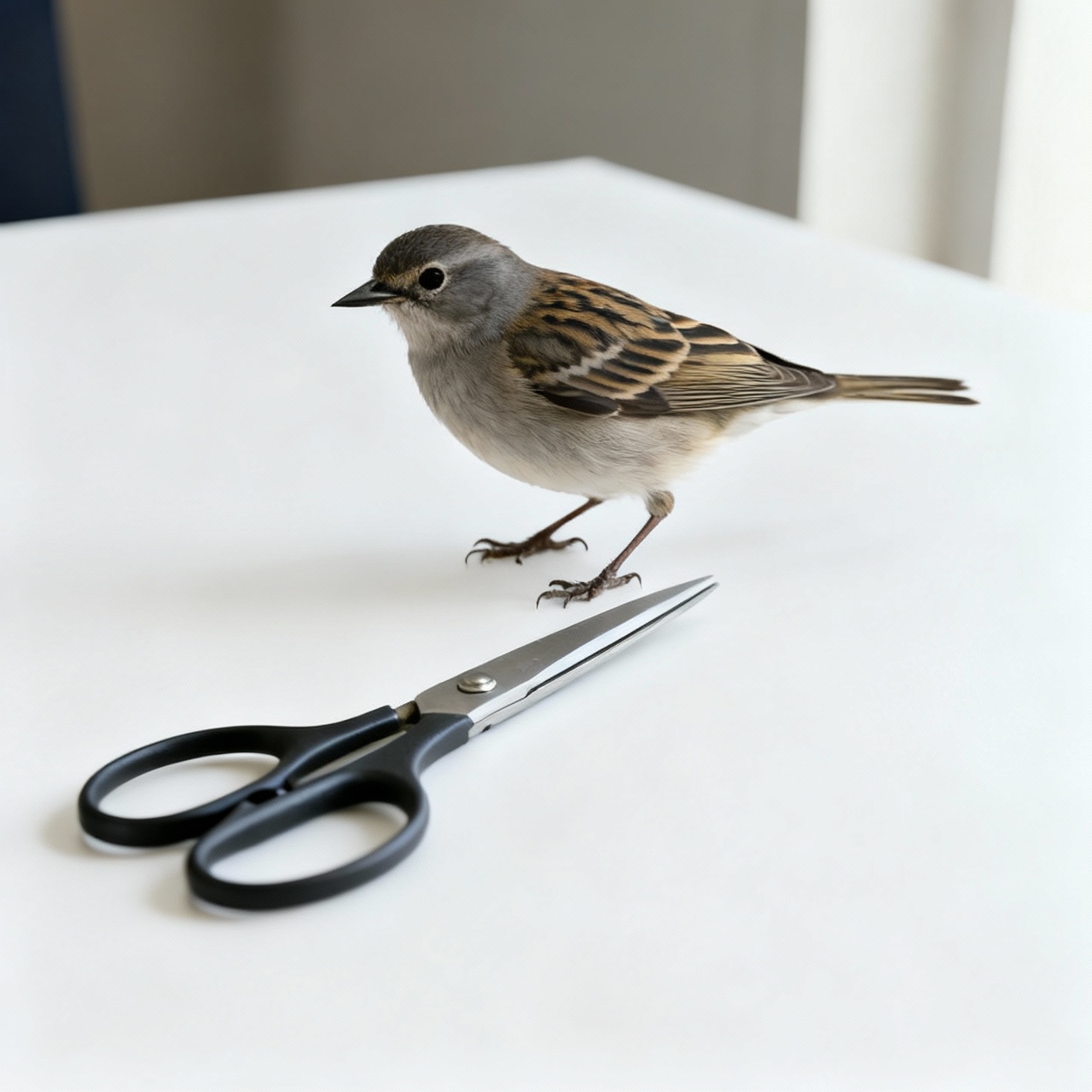}} & & \multirow{7}{*}{\includegraphics[width=0.16\textwidth]{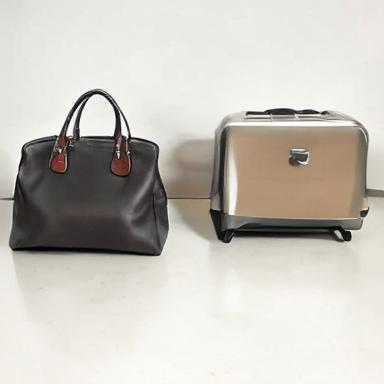}} &\\
 & & &\\
 &HPS: 0.2795 & &HPS: 0.2272\\
 &GDino: 1.0 & &GDino: 0.9\\
 &ORM: 0.0  & &ORM: 0.6027\\
 &ArtifactR: 0.8355  & &ArtifactR: 0.7879\\
 & & &\\

 \midrule
  \multirow{7}{*}{\includegraphics[width=0.16\textwidth]{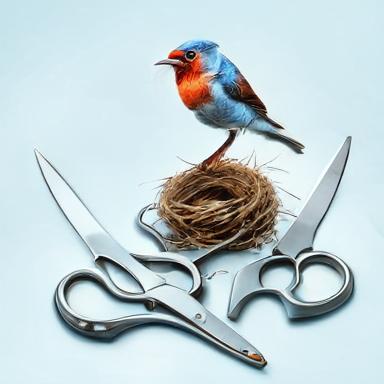}} & & \multirow{7}{*}{\includegraphics[width=0.16\textwidth]{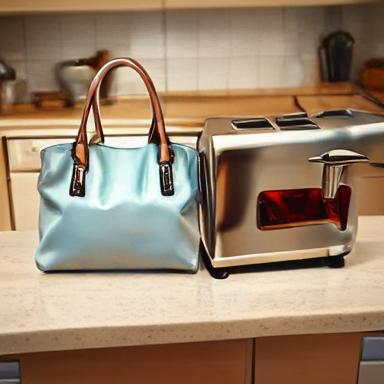}} &\\
 & & &\\
 &HPS: 0.2988 & &HPS: 0.2896\\
 &GDino: 1.0 & &GDino: 0.9\\
 &ORM: 0.0  & &ORM: 1.0\\
 &ArtifactR: 0.4844 & &ArtifactR: 0.4378 \\
 & & &\\

\bottomrule
\end{tabular}
}
\end{table}

We then evaluate existing reward models on this curated dataset to assess whether they can correctly assign higher scores to artifact-free images compared with artifact-containing images.
As summarized in Table~\ref{tab:artifacts}, none of the existing reward models consistently penalize images with clear structural issues. In many cases, they assign equal or even higher scores to flawed images. GDino, ORM and UnifiedReward, in particular, frequently produce identical scores across pairs (``Tie''), revealing a failure to detect fine-grained geometric inconsistencies.
This results expose a centural limitation in current T2I reward designs: they lack sensitivity to fine-grained local structural coherence and object plausibility, allowing models to exploit reward loopholes and generate visually implausible yet high-reward images.

\subsection{ArtifactReward: Complementary Reward via Automatic Prompt Optimization}
\label{sec:artifact_reward}
To address this limitation, we introduce \textbf{\textit{ArtifactReward}}, a complementary reward explicitly designed to penalize unrealistic or artifact-containing generations.

Rather than relying on large-scale human annotation, we leverage automatic prompt optimization (APO) \cite{pryzant2023automatic, shin2020autopo, ape, deng2022rlpo} to construct a lightweight, data-efficient proxy reward from limited labeled examples.

Specificly, we use Qwen2.5-VL-7B-Instruct~\cite{bai2025qwen2} as the backbone visual-language model. Details of the APO procedure for constructing ArtifactReward are provided in Supplementary~\ref{supp:apo}.
At inference time, the model is prompted with the optimized instruction and asked to determine whether the input image contains any unrealistic artifacts. The model outputs probabilities for the ``yes'' and ``no'' tokens. We define the ArtifactReward score as the normalized probability assigned to ``no'', which corresponds to the estimated likelihood that the image is artifact-free:

\begin{equation}
\label{artifact}
    R_{\text{ArtifactReward}}
    = \frac{1}{1+ e^{[\text{log}(p_{\text{yes}}) - \text{log}(p_{\text{no}})]}}.
\end{equation}

Through APO, the instruction is iteratively refined to maximize discrimination between artifact-free and artifact-containing samples. The resulting ArtifactReward demonstrates strong ability to detect structural inconsistencies and fine-grained artifacts (Table~\ref{tab:artifacts}), outperforming existing rewards in recognizing unrealistic generations.

We further evaluate ArtifactReward evolution in different existing reward configurations in training. As shown in Figure~\ref{fig:dynamic:arti_a}-\ref{fig:dynamic:arti_d}, none of existing reward models produce a meaningful increase in ArtifactReward scores during training, reinforcing that current rewards do not effectively suppress artifact generation.

\subsection{Integrating ArtifactReward into RL}

We incorporate ArtifactReward into the four prior RL settings (HPS, GDino, ORM, and HPS + GDino) and the T2I-R1~\cite{jiang2025t2i} setting.
As illustrated in Table~\ref{tab:image}, incorporating ArtifactReward consistently \textbf{reduces the prevalence of visual artifacts} in generated images. The resulting images exhibit reduced artifacts, \textbf{improved realism} and more \textbf{balanced color composition}.

\subsection{Benchmark Evaluation}
To assess the generalizability of ArtifactReward, we evaluate all training variants on two recent benchmarks:
\begin{itemize}
    \item \textbf{WISE}~\cite{niu2025wise}: uses GPT-4o~\cite{hurst2024gpt} to evaluate consistency, realism, and aesthetic quality of generated images;
    \item \textbf{LLM4LLM}~\cite{wang2025lmm4lmm}: uses two fine-tuned LLM judges to compute Perception (clarity, authenticity, aesthetics) and Correspondence (text-image alignment).
\end{itemize}

\begin{table}
    \small
  \caption{\small Performance on WISE~\cite{niu2025wise} benchmark trained on Janus-Pro-1B~\cite{chen2025janus}. WiScore = 0.7 $\times$ Consistency + 0.2 $\times$ Realism + 0.1 $\times$ Aesthetic by the default setting.}
  \label{tab:wise}
  \centering
  \resizebox{1.0\linewidth}{!}
  {
  \begin{tabular}{l|ccc|c}
    \toprule
    \textbf{Method}  &Consistency &Realism &Aesthetic &\textbf{WiScore}\\
    \midrule
    Janus-Pro-1B~\cite{chen2025janus}  &0.2420 &0.3550 &0.3790 &0.2783 \\
    \midrule
    HPS~\cite{wu2023human}  &0.2960 &0.4515 &0.6115 &0.3587 \\
    HPS + ArtifactR  &0.3000 &0.5760 &0.5470 &0.3799 \\
    \midrule
    GDino~\cite{liu2024grounding}  &0.2905 &0.4100 &0.4395 &0.3293 \\
    GDino + ArtifactR  &0.3150 &0.5550 &0.5065 &0.3821 \\
    \midrule
    ORM~\cite{guo2025can}  &0.3210 &0.4750 &0.5060 &0.3703 \\
    ORM + ArtifactR  &0.3335 &0.5600 &0.5095 &0.3964 \\
    \midrule
    HPS + GDino  &0.3185 &0.4805 &\textbf{0.5960} &0.3787 \\
    HPS + GDino + ArtifactR  &0.3335 &\textbf{0.5765} &0.5210 &0.4009 \\
    \midrule
    T2I-R1~\cite{jiang2025t2i}  &0.3370 &0.4945 &0.5675 &0.3916 \\
    T2I-R1 + ArtifactR  &\textbf{0.3435} &0.5580 &0.5195 &\textbf{0.4040} \\
    \bottomrule
  \end{tabular}
  }
\end{table}


Table~\ref{tab:wise} (and Supplementary Table~\ref{tab:supp_7b_wise}) show the results on WISE benchmark. Across all reward configurations, integrating ArtifactReward leads to consistent performance improvements, particularly in realism and consistency. The primary gain originates from ArtifactReward’s explicit suppression of structural distortions and over-saturated appearances. Consequently, images appear more natural and coherent, despite sometimes at a modest cost to aesthetic scores, which is an expected tradeoff given the reduced emphasis on exaggerated colorfulness.

\begin{table}
    \small
  \caption{\small Performance on LLM4LLM~\cite{wang2025lmm4lmm} benchmark trained on Janus-Pro-1B~\cite{chen2025janus}.}
  \label{tab:llm4llm}
  \centering
  \resizebox{1.0\linewidth}{!}
  {
  \begin{tabular}{l|cc|c}
    \toprule
    \textbf{Method}  &Perception &Correspondence &All \\
    \midrule
    Janus-Pro-1B~\cite{chen2025janus}  &0.4056 &0.5058  &0.9114 \\
    \midrule
    HPS~\cite{wu2023human}  &0.4240 &0.5067 &0.9307 \\
    HPS + ArtifactR  &0.4464 &0.5089 &0.9553  \\
    \midrule
    GDino~\cite{liu2024grounding}  &0.4230 &0.5371 &0.9601  \\
    GDino + ArtifactR  &0.4336 &0.5416 &0.9751 \\
    \midrule
    ORM~\cite{guo2025can}  &0.4347 &0.5418 &0.9765 \\
    ORM + ArtifactR  &0.4479 &0.5497 &\textbf{0.9976} \\
    \midrule
    HPS + GDino  &0.4482 &0.5368 &0.9850 \\
    HPS + GDino + ArtifactR  &\textbf{0.4494} &0.5441 &0.9934 \\
    \midrule
    T2I-R1~\cite{jiang2025t2i}  &0.4448 &0.5390 &0.9838 \\
    T2I-R1 + ArtifactR  &0.4463 &\textbf{0.5458} &0.9921 \\
    \bottomrule
  \end{tabular}
  }
\end{table}

Table~\ref{tab:llm4llm} (and Supplementary Table~\ref{tab:supp_7b_llm4llm}) show the results on LLM4LLM benchmark. Incorporating ArtifactReward yields consistent gains across most training configurations. On the Perception dimension, ArtifactReward almost improves over all baselines, indicating that reducing structural artifacts directly enhances the perceived authenticity of generated images. Similarly, the Correspondence dimension also benefits from ArtifactReward, often achieving the best or near-best scores under each reward setting.

\begin{figure*}[htbp]
    \centering
    
    \begin{subfigure}[b]{0.21\textwidth}
        \centering
        \includegraphics[width=\textwidth]{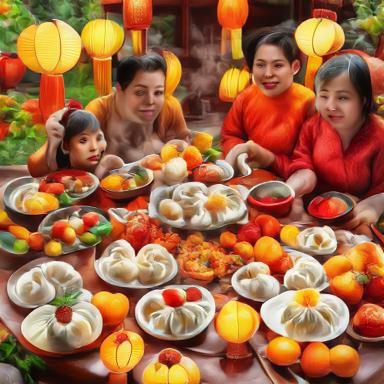}
        \caption{HPS}
        \label{fig:wise:hps}
    \end{subfigure}
    \hfill
    \begin{subfigure}[b]{0.21\textwidth}
        \centering
        \includegraphics[width=\textwidth]{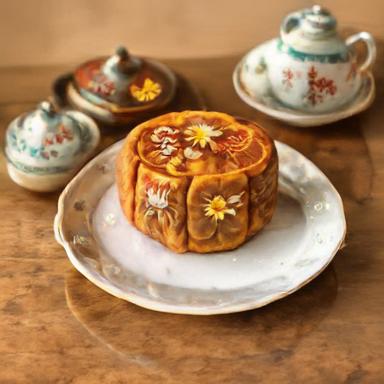}
        \caption{HPS + ArtifactR}
        \label{fig:wise:hps_arti}
    \end{subfigure}
    \hfill
    \begin{subfigure}[b]{0.21\textwidth}
        \centering
        \includegraphics[width=\textwidth]{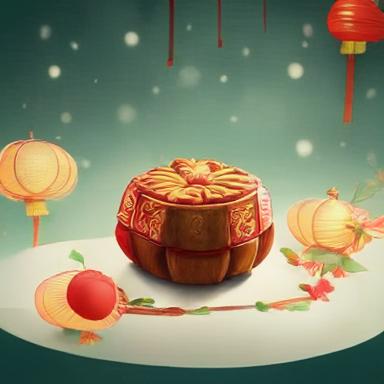}
        \caption{GDino}
        \label{fig:wise:gdino}
    \end{subfigure}
    \hfill
    \begin{subfigure}[b]{0.21\textwidth}
        \centering
        \includegraphics[width=\textwidth]{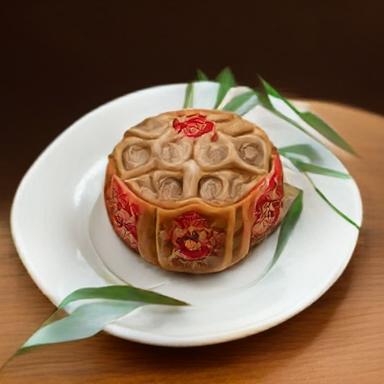}
        \caption{GDino + ArtifactR}
        \label{fig:wise:gdino_arti}
    \end{subfigure}


    \vspace{0.3cm} 

    \begin{subfigure}[b]{0.21\textwidth}
        \centering
        \includegraphics[width=\textwidth]{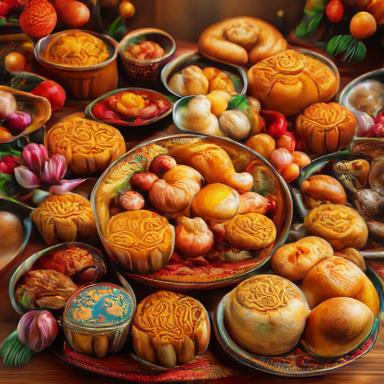}
        \caption{HPS + GDino}
        \label{fig:wise:hps_gdino}
    \end{subfigure}
    \hfill
    \begin{subfigure}[b]{0.21\textwidth}
        \centering
        \includegraphics[width=\textwidth]{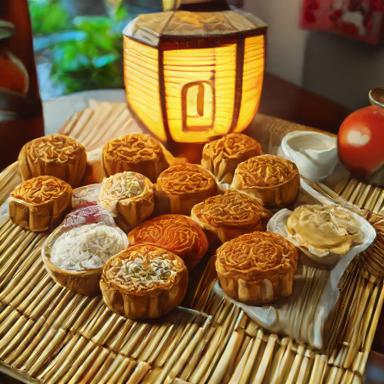}
        \caption{HPS + GDino + ArtifactR}
        \label{fig:wise:hps_gdino_arti}
    \end{subfigure}
    \hfill
    \begin{subfigure}[b]{0.21\textwidth}
        \centering
        \includegraphics[width=\textwidth]{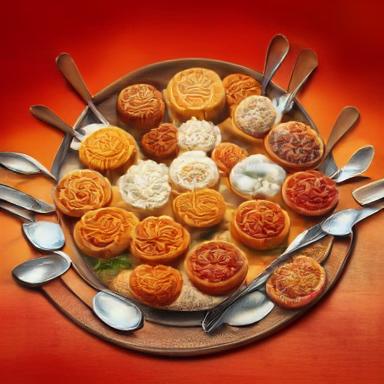}
        \caption{ORM}
        \label{fig:wise:orm}
    \end{subfigure}
    \hfill
    \begin{subfigure}[b]{0.21\textwidth}
        \centering
        \includegraphics[width=\textwidth]{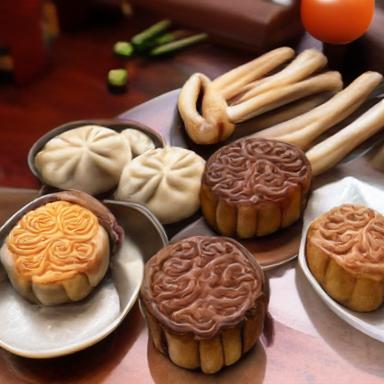}
        \caption{ORM + ArtifactR}
        \label{fig:wise:orm_arti}
    \end{subfigure}

    \caption{\small 
    Images generated with prompt ``\textit{Traditional food of the Mid-Autumn Festival}'' in WISE~\cite{niu2025wise} benchmark under different training reward configurations. This prompt expect the image to show mooncakes.
    }
    \label{fig:wise}

\end{figure*}
\begin{figure*}[htbp]
    \centering
    
    \begin{subfigure}[b]{0.21\textwidth}
        \centering
        \includegraphics[width=\textwidth]{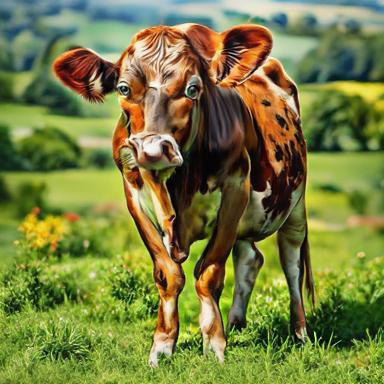}
        \caption{HPS}
        \label{fig:llm4llm:hps}
    \end{subfigure}
    \hfill
    \begin{subfigure}[b]{0.21\textwidth}
        \centering
        \includegraphics[width=\textwidth]{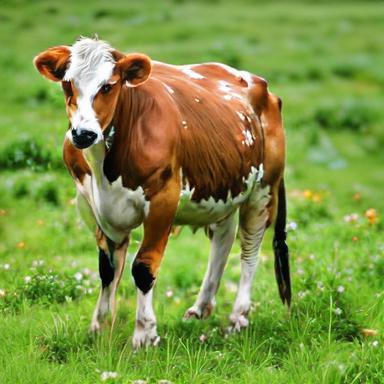}
        \caption{HPS + ArtifactR}
        \label{fig:llm4llm:hps_arti}
    \end{subfigure}
    \hfill
    \begin{subfigure}[b]{0.21\textwidth}
        \centering
        \includegraphics[width=\textwidth]{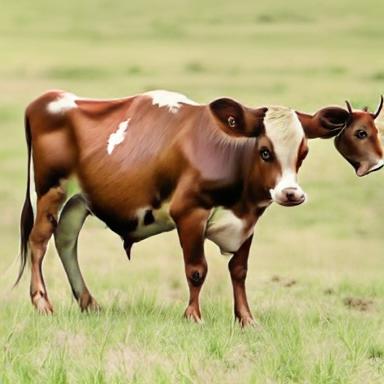}
        \caption{GDino}
        \label{fig:llm4llm:gdino}
    \end{subfigure}
    \hfill
    \begin{subfigure}[b]{0.21\textwidth}
        \centering
        \includegraphics[width=\textwidth]{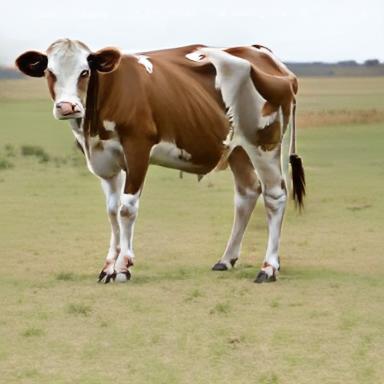}
        \caption{GDino + ArtifactR}
        \label{fig:llm4llm:gdino_arti}
    \end{subfigure}


    \vspace{0.3cm} 

    \begin{subfigure}[b]{0.21\textwidth}
        \centering
        \includegraphics[width=\textwidth]{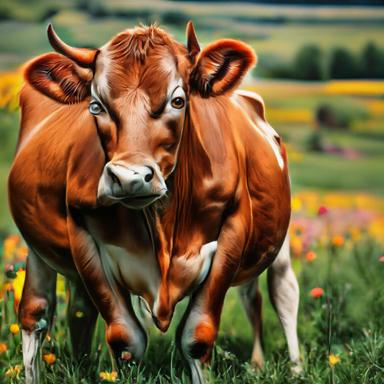}
        \caption{HPS + GDino}
        \label{fig:llm4llm:hps_gdino}
    \end{subfigure}
    \hfill
    \begin{subfigure}[b]{0.21\textwidth}
        \centering
        \includegraphics[width=\textwidth]{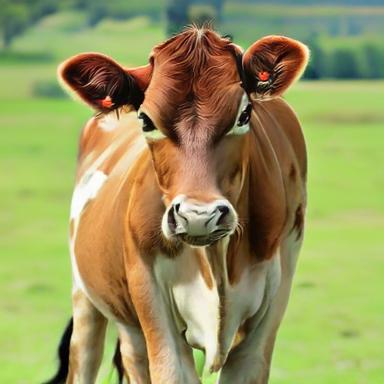}
        \caption{HPS + GDino + ArtifactR}
        \label{fig:llm4llm:hps_gdino_arti}
    \end{subfigure}
    \hfill
    \begin{subfigure}[b]{0.21\textwidth}
        \centering
        \includegraphics[width=\textwidth]{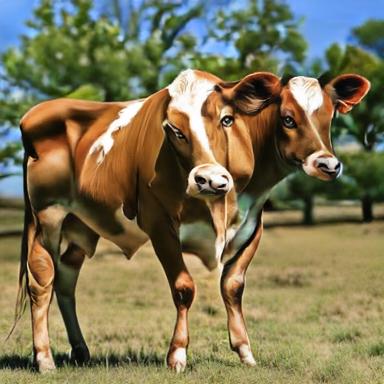}
        \caption{ORM}
        \label{fig:llm4llm:orm}
    \end{subfigure}
    \hfill
    \begin{subfigure}[b]{0.21\textwidth}
        \centering
        \includegraphics[width=\textwidth]{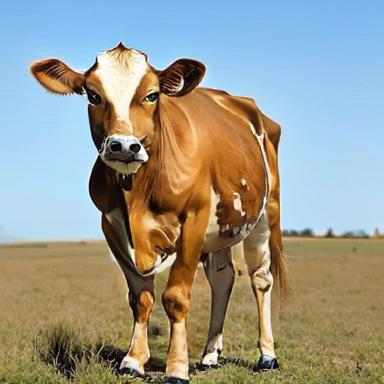}
        \caption{ORM + ArtifactR}
        \label{fig:llm4llm:orm_arti}
    \end{subfigure}

    \caption{\small 
    Images generated with prompt ``\textit{a photo of a cow}'' in LLM4LLM~\cite{wang2025lmm4lmm} under different training reward configurations.
    }
    \label{fig:llm4llm}

\end{figure*}

Figure~\ref{fig:wise} and Figure~\ref{fig:llm4llm} illustrate qualitative examples from these two benchmarks. In Figure~\ref{fig:wise}, images generated with the help of our ArtifactReward exhibit noticeably improved realism and better alignment with the prompt intent, while also reducing the excessive color saturation often produced by HPS-trained models. In Figure~\ref{fig:llm4llm}, we also observe that HPS-based training tends to introduce unnatural colors, whereas GDino and ORM-based training frequently results in duplicated or geometrically distorted objects. In contrast, incorporating our ArtifactReward effectively suppresses these issues, yielding images with more coherent structure and plausible visual attributes.

\vspace{-8pt}
\paragraph{Summary.}
Together, these results demonstrate that ArtifactReward generalizes well across diverse reward settings and architectures, consistently improving both perceptual quality and semantic correspondence. By explicitly targeting perceptual and structural plausibility, failure modes that existing rewards systematically overlook, ArtifactReward acts as an effective and general-purpose regularizer against T2I reward hacking. It complements existing rewards that emphasize aesthetics or compositional alignment. Although aesthetic scores may slightly decrease due to reduced over-saturation, the substantial improvements in realism, alignment consistency, and cross-benchmark robustness highlight ArtifactReward’s value as a meaningful step toward more human-aligned T2I reward design.

\section{Conclusion}
\label{sec:conclusion}

In this work, we present the first systematic study of reward hacking in Text-to-Image RL post-training. Through extensive analysis across widely used reward models, which spans aesthetics, human preference, and image-text alignment, we reveal that existing rewards not only induce reward-specific biases but also share a universal blind spot: the persistent production of structural artifacts that current reward functions fail to penalize. To address this gap, we introduce ArtifactReward, a complementary reward learned via automatic prompt optimization to detect fine-grained visual artifacts from limited supervision. When integrated into multiple RL training configurations, ArtifactReward consistently reduces artifact prevalence, improves realism, and enhances benchmark performance.

\section*{Acknowledgments}
This work is supported by NSF 2048280, 2325121, 2244760, 2331966 and ONR N00014-23-1-2300:P00001
{
    \small
    \bibliographystyle{ieeenat_fullname}
    \bibliography{main}
}

\clearpage
\setcounter{page}{1}
\maketitlesupplementary

\section{Automatic Prompt Optimization for ArtifactReward}
\label{supp:apo}
We employ automatic prompt optimization (APO)~\cite{pryzant2023automatic} to optimize the prompt used for ArtifactReward. Qwen2.5-VL-7B-Instruct~\cite{bai2025qwen2} serves as our backbone vision-language model, and APO is used to optimize the prompt so that the model can more reliably distinguish artifact-containing images from artifact-free ones.

Our initial prompt is given as:
\begin{tcolorbox}[title={Initial prompt.},
boxrule=2pt, arc=0mm, breakable, colback=white, coltitle=white,
top=-2pt, bottom=-2pt]
\begin{lstlisting}[language=markdown]
Is there any artifacts in the image that look not realistic?
\end{lstlisting}
\end{tcolorbox}

We compute the ArtifactReward using the normalized probability of the model answering ``no'', where a higher ``no'' probability indicates a higher likelihood of being artifact-free (See Eq.~\ref{artifact}).
During optimization, we maximize the following log-likelihood objective:
\begin{equation}
\label{eq:apo_goal}
    LL = y \text{log}(p_{\text{yes}}) + (1-y) \text{log}(1-p_{\text{yes}}),
\end{equation}
where $y$ is the ground-truth label, and $p_{\text{yes}}=1-p_{\text{no}}$ denotes the normalized probability assigned to the ``yes'' token.
Algorithm~\ref{alg:apo} provides the complete APO procedure, with the scoring function $S(\cdot)$ defined as the log-likelihood above.

\begin{algorithm}
\small
\caption{Automatic Prompt Optimization for ArtifactReward}  
\label{alg:apo}
\begin{algorithmic}[1]
\Require {$p_0$: initial prompt, $N$: iterations, $\mathcal{D}={(x,y)}$: training dataset where $x$ is the input image and $y$ is the artifact label, $b$: top prompts retained per iteration, $l$: number of error examples for reflections, $S(\cdot)$: scoring function}
\State $P_0\leftarrow \{p_0\}$ \Comment{Initialize candidate prompt set}
\For{$t=1$ to $N$}
    \State $P_c\leftarrow P_{t-1}$
    \For{$p \in P_{t-1}$}
        \State $J_{\mathrm{error}}=\{(x_i,y_i) \mid p^{i}_{\text{yes}}>0.5 \hspace{0.5em} \text{if} \hspace{0.5em} y_i=0, \hspace{0.5em} \text{or} \hspace{0.5em} p^{i}_{\text{yes}}<0.5 \hspace{0.5em} \text{if} \hspace{0.5em} y_i=1$\}
        \State $J_{\mathrm{error}}^i \subset J_{\mathrm{error}}$ is a sampled subset for each $i = 1,\dots,l$
        \State $G=\bigcup_{i=1,...l} \mathrm{Reflect}(p,J_{\mathrm{error}}^i)$
        \State $H=\bigcup_{i=1,...l} \mathrm{Modify}(p,g_i,J_{\mathrm{error}}^i)$
        \State $P_c\leftarrow P_c\cup H$
    \EndFor
    \State $\mathcal{S}_c=\{S(p) \mid p\in P_c\}$ \Comment{Evaluate prompts}
    \State $P_t\leftarrow \{p\in P_c \mid S(p)\geq \tau\}$, where $\tau$ is the $b^\text{th}$ highest score in $\mathcal{S}_c$
\EndFor
\State \textbf{Return} $p^*\leftarrow \arg\max_{p\in P_N}S(p)$
\end{algorithmic}
\end{algorithm}

An example of the optimized prompt is:
\begin{tcolorbox}[title={Optimized prompt.},
boxrule=2pt, arc=0mm, breakable, colback=white, coltitle=white,
top=-2pt, bottom=-2pt]
\begin{lstlisting}[language=markdown]
Analyze the images to identify any unintentional digital artifacts, concentrating on irregular lighting, object placement, blending errors, or anomalies that could affect realism. Disregard any deliberate artistic styles or intentional surreal elements. Respond with YES if artifacts are detected; if not, respond with NO.
\end{lstlisting}
\end{tcolorbox}
This optimized prompt is more explicit and structurally aligned with common artifact patterns, enabling the model to produce more accurate and reliable artifact predictions.
As shown in Table~\ref{tab:artifact_img}, our ArtifactReward consistently assigns higher scores to artifact-free images, clearly separating clean generations from those with distortions. This demonstrates the effectiveness of our optimized prompt in enabling a reward model that can reliably detect and penalize generated artifacts.

\section{Additional Experiment Results}

\subsection{Additional Training Dynamics on Janus-Pro-7B}
\begin{figure*}[htbp]
    \centering
    \begin{subfigure}[b]{0.24\textwidth}
        \centering
        \includegraphics[width=\textwidth]{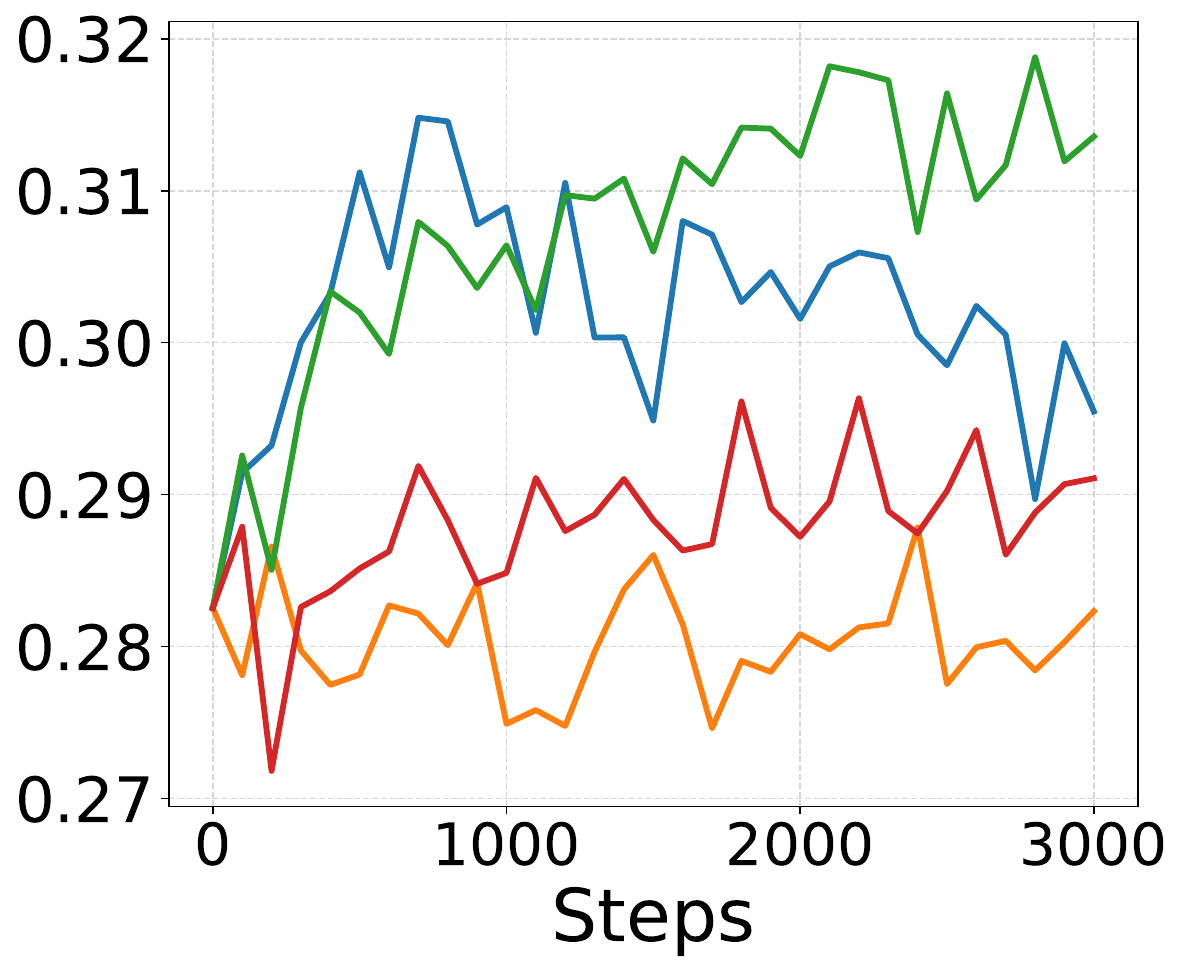}
        \caption{HPS - Spatial}
        \label{fig:dyn_supp:hps}
    \end{subfigure}
    \hfill
    \begin{subfigure}[b]{0.24\textwidth}
        \centering
        \includegraphics[width=\textwidth]{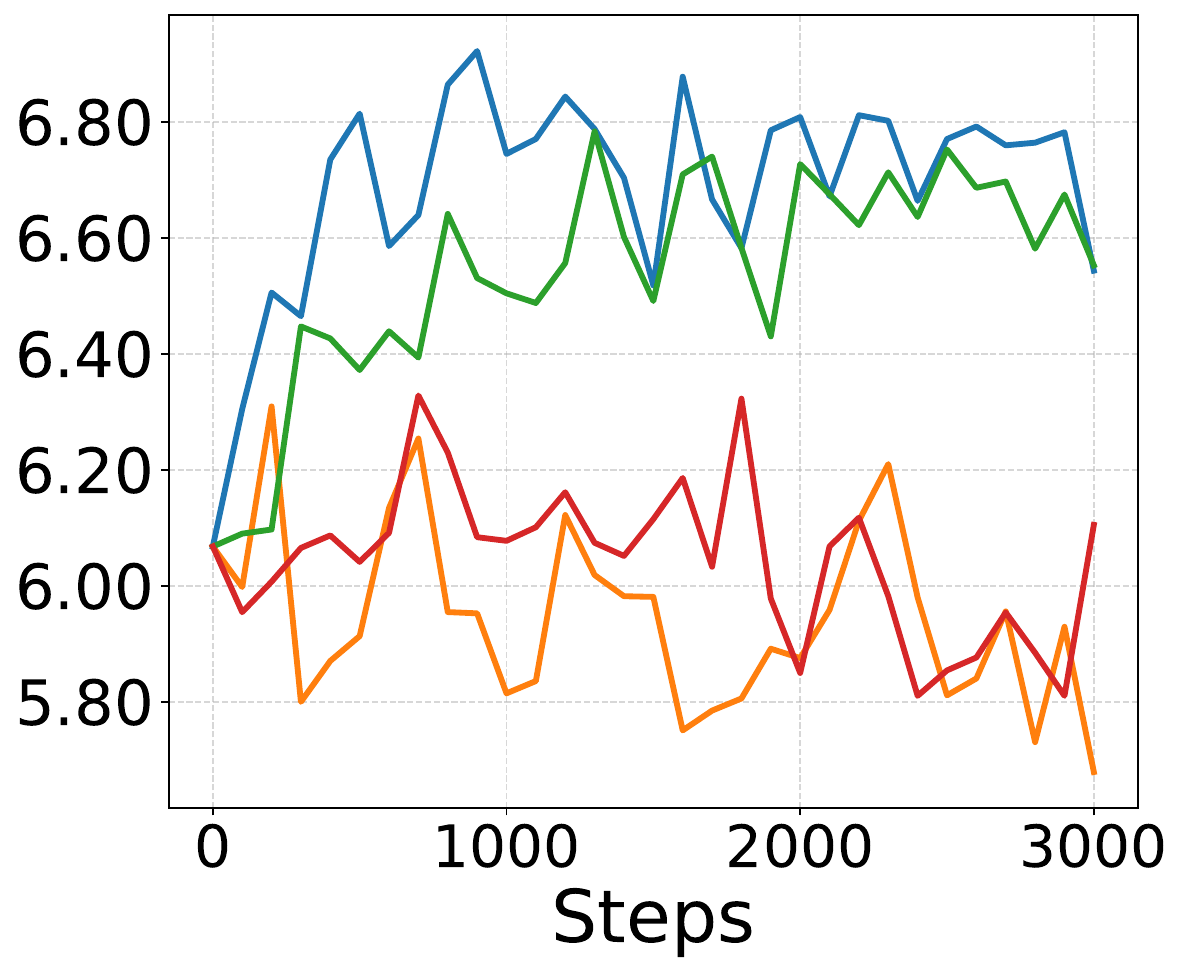}
        \caption{Aesthetic Score - Color}
        \label{fig:dyn_supp:aes}
    \end{subfigure}
    \hfill
    \begin{subfigure}[b]{0.24\textwidth}
        \centering
        \includegraphics[width=\textwidth]{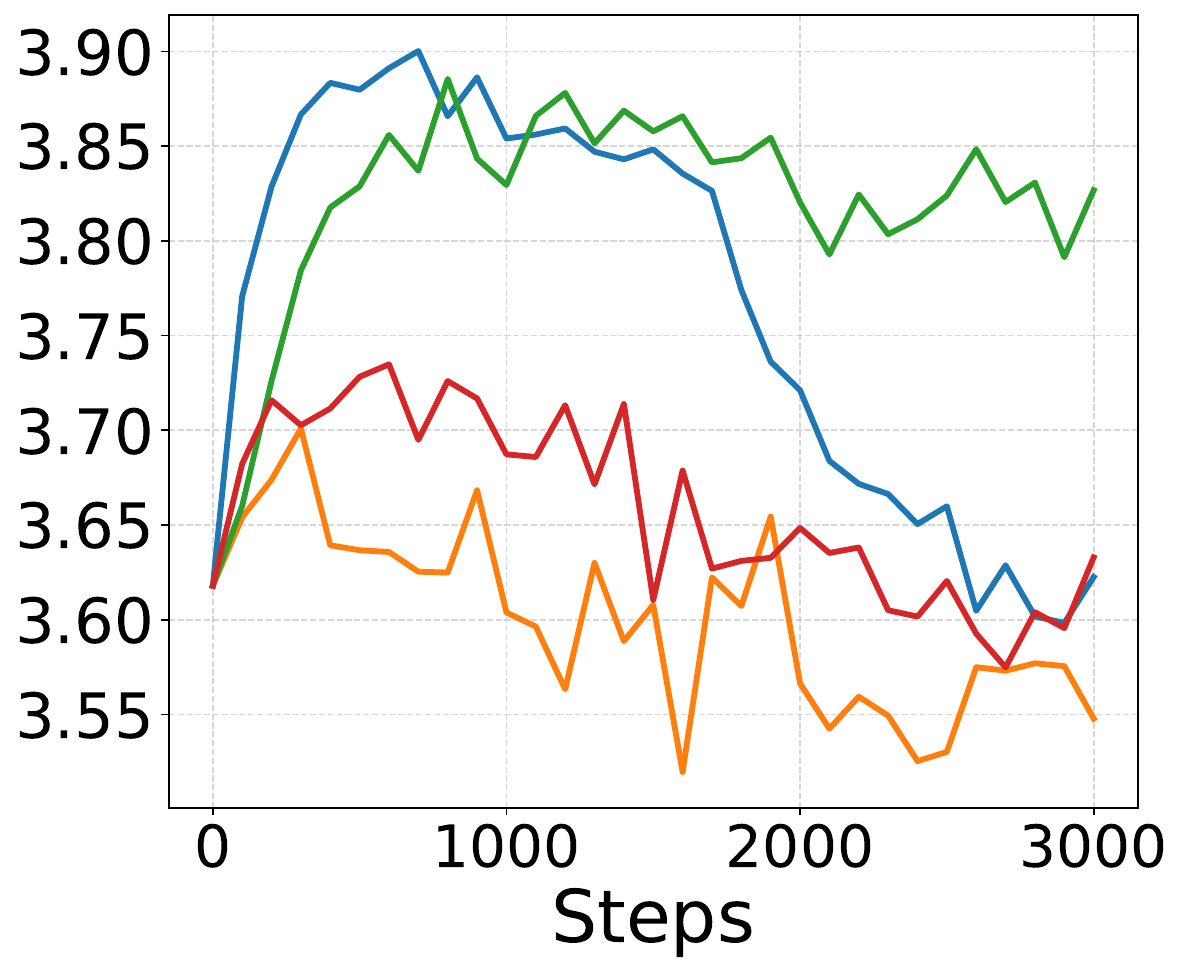}
        \caption{DeQA - Spatial}
        \label{fig:dyn_supp:deqa}
    \end{subfigure}
    \hfill
    \begin{subfigure}[b]{0.24\textwidth}
        \centering
        \includegraphics[width=\textwidth]{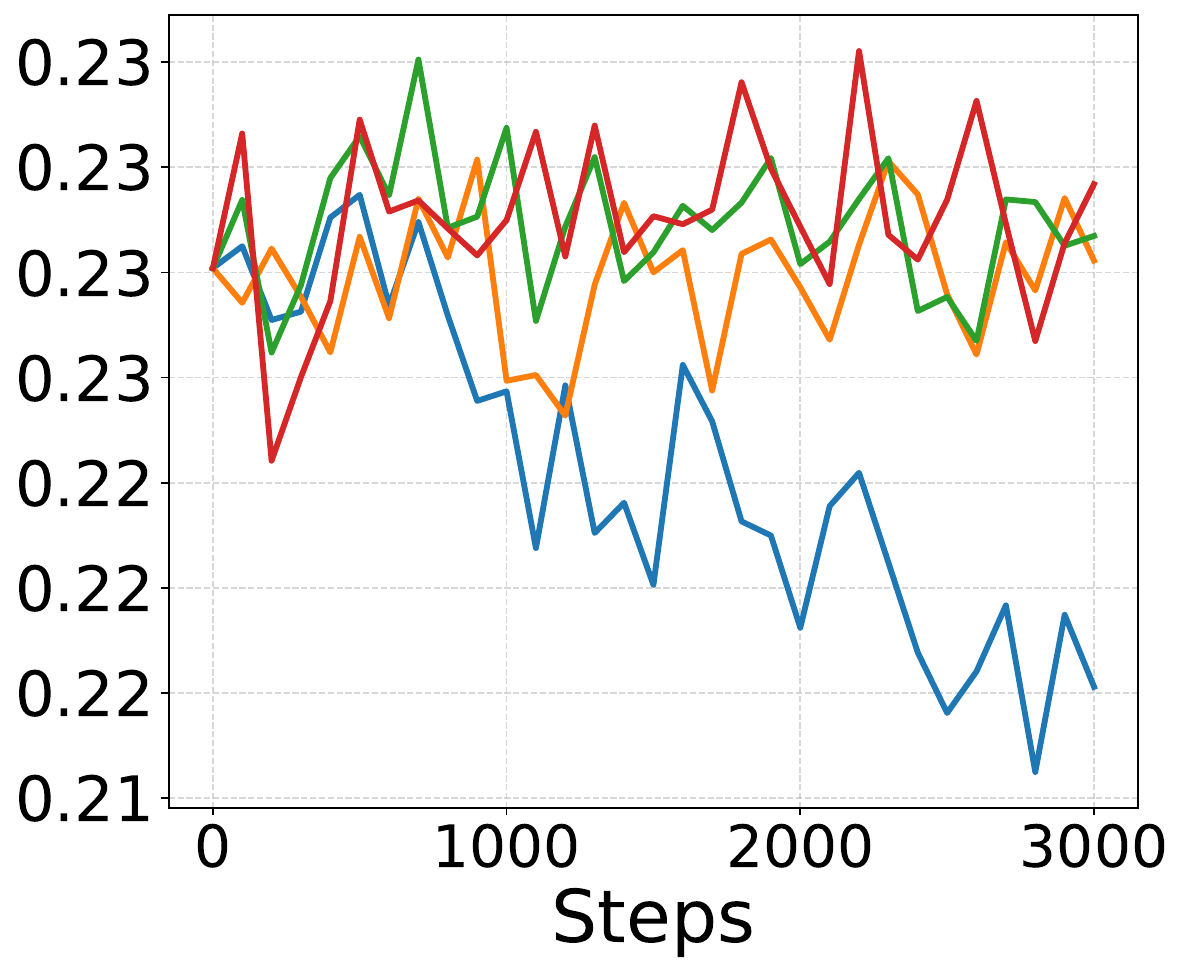}
        \caption{PickScore - Object}
        \label{fig:dyn_supp:pick}
    \end{subfigure}

    \vspace{0.3cm} 

    \centering
    \begin{subfigure}[b]{0.24\textwidth}
        \centering
        \includegraphics[width=\textwidth]{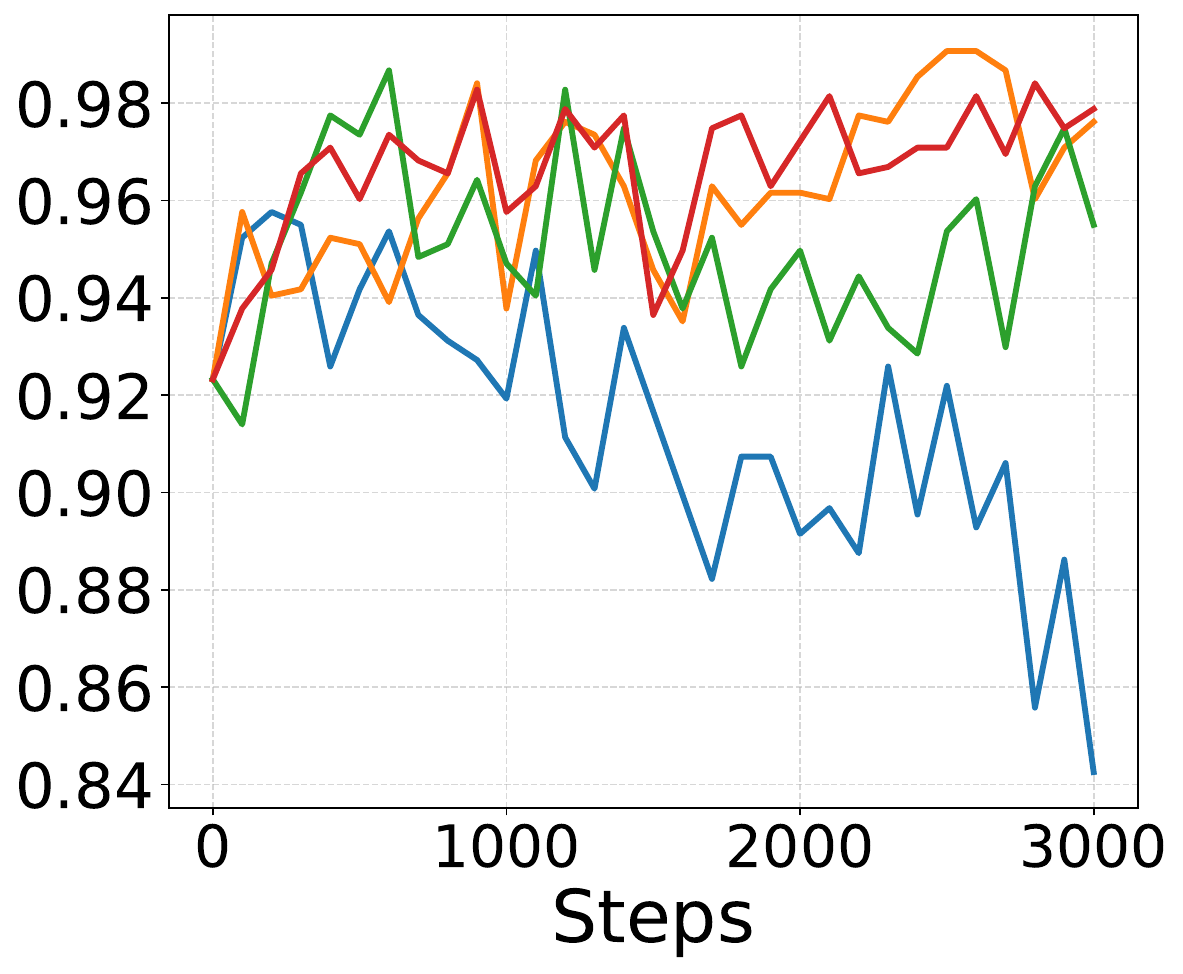}
        \caption{GDino - Texture}
        \label{fig:dyn_supp:gdino}
    \end{subfigure}
    \hfill
    \begin{subfigure}[b]{0.24\textwidth}
        \centering
        \includegraphics[width=\textwidth]{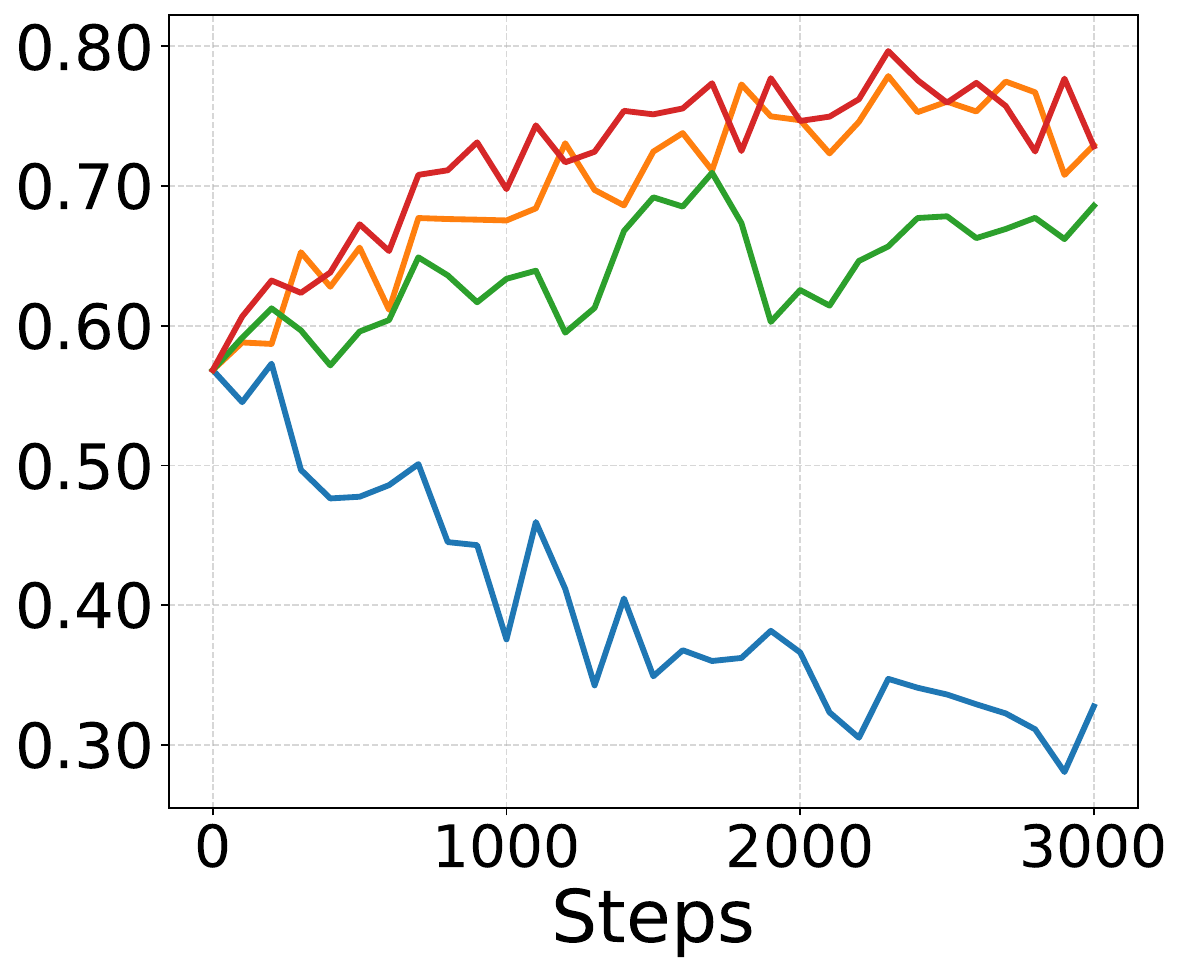}
        \caption{ORM - Spatial}
        \label{fig:dyn_supp:orm}
    \end{subfigure}
    \hfill
    \begin{subfigure}[b]{0.24\textwidth}
        \centering
        \includegraphics[width=\textwidth]{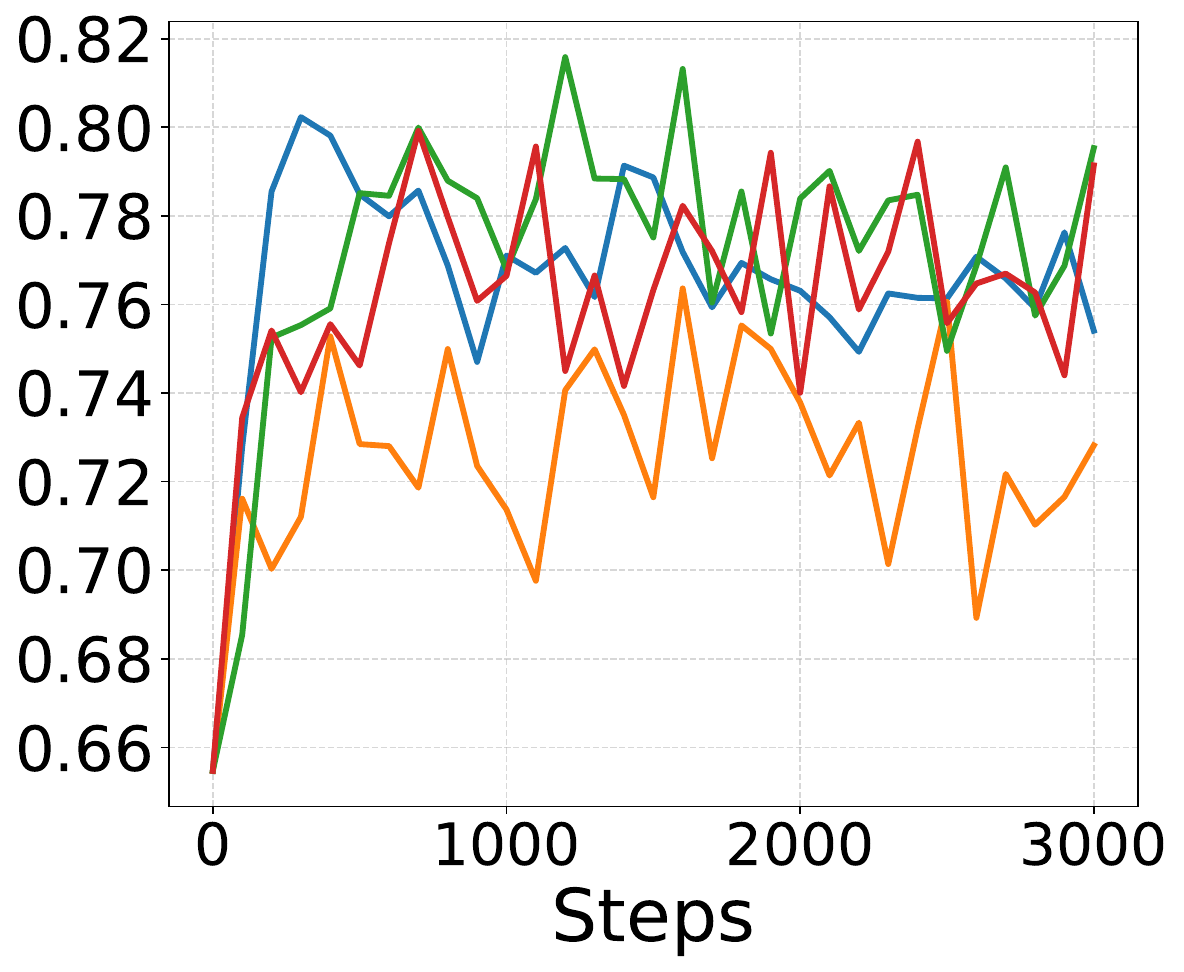}
        \caption{VQAScore - Numeracy}
        \label{fig:dyn_supp:vqa}
    \end{subfigure}
    \hfill
    \begin{subfigure}[b]{0.24\textwidth}
        \centering
        \includegraphics[width=\textwidth]{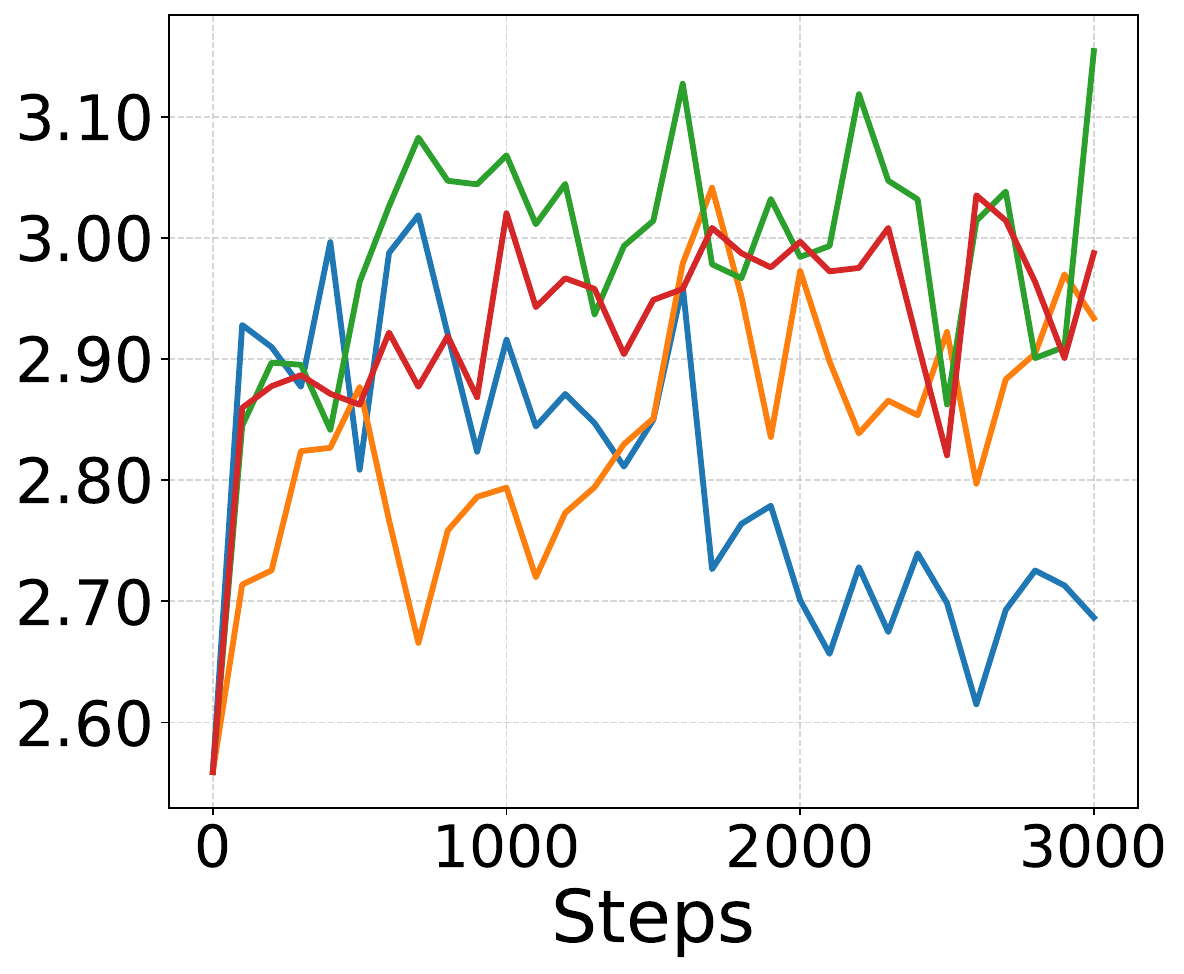}
        \caption{Unified - Numeracy}
        \label{fig:dyn_supp:uni}
    \end{subfigure}

    \vspace{0.3cm} 

    \begin{subfigure}[b]{0.24\textwidth}
        \centering
        \includegraphics[width=\textwidth]{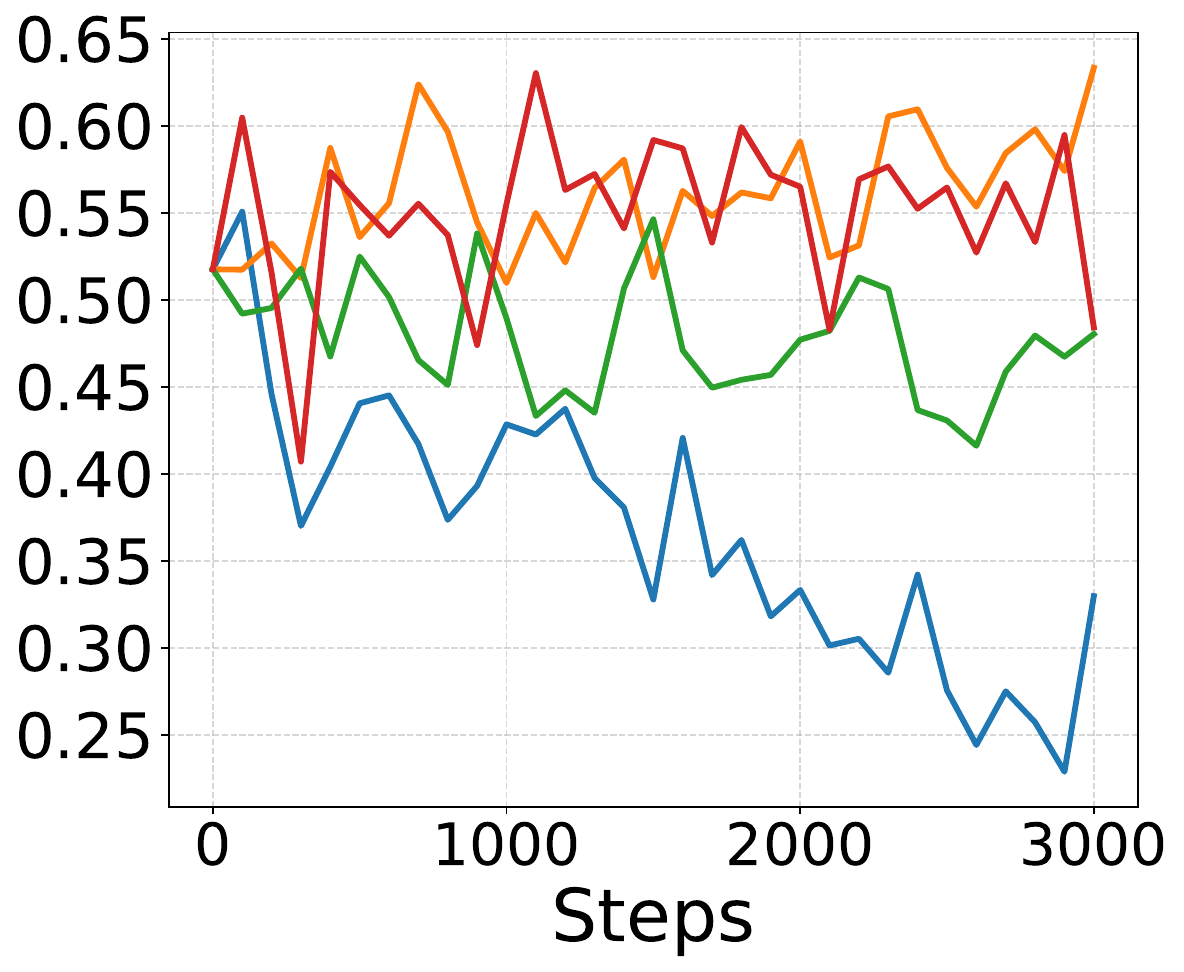}
        \caption{ArtifactR - Object}
        \label{fig:dyn_supp:arti_a}
    \end{subfigure}
    \hfill
    \begin{subfigure}[b]{0.24\textwidth}
        \centering
        \includegraphics[width=\textwidth]{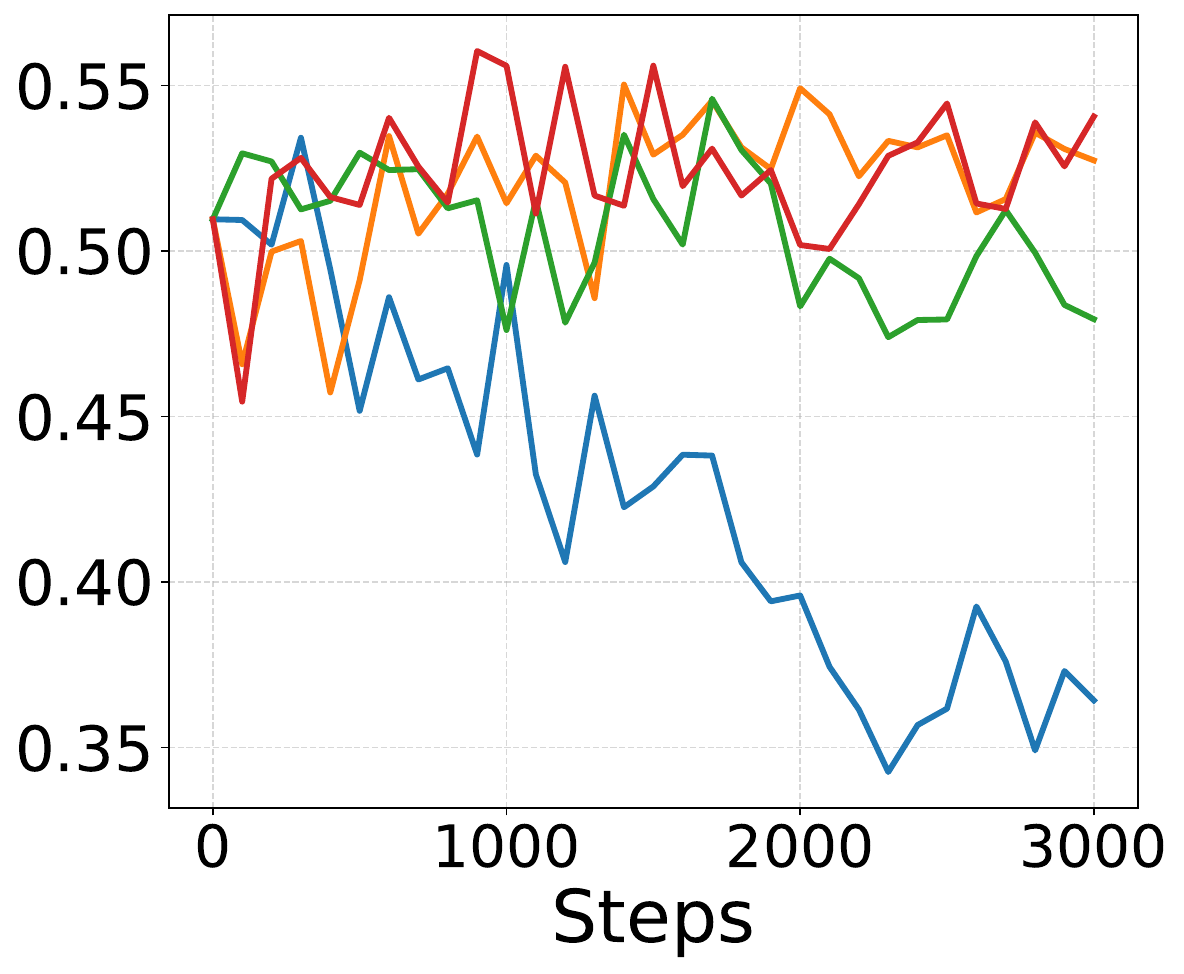}
        \caption{ArtifactR - Shape}
        \label{fig:dyn_supp:arti_b}
    \end{subfigure}
    \hfill
    \begin{subfigure}[b]{0.24\textwidth}
        \centering
        \includegraphics[width=\textwidth]{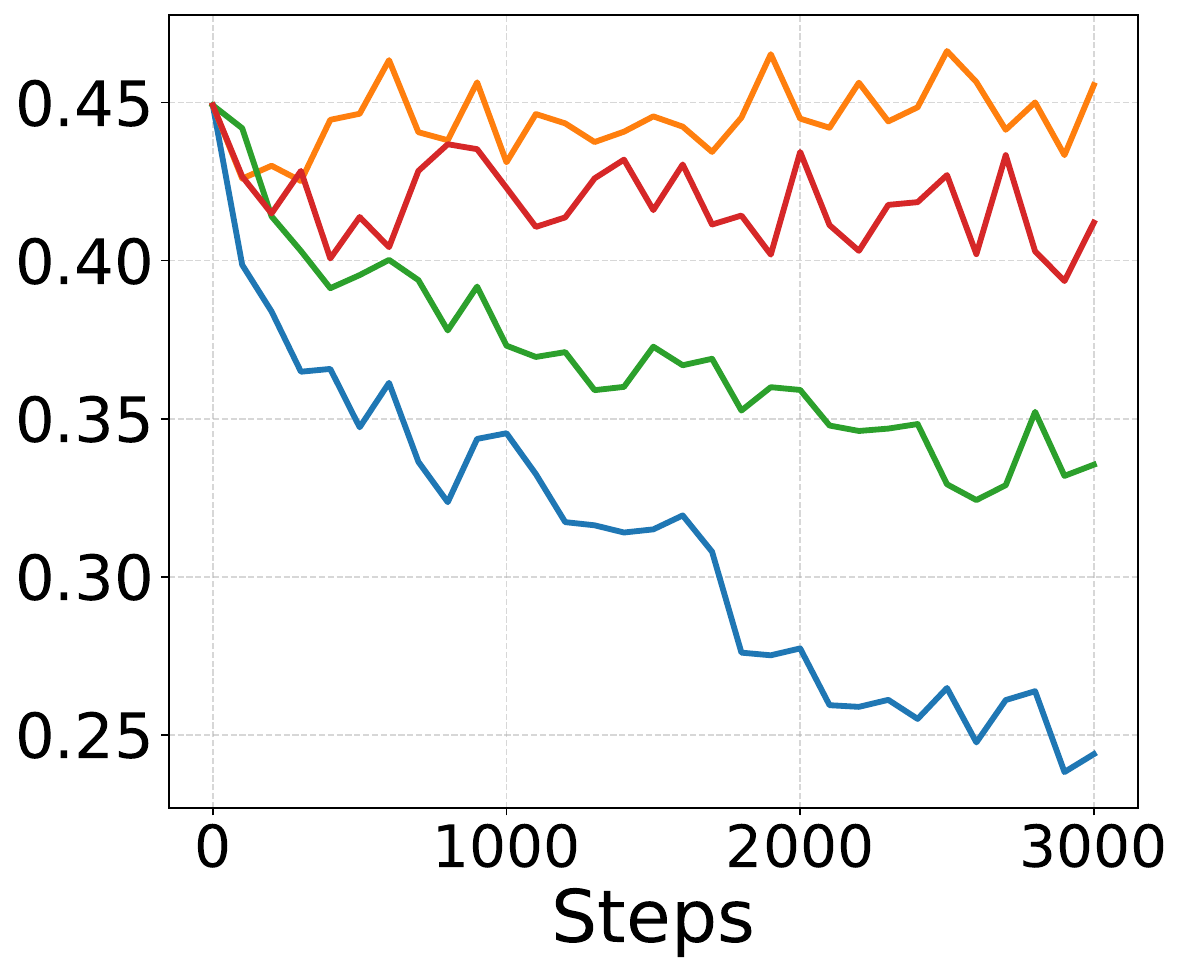}
        \caption{ArtifactR - Spatial}
        \label{fig:dyn_supp:arti_c}
    \end{subfigure}
    \hfill
    \begin{subfigure}[b]{0.24\textwidth}
        \centering
        \includegraphics[width=\textwidth]{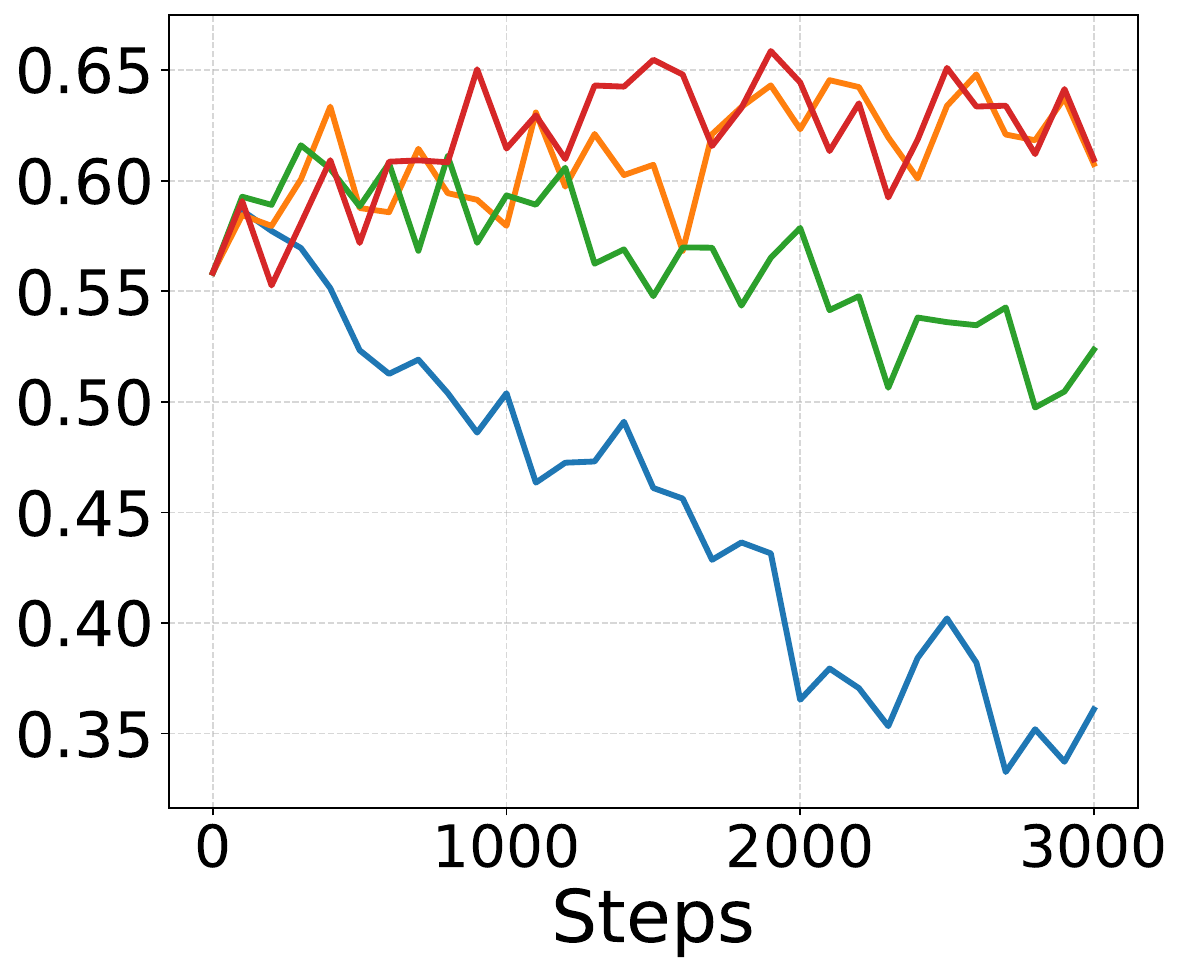}
        \caption{ArtifactR - Texture}
        \label{fig:dyn_supp:arti_d}
    \end{subfigure}

    \caption{\small 
    Evolution of metrics over training steps with different categories of prompts trained on Janus-Pro-7B~\cite{chen2025janus}. {\color{tabblue}Blue} color denotes the model trained with HPS~\cite{wu2023human}; {\color{taborange}Orange} color denotes the model trained with GDino~\cite{liu2024grounding}; {\color{tabgreen}Green} color denotes the model trained with HPS and GDino; {\color{tabred}Red} color denotes the model trained with finetuned ORM~\cite{guo2025can}.
    }
    \label{fig:dyn_supp}

\end{figure*}


We conducted additional experiments using the larger 7B variant of Janus-Pro~\cite{chen2025janus}. The training dynamics of different evaluation metrics are shown in Figure~\ref{fig:dyn_supp}. Overall, we observe trends similar to those seen with the 1B model. Optimization under a given reward tends to improve metrics with closely related inductive biases, while degrading performance on metrics that capture different aspects of image quality. For example, the model trained with the GDino reward shows increased GDino and ORM scores, as both scores emphasize object recognition in generated images. In contrast, the model fails to improve HPS, Aesthetic Score, and DeQA, which focus more on color quality and visual appeal.
Moreover, across all reward settings, we find that none effectively reduce artifacts in the generated images (Figures~\ref{fig:dyn_supp:arti_a}–\ref{fig:dyn_supp:arti_d}).

\subsection{Detailed Results of WISE benchmark}
\begin{table}
    \small
  \caption{\small Performance on WISE~\cite{niu2025wise} benchmark trained on Janus-Pro-7B~\cite{chen2025janus}. WiScore = 0.7 $\times$ Consistency + 0.2 $\times$ Realism + 0.1 $\times$ Aesthetic by the default setting.}
  \label{tab:supp_7b_wise}
  \centering
  \resizebox{1.0\linewidth}{!}
  {
  \begin{tabular}{l|ccc|c}
    \toprule
    \textbf{Method}  &Consistency &Realism &Aesthetic &\textbf{WiScore}\\
    \midrule
    Janus-Pro-7B~\cite{chen2025janus}  &0.4340 &0.5470 &0.5528 &0.4685 \\
    \midrule
    HPS~\cite{wu2023human}  &0.4160 &0.4745 &0.6530 &0.4514 \\
    HPS + ArtifactR  &0.4255 &\textbf{0.6605} &0.5815 &0.4881 \\
    \midrule
    GDino~\cite{liu2024grounding}  &0.4445 &0.5170 &0.5435 &0.4689 \\
    GDino + ArtifactR  &0.4470 &0.6295 &0.5465 &0.4935 \\
    \midrule
    ORM~\cite{guo2025can}  &0.4420 &0.5800 &0.5975 &0.4852 \\
    ORM + ArtifactR  &0.4665 &0.6355 &0.5840 &0.5121 \\
    \midrule
    HPS + GDino  &0.4780 &0.5697 &\textbf{0.6755} &0.5161 \\
    HPS + GDino + ArtifactR  &0.4775 &0.6470 &0.5880 &0.5225 \\
    \midrule
    T2I-R1~\cite{jiang2025t2i}  &0.4655 &0.5925 &0.6730 &0.5157 \\
    T2I-R1 + ArtifactR  &\textbf{0.4945} &0.6450 &0.6000 &\textbf{0.5352} \\
    \bottomrule
  \end{tabular}
  }
\end{table}

\begin{figure*}[htbp]
    \centering

    \begin{subfigure}[b]{1.0\textwidth}
        \centering
        \includegraphics[width=\textwidth]{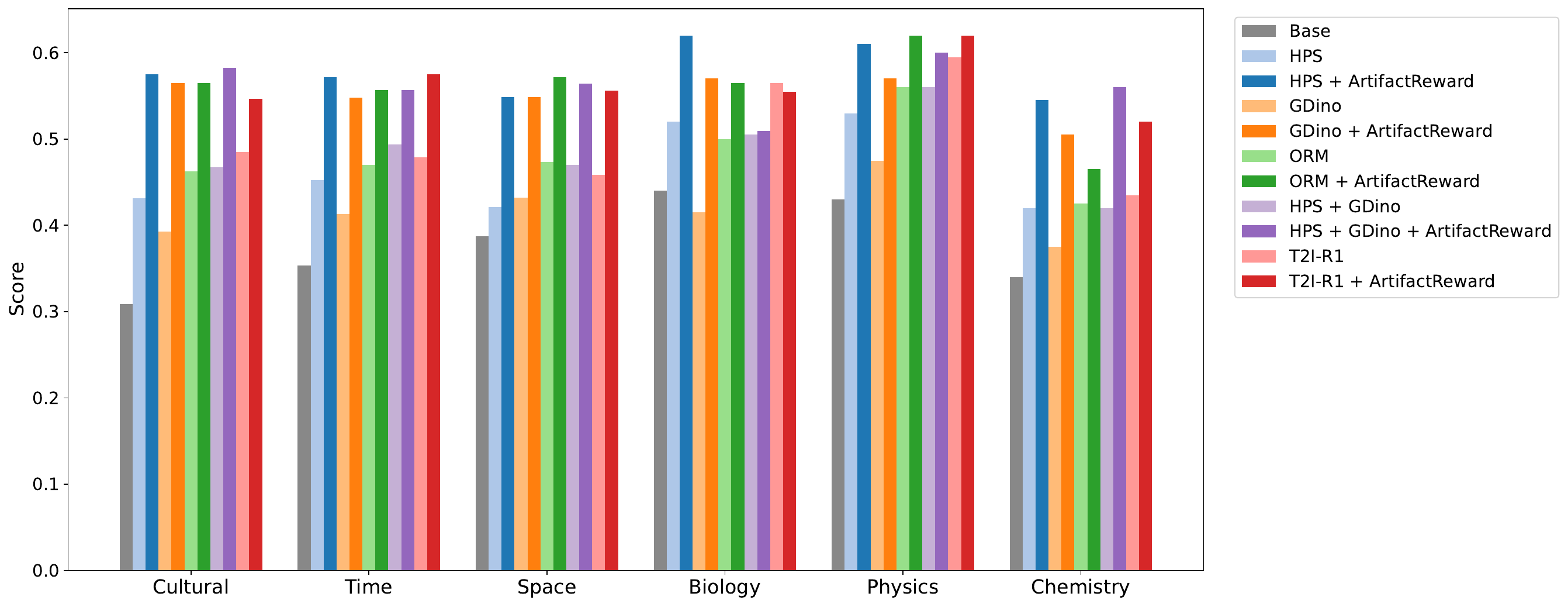}
        \caption{Realism}
        \label{fig:wise_1b_real}
    \end{subfigure}


    \vspace{0.3cm} 

    \begin{subfigure}[b]{1.0\textwidth}
        \centering
        \includegraphics[width=\textwidth]{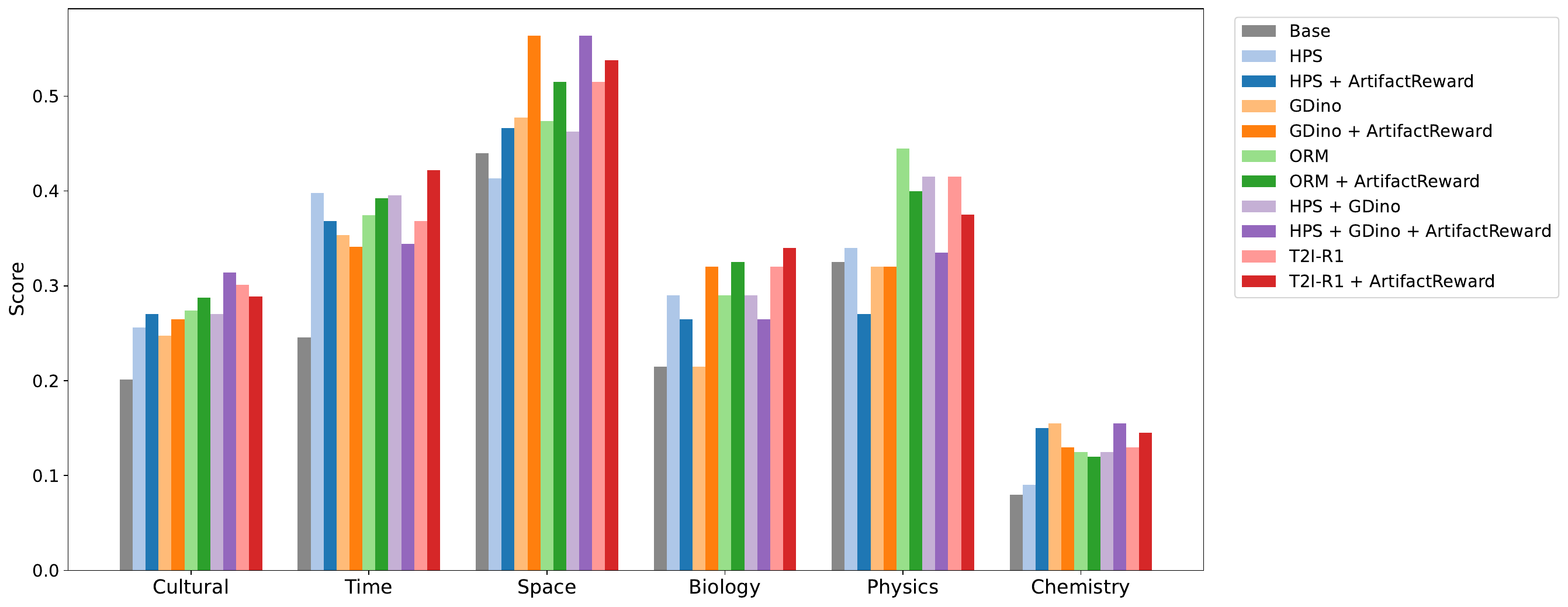}
        \caption{Consistency}
        \label{fig:wise_1b_consist}
    \end{subfigure}

    \vspace{0.3cm} 

    \begin{subfigure}[b]{1.0\textwidth}
        \centering
        \includegraphics[width=\textwidth]{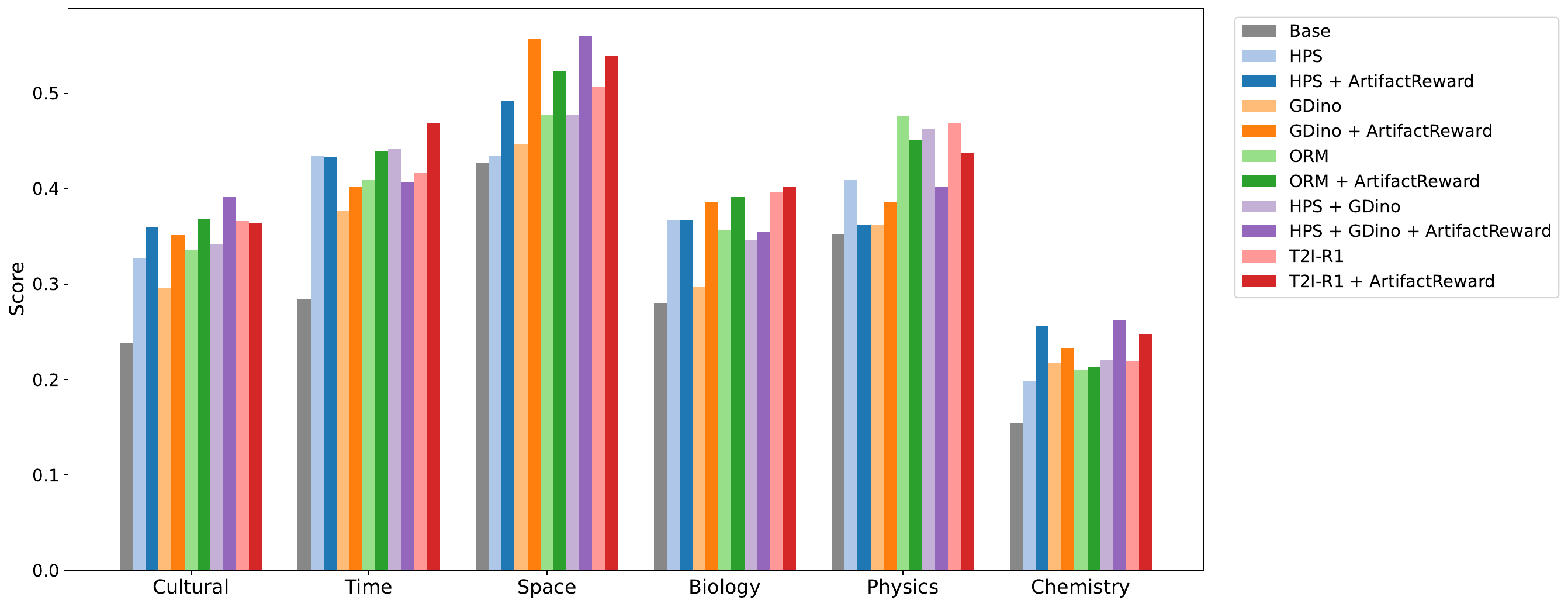}
        \caption{WiScore}
        \label{fig:wise_1b_wiscore}
    \end{subfigure}

    \caption{\small 
    Performance on WISE~\cite{niu2025wise} benchmark across different categories trained on Janus-Pro-1B~\cite{chen2025janus}.
    }
    \label{fig:wise_1b_res}

\end{figure*}

\begin{figure*}[htbp]
    \centering
    
    \begin{subfigure}[b]{0.22\textwidth}
        \centering
        \includegraphics[width=\textwidth]{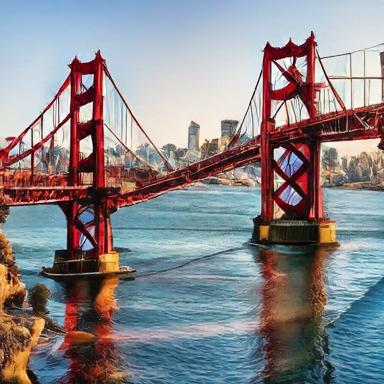}
        \caption{HPS}
        \label{fig:supp_wise:hps}
    \end{subfigure}
    \hfill
    \begin{subfigure}[b]{0.22\textwidth}
        \centering
        \includegraphics[width=\textwidth]{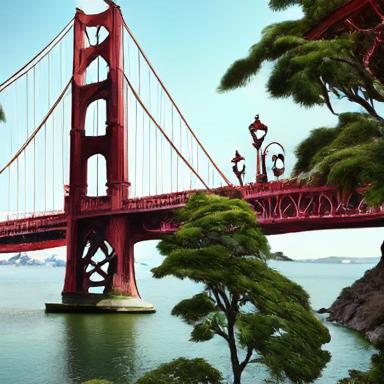}
        \caption{HPS + ArtifactR}
        \label{fig:supp_wise:hps_arti}
    \end{subfigure}
    \hfill
    \begin{subfigure}[b]{0.22\textwidth}
        \centering
        \includegraphics[width=\textwidth]{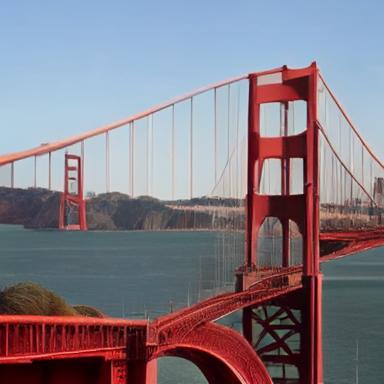}
        \caption{GDino}
        \label{fig:supp_wise:gdino}
    \end{subfigure}
    \hfill
    \begin{subfigure}[b]{0.22\textwidth}
        \centering
        \includegraphics[width=\textwidth]{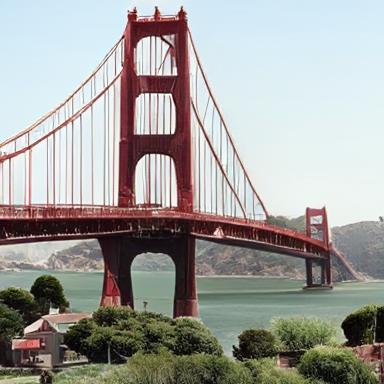}
        \caption{GDino + ArtifactR}
        \label{fig:supp_wise:gdino_arti}
    \end{subfigure}


    \vspace{0.3cm} 

    \begin{subfigure}[b]{0.22\textwidth}
        \centering
        \includegraphics[width=\textwidth]{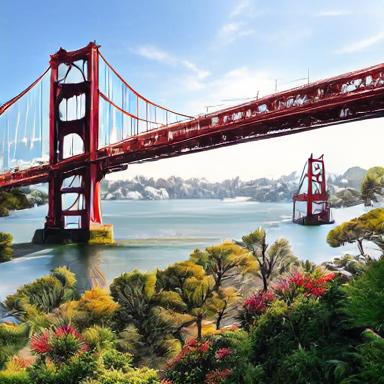}
        \caption{HPS + GDino}
        \label{fig:supp_wise:hps_gdino}
    \end{subfigure}
    \hfill
    \begin{subfigure}[b]{0.22\textwidth}
        \centering
        \includegraphics[width=\textwidth]{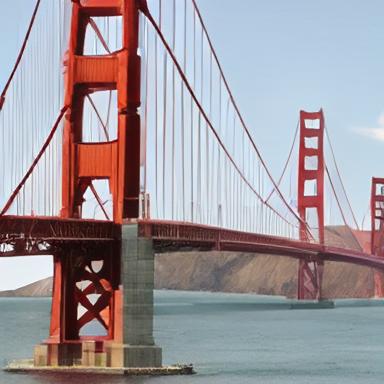}
        \caption{HPS + GDino + ArtifactR}
        \label{fig:supp_wise:hps_gdino_arti}
    \end{subfigure}
    \hfill
    \begin{subfigure}[b]{0.22\textwidth}
        \centering
        \includegraphics[width=\textwidth]{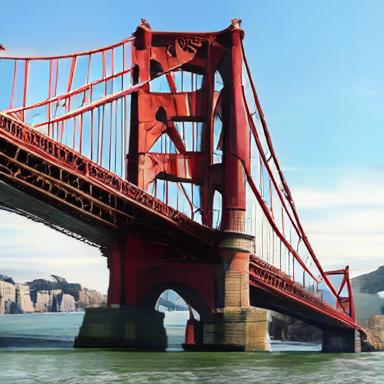}
        \caption{ORM}
        \label{fig:supp_wise:orm}
    \end{subfigure}
    \hfill
    \begin{subfigure}[b]{0.22\textwidth}
        \centering
        \includegraphics[width=\textwidth]{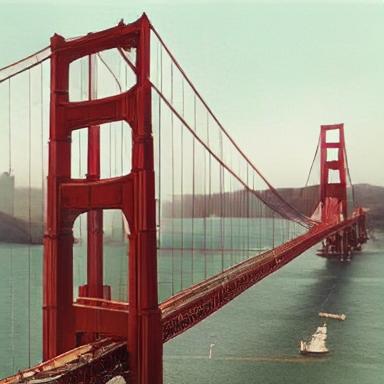}
        \caption{ORM + ArtifactR}
        \label{fig:supp_wise:orm_arti}
    \end{subfigure}

    \caption{\small 
    Images generated with prompt ``\textit{An iconic bridge, known for its red hue and location over a famous bay in San Francisco}'' in WISE~\cite{niu2025wise} benchmark under different training reward configurations trained on Janus-Pro-1B~\cite{chen2025janus}.
    }
    \label{fig:supp_wise}

\end{figure*}

\begin{figure*}[htbp]
    \centering
    
    \begin{subfigure}[b]{0.22\textwidth}
        \centering
        \includegraphics[width=\textwidth]{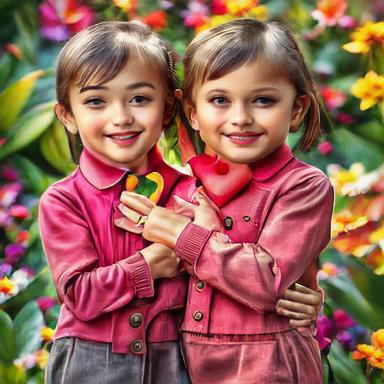}
        \caption{HPS}
        \label{fig:supp_wise2:hps}
    \end{subfigure}
    \hfill
    \begin{subfigure}[b]{0.22\textwidth}
        \centering
        \includegraphics[width=\textwidth]{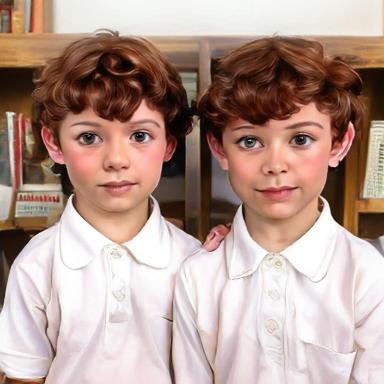}
        \caption{HPS + ArtifactR}
        \label{fig:supp_wise2:hps_arti}
    \end{subfigure}
    \hfill
    \begin{subfigure}[b]{0.22\textwidth}
        \centering
        \includegraphics[width=\textwidth]{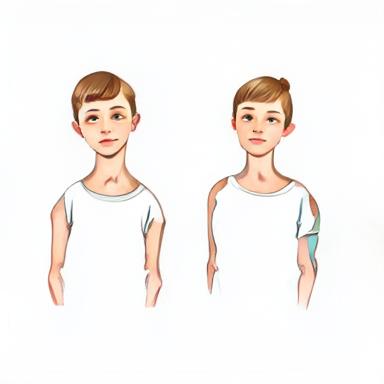}
        \caption{GDino}
        \label{fig:supp_wise2:gdino}
    \end{subfigure}
    \hfill
    \begin{subfigure}[b]{0.22\textwidth}
        \centering
        \includegraphics[width=\textwidth]{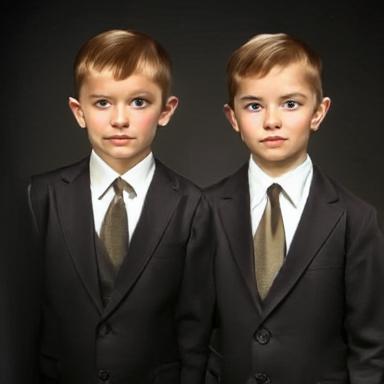}
        \caption{GDino + ArtifactR}
        \label{fig:supp_wise2:gdino_arti}
    \end{subfigure}


    \vspace{0.3cm} 

    \begin{subfigure}[b]{0.22\textwidth}
        \centering
        \includegraphics[width=\textwidth]{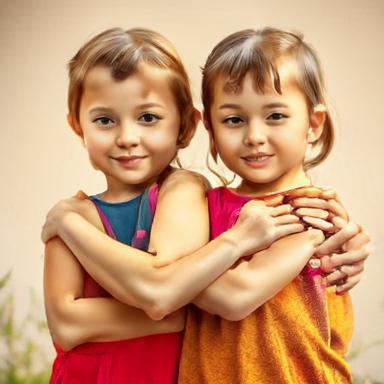}
        \caption{HPS + GDino}
        \label{fig:supp_wise2:hps_gdino}
    \end{subfigure}
    \hfill
    \begin{subfigure}[b]{0.22\textwidth}
        \centering
        \includegraphics[width=\textwidth]{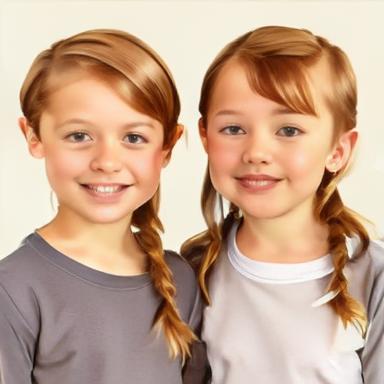}
        \caption{HPS + GDino + ArtifactR}
        \label{fig:supp_wise2:hps_gdino_arti}
    \end{subfigure}
    \hfill
    \begin{subfigure}[b]{0.22\textwidth}
        \centering
        \includegraphics[width=\textwidth]{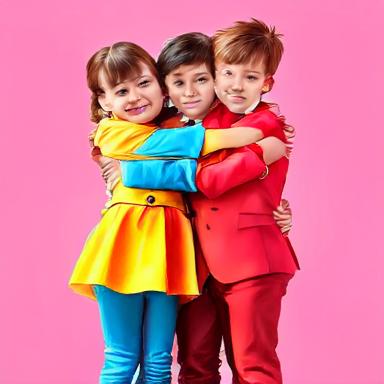}
        \caption{ORM}
        \label{fig:supp_wise2:orm}
    \end{subfigure}
    \hfill
    \begin{subfigure}[b]{0.22\textwidth}
        \centering
        \includegraphics[width=\textwidth]{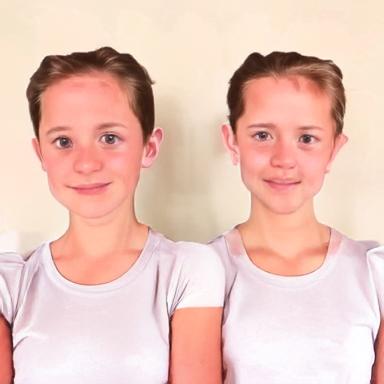}
        \caption{ORM + ArtifactR}
        \label{fig:supp_wise2:orm_arti}
    \end{subfigure}

    \caption{\small 
    Images generated with prompt ``\textit{Siblings of identical twins}'' in WISE~\cite{niu2025wise} benchmark under different training reward configurations trained on Janus-Pro-1B~\cite{chen2025janus}.
    }
    \label{fig:supp_wise2}

\end{figure*}

Figure~\ref{fig:wise_1b_res} presents the full WISE~\cite{niu2025wise} benchmark results across different prompt subcategories and evaluation metrics for Janus-Pro-1B~\cite{chen2025janus} trained with various reward configurations.
Across nearly all categories, incorporating our ArtifactReward leads to substantial improvements in image realism, consistently outperforming models trained with baseline rewards. In addition, ArtifactReward enhances image consistency in most subcategories, indicating that reducing structural artifacts not only improves visual plausibility but also strengthens text-image alignment across diverse prompt types.
More image illustrations can be found in Figure~\ref{fig:supp_wise} and Figure~\ref{fig:supp_wise2}.

\begin{figure*}[htbp]
    \centering

    \begin{subfigure}[b]{1.0\textwidth}
        \centering
        \includegraphics[width=\textwidth]{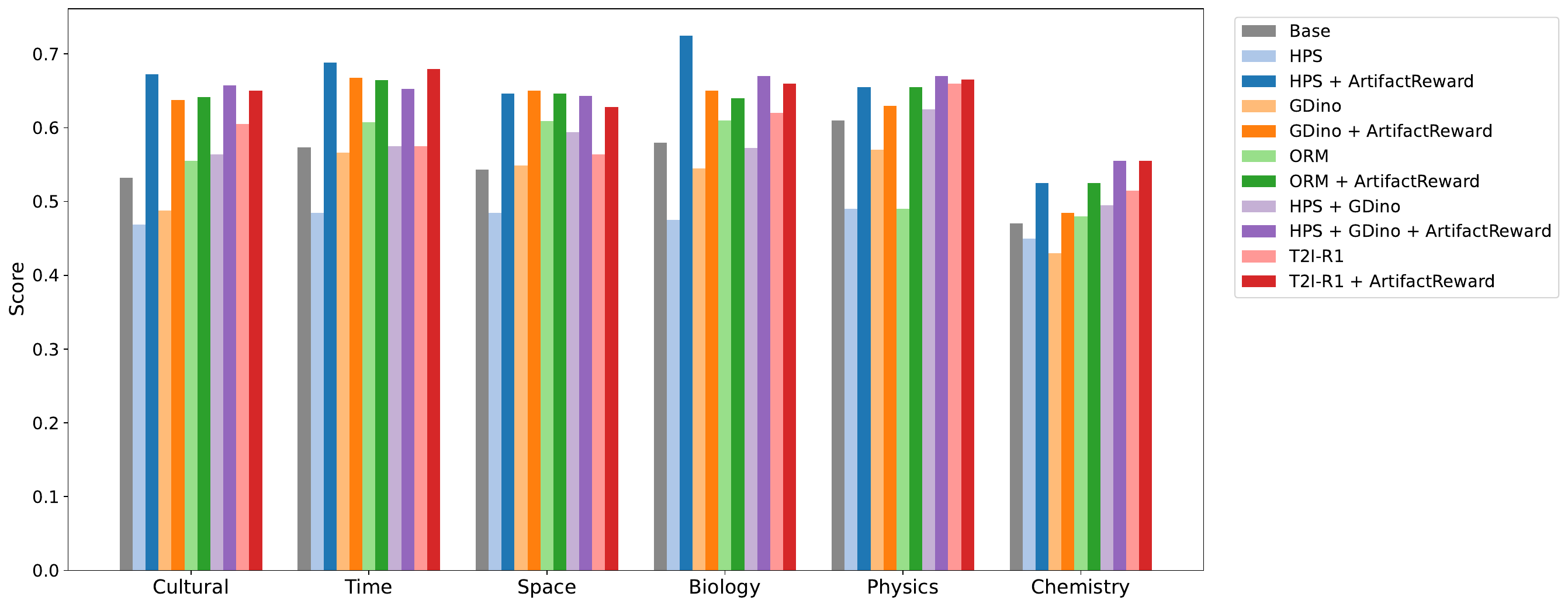}
        \caption{Realism}
        \label{fig:wise_7b_real}
    \end{subfigure}


    \vspace{0.3cm} 

    \begin{subfigure}[b]{1.0\textwidth}
        \centering
        \includegraphics[width=\textwidth]{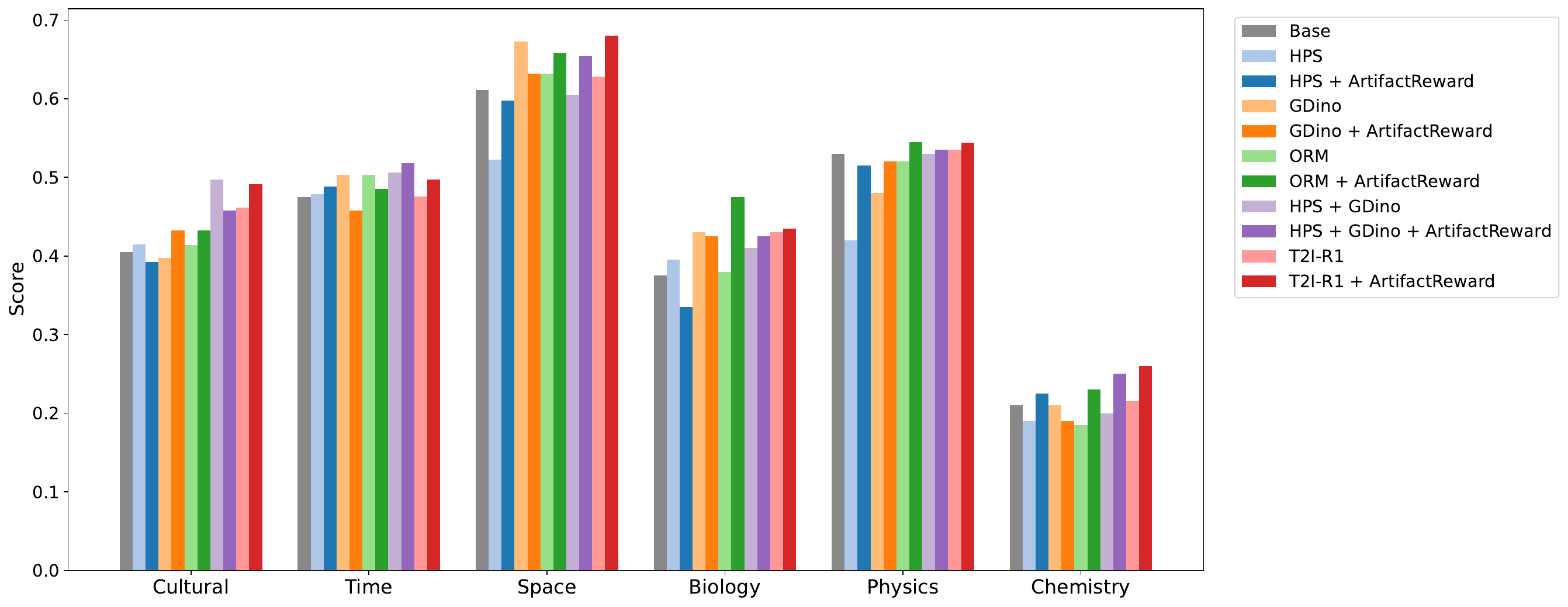}
        \caption{Consistency}
        \label{fig:wise_7b_consist}
    \end{subfigure}

    \vspace{0.3cm} 

    \begin{subfigure}[b]{1.0\textwidth}
        \centering
        \includegraphics[width=\textwidth]{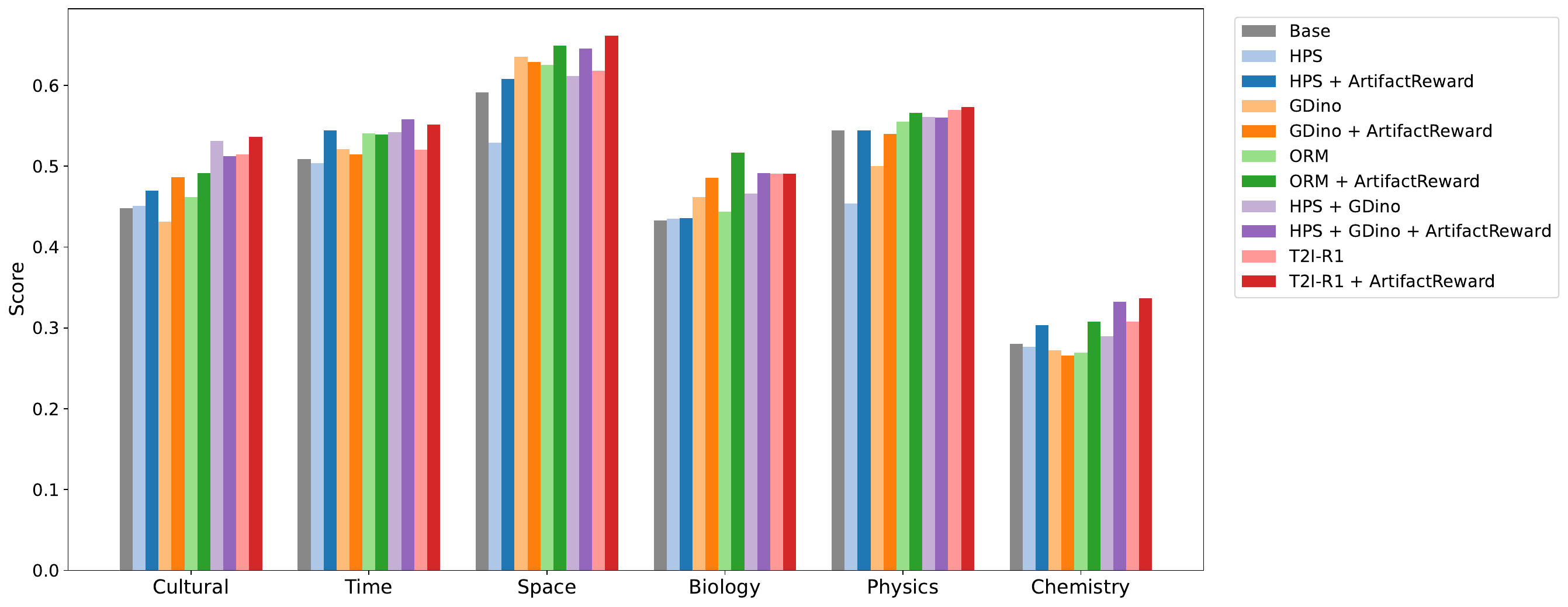}
        \caption{WiScore}
        \label{fig:wise_7b_wiscore}
    \end{subfigure}

    \caption{\small 
    Performance on WISE~\cite{niu2025wise} benchmark across different categories trained on Janus-Pro-7B~\cite{chen2025janus}
    }
    \label{fig:wise_7b_res}

\end{figure*}

We additionally evaluate our method on the larger Janus-Pro-7B~\cite{chen2025janus} model. As shown in Table~\ref{tab:supp_7b_wise} and Figure~\ref{fig:wise_7b_res}, the results mirror those observed with the 1B model, demonstrating that ArtifactReward generalizes effectively across model scales.

\subsection{Detailed Results of LLM4LLM benchmark}
\begin{table}
    \small
  \caption{\small Performance on LLM4LLM~\cite{wang2025lmm4lmm} benchmark trained on Janus-Pro-7B~\cite{chen2025janus}.}
  \label{tab:supp_7b_llm4llm}
  \centering
  \resizebox{1.0\linewidth}{!}
  {
  \begin{tabular}{l|cc|c}
    \toprule
    \textbf{Method}  &Perception &Correspondence &All \\
    \midrule
    Janus-Pro-7B~\cite{chen2025janus}  &0.4390 &0.5416 &0.9806 \\
    \midrule
    HPS~\cite{wu2023human}  &0.4360 &0.5205 &0.9564 \\
    HPS + ArtifactR  &0.4454 &0.5278 &0.9732  \\
    \midrule
    GDino~\cite{liu2024grounding}  &0.4417 &0.5570 &0.9987  \\
    GDino + ArtifactR  &0.4506 &0.5519 &1.0024 \\
    \midrule
    ORM~\cite{guo2025can}  &0.4499 &0.5561 &1.0059 \\
    ORM + ArtifactR &0.4576 &0.5613 &1.0188 \\
    \midrule
    HPS + GDino  &\textbf{0.4615} &0.5540 &1.0156 \\
    HPS + GDino + ArtifactR  &0.4558 &0.5594 &1.0152 \\
    \midrule
    T2I-R1~\cite{jiang2025t2i}  &0.4580 &0.5559 &1.0139 \\
    T2I-R1 + ArtifactR  &0.4608 &\textbf{0.5622} &\textbf{1.0230} \\
    \bottomrule
  \end{tabular}
  }
\end{table}

\begin{figure*}[htbp]
    \centering

    \begin{subfigure}[b]{1.0\textwidth}
        \centering
        \includegraphics[width=\textwidth]{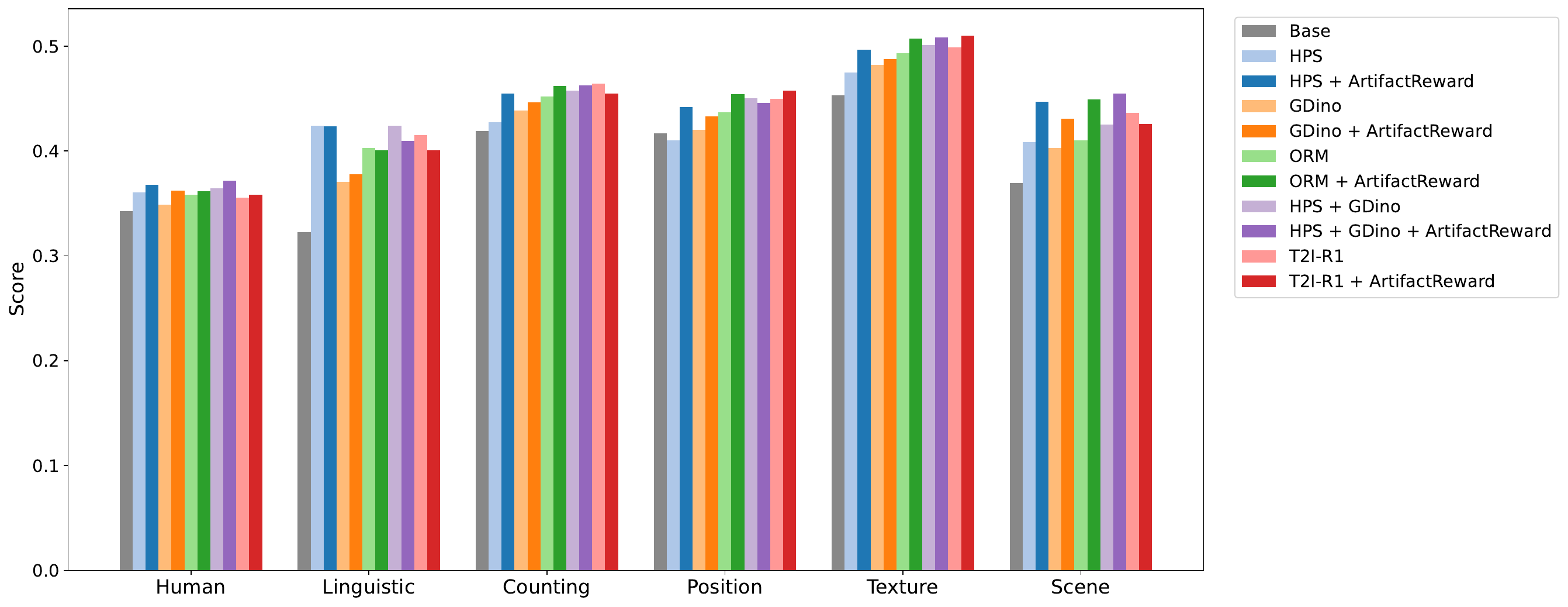}
        \caption{Perception}
        \label{fig:llm4llm_1b_percept}
    \end{subfigure}


    \vspace{0.3cm} 

    \begin{subfigure}[b]{1.0\textwidth}
        \centering
        \includegraphics[width=\textwidth]{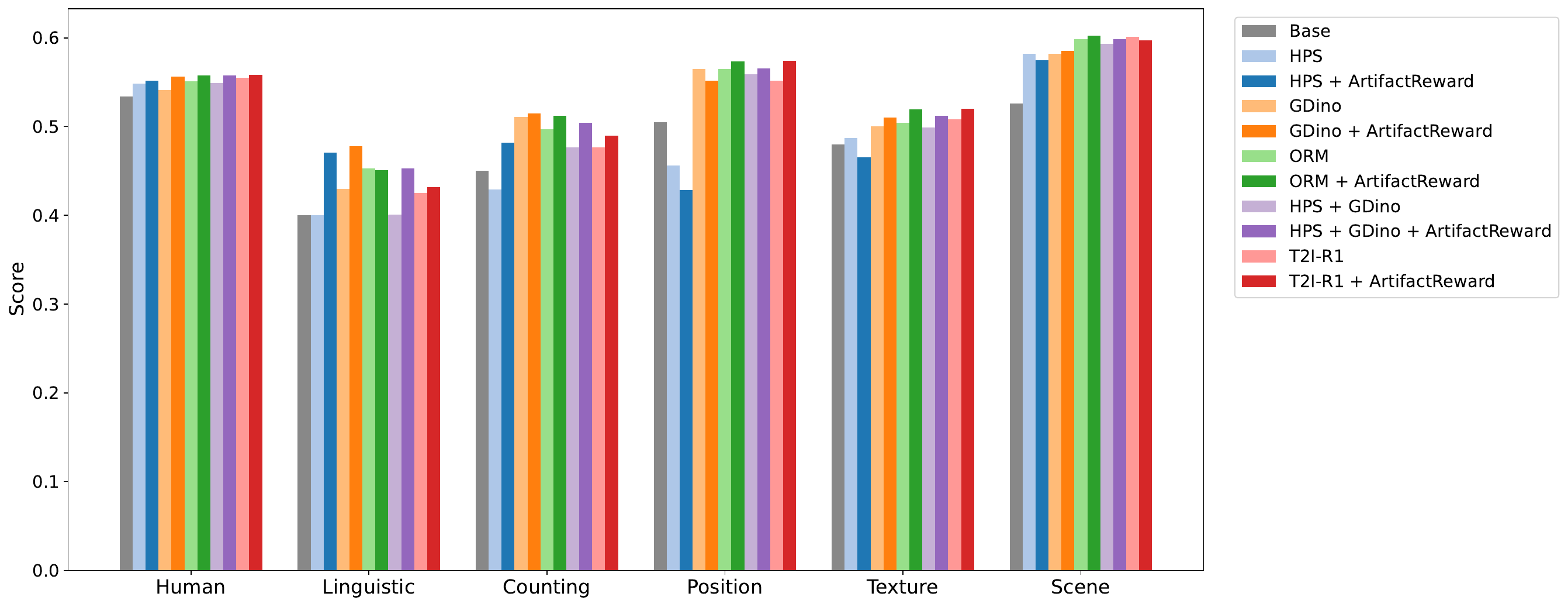}
        \caption{Correspondence}
        \label{fig:llm4llm_1b_corres}
    \end{subfigure}

    \caption{\small 
    Performance on LLM4LLM~\cite{wang2025lmm4lmm} benchmark across different categories trained on Janus-Pro-1B~\cite{chen2025janus}
    }
    \label{fig:llm4llm_1b_res}

\end{figure*}

\begin{figure*}[htbp]
    \centering
    
    \begin{subfigure}[b]{0.22\textwidth}
        \centering
        \includegraphics[width=\textwidth]{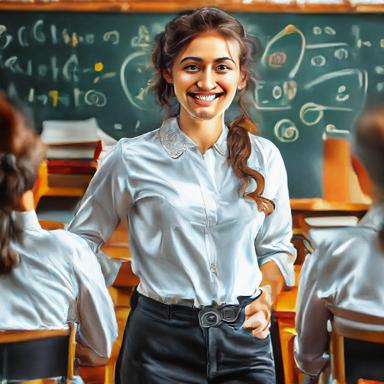}
        \caption{HPS}
        \label{fig:supp_llm4llm:hps}
    \end{subfigure}
    \hfill
    \begin{subfigure}[b]{0.22\textwidth}
        \centering
        \includegraphics[width=\textwidth]{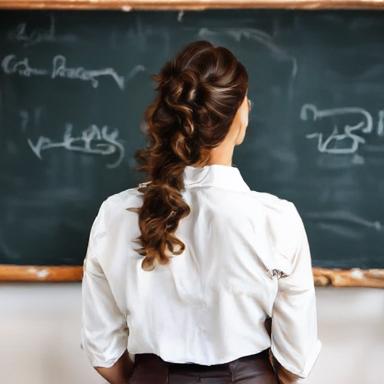}
        \caption{HPS + ArtifactR}
        \label{fig:supp_llm4llm:hps_arti}
    \end{subfigure}
    \hfill
    \begin{subfigure}[b]{0.22\textwidth}
        \centering
        \includegraphics[width=\textwidth]{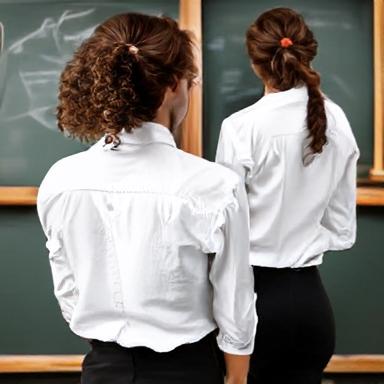}
        \caption{GDino}
        \label{fig:supp_llm4llm:gdino}
    \end{subfigure}
    \hfill
    \begin{subfigure}[b]{0.22\textwidth}
        \centering
        \includegraphics[width=\textwidth]{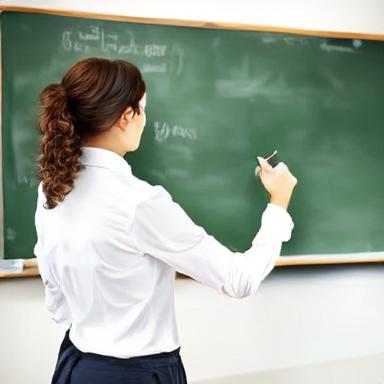}
        \caption{GDino + ArtifactR}
        \label{fig:supp_llm4llm:gdino_arti}
    \end{subfigure}


    \vspace{0.3cm} 

    \begin{subfigure}[b]{0.22\textwidth}
        \centering
        \includegraphics[width=\textwidth]{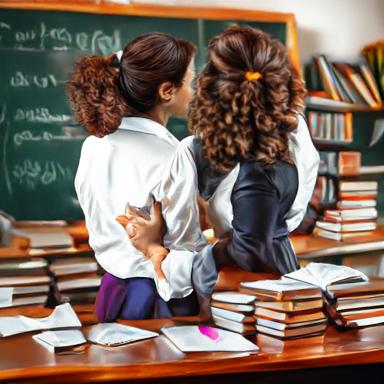}
        \caption{HPS + GDino}
        \label{fig:supp_llm4llm:hps_gdino}
    \end{subfigure}
    \hfill
    \begin{subfigure}[b]{0.22\textwidth}
        \centering
        \includegraphics[width=\textwidth]{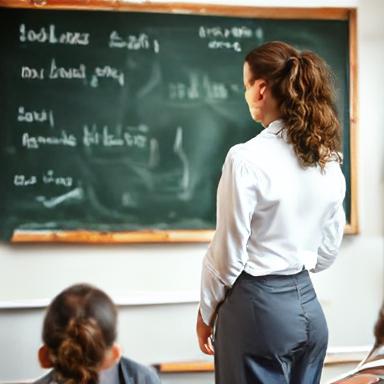}
        \caption{HPS + GDino + ArtifactR}
        \label{fig:supp_llm4llm:hps_gdino_arti}
    \end{subfigure}
    \hfill
    \begin{subfigure}[b]{0.22\textwidth}
        \centering
        \includegraphics[width=\textwidth]{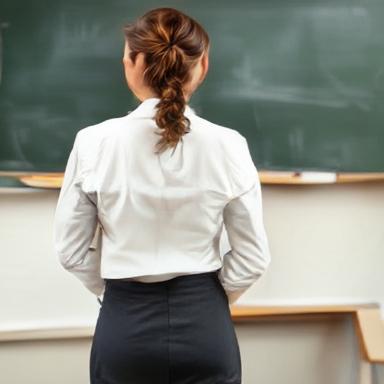}
        \caption{ORM}
        \label{fig:supp_llm4llm:orm}
    \end{subfigure}
    \hfill
    \begin{subfigure}[b]{0.22\textwidth}
        \centering
        \includegraphics[width=\textwidth]{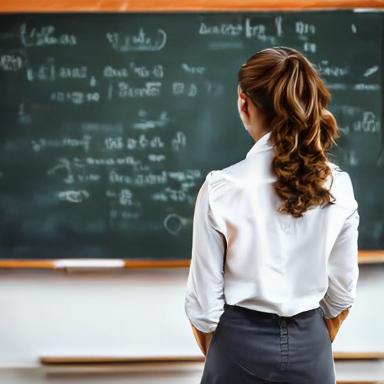}
        \caption{ORM + ArtifactR}
        \label{fig:supp_llm4llm:orm_arti}
    \end{subfigure}

    \caption{\small 
    Images generated with prompt ``\textit{A teacher in a white blouse stands at the blackboard, her curly brown hair tied back in a ponytail.}'' in LLM4LLM~\cite{wang2025lmm4lmm} benchmark under different training reward configurations trained on Janus-Pro-1B~\cite{chen2025janus}.
    }
    \label{fig:supp_llm4llm}

\end{figure*}

\begin{figure*}[htbp]
    \centering
    
    \begin{subfigure}[b]{0.22\textwidth}
        \centering
        \includegraphics[width=\textwidth]{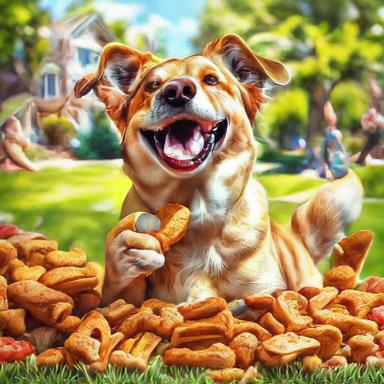}
        \caption{HPS}
        \label{fig:supp_llm4llm2:hps}
    \end{subfigure}
    \hfill
    \begin{subfigure}[b]{0.22\textwidth}
        \centering
        \includegraphics[width=\textwidth]{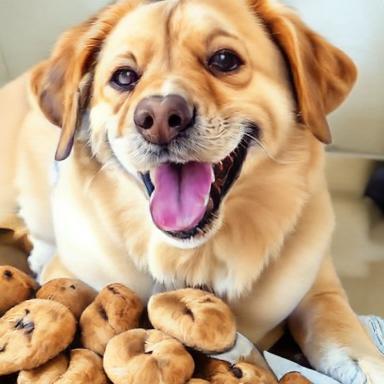}
        \caption{HPS + ArtifactR}
        \label{fig:supp_llm4llm2:hps_arti}
    \end{subfigure}
    \hfill
    \begin{subfigure}[b]{0.22\textwidth}
        \centering
        \includegraphics[width=\textwidth]{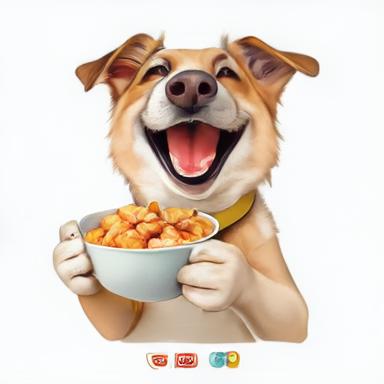}
        \caption{GDino}
        \label{fig:supp_llm4llm2:gdino}
    \end{subfigure}
    \hfill
    \begin{subfigure}[b]{0.22\textwidth}
        \centering
        \includegraphics[width=\textwidth]{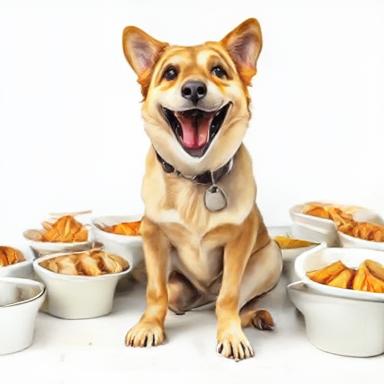}
        \caption{GDino + ArtifactR}
        \label{fig:supp_llm4llm2:gdino_arti}
    \end{subfigure}


    \vspace{0.3cm} 

    \begin{subfigure}[b]{0.22\textwidth}
        \centering
        \includegraphics[width=\textwidth]{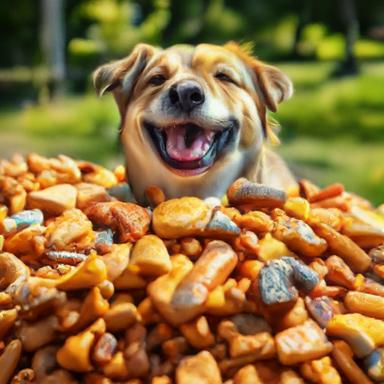}
        \caption{HPS + GDino}
        \label{fig:supp_llm4llm2:hps_gdino}
    \end{subfigure}
    \hfill
    \begin{subfigure}[b]{0.22\textwidth}
        \centering
        \includegraphics[width=\textwidth]{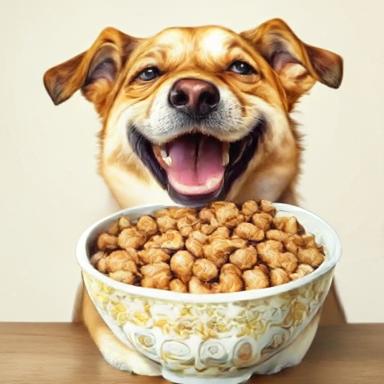}
        \caption{HPS + GDino + ArtifactR}
        \label{fig:supp_llm4llm2:hps_gdino_arti}
    \end{subfigure}
    \hfill
    \begin{subfigure}[b]{0.22\textwidth}
        \centering
        \includegraphics[width=\textwidth]{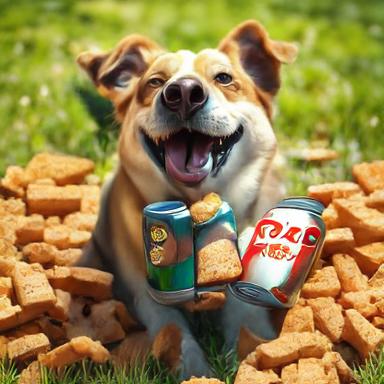}
        \caption{ORM}
        \label{fig:supp_llm4llm2:orm}
    \end{subfigure}
    \hfill
    \begin{subfigure}[b]{0.22\textwidth}
        \centering
        \includegraphics[width=\textwidth]{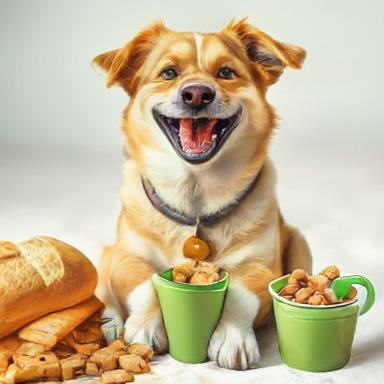}
        \caption{ORM + ArtifactR}
        \label{fig:supp_llm4llm2:orm_arti}
    \end{subfigure}

    \caption{\small 
    Images generated with prompt ``\textit{a dog is smiling with happy emotion. He find a lot of delicious food.}'' in LLM4LLM~\cite{wang2025lmm4lmm} benchmark under different training reward configurations trained on Janus-Pro-1B~\cite{chen2025janus}.
    }
    \label{fig:supp_llm4llm2}

\end{figure*}

Figure~\ref{fig:llm4llm_1b_res} presents the LLM4LLM~\cite{wang2025lmm4lmm} benchmark results across different prompt subcategories and evaluation metrics for Janus-Pro-1B~\cite{chen2025janus} trained with various reward configurations.
More image illustrations can be found in Figure~\ref{fig:supp_llm4llm} and Figure~\ref{fig:supp_llm4llm2}.

\begin{figure*}[htbp]
    \centering

    \begin{subfigure}[b]{1.0\textwidth}
        \centering
        \includegraphics[width=\textwidth]{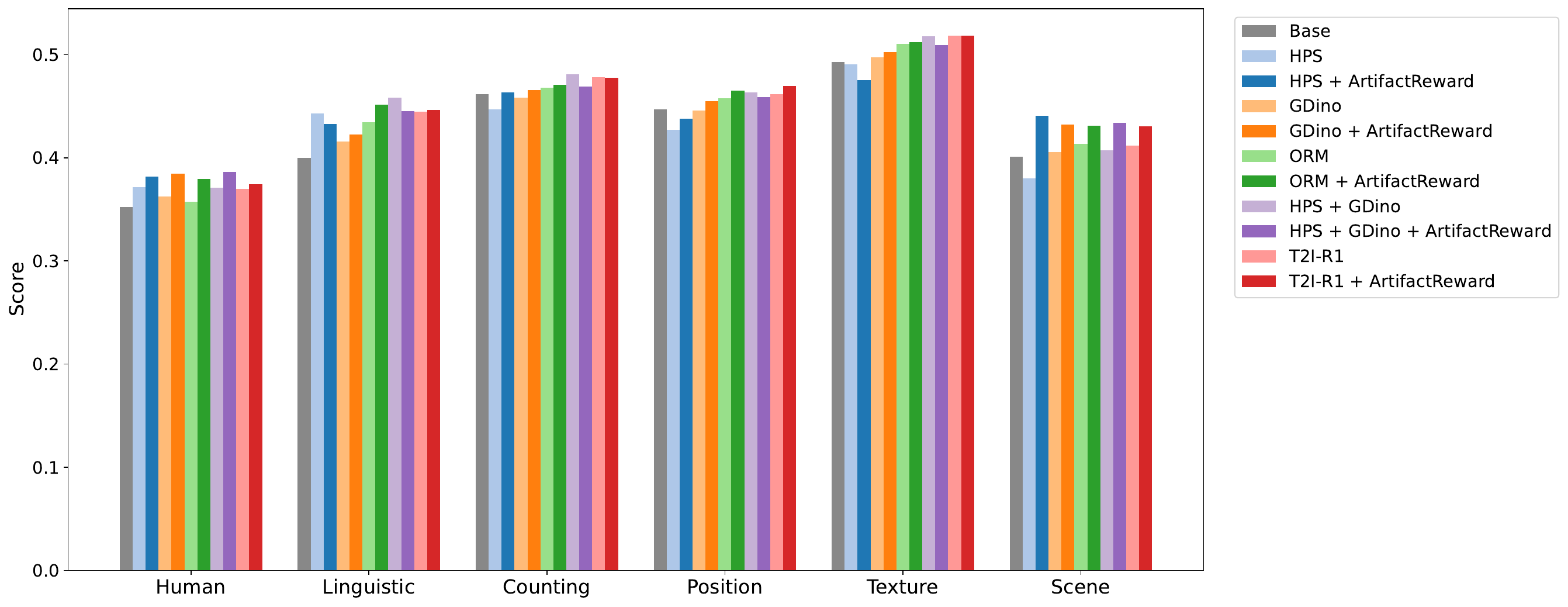}
        \caption{Perception}
        \label{fig:llm4llm_7b_percept}
    \end{subfigure}


    \vspace{0.3cm} 

    \begin{subfigure}[b]{1.0\textwidth}
        \centering
        \includegraphics[width=\textwidth]{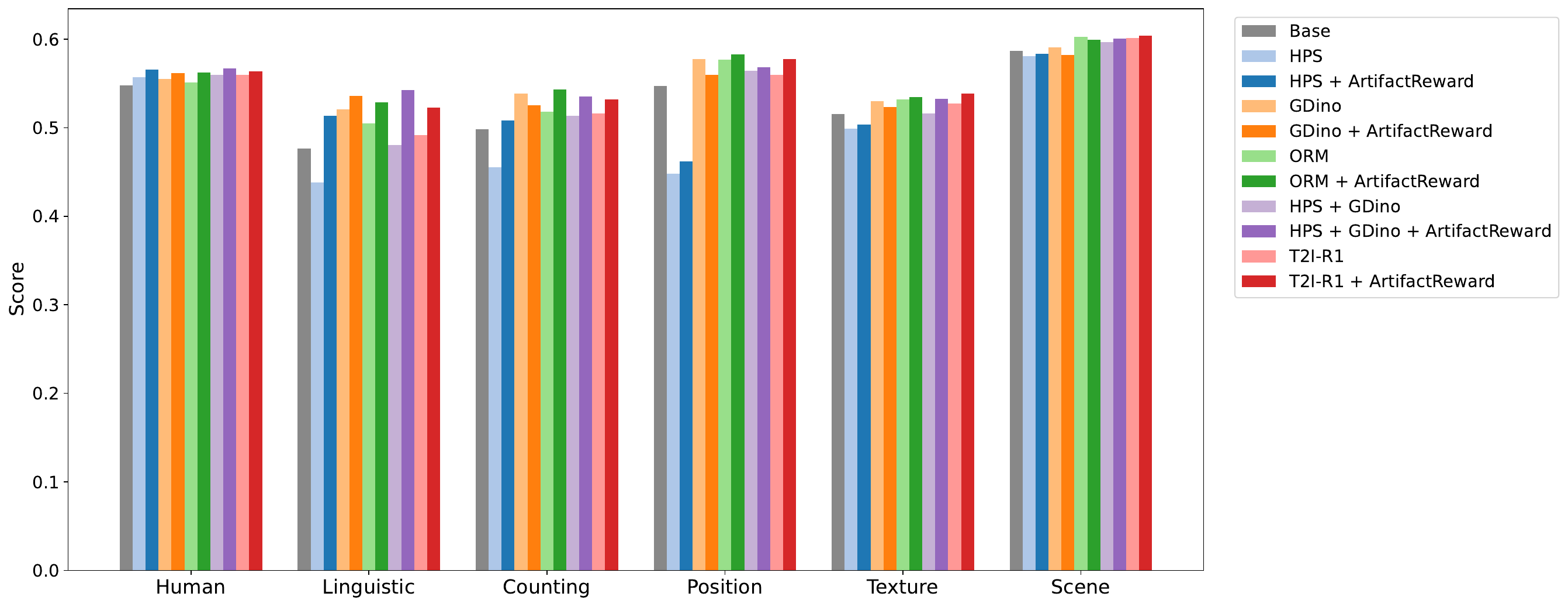}
        \caption{Correspondence}
        \label{fig:llm4llm_7b_corres}
    \end{subfigure}

    \caption{\small 
    Performance on LLM4LLM~\cite{wang2025lmm4lmm} benchmark across different categories trained on Janus-Pro-7B~\cite{chen2025janus}
    }
    \label{fig:llm4llm_7b_res}

\end{figure*}

Additional results trained on Janus-Pro-7B~\cite{chen2025janus} are shown in Table~\ref{tab:supp_7b_llm4llm} and Figure~\ref{fig:llm4llm_7b_res}.

\subsection{Detailed Results of EVALALIGN benchmark}

\begin{table}[t]
\centering
\caption{Performance on EvalAlign~\cite{tan2024evalalign} benchmark trained on Janus-Pro~\cite{chen2025janus}.}
\label{tab:evalalign_1b}
\resizebox{0.65\linewidth}{!}{
\begin{tabular}{l|cc}
\toprule
\multirow{2}{*}{\textbf{Method}} & \multicolumn{2}{c}{\textbf{Faithfulness}} \\

 & 1B & 7B \\
\midrule
Janus-Pro~\cite{chen2025janus}         & 0.7642 & 0.8694 \\
\midrule
HPS~\cite{wu2023human}                 & 0.7569 & 0.8538 \\
HPS + ArtifactR           & \textbf{0.9302} & \textbf{1.0676} \\
\midrule
GDino~\cite{liu2024grounding}               & 0.7832 & 0.8956 \\
GDino + ArtifactR         & \textbf{0.8952} & \textbf{0.9837} \\
\midrule
ORM~\cite{guo2025can}                 & 0.7560 & 0.8607 \\
ORM + ArtifactR           & \textbf{0.9140} & \textbf{0.9245} \\
\midrule
HPS + GDino              & 0.7840 & 0.8956 \\
HPS + GDino + ArtifactR   & \textbf{0.8756} & \textbf{0.9837} \\
\midrule
T2I-R1~\cite{jiang2025t2i}              & 0.7692 & 0.9059 \\
T2I-R1 + ArtifactR        & \textbf{0.9015} & \textbf{0.9638} \\
\bottomrule
\end{tabular}
}
\end{table}

We additionally evaluate our method on \textbf{EVALALIGN} \cite{tan2024evalalign}, a fine-grained, MLLM-based evaluation benchmark for text-to-image models. It uses image faithfulness, how accurately the visual content matches reality or semantics, as one of their image quality measurements.
To highlight the impact of our method on reducing such artifacts, we focus on this faithfulness metric.

\begin{figure*}[htbp]
    \centering

    \begin{subfigure}[b]{1.0\textwidth}
        \centering
        \includegraphics[width=\textwidth]{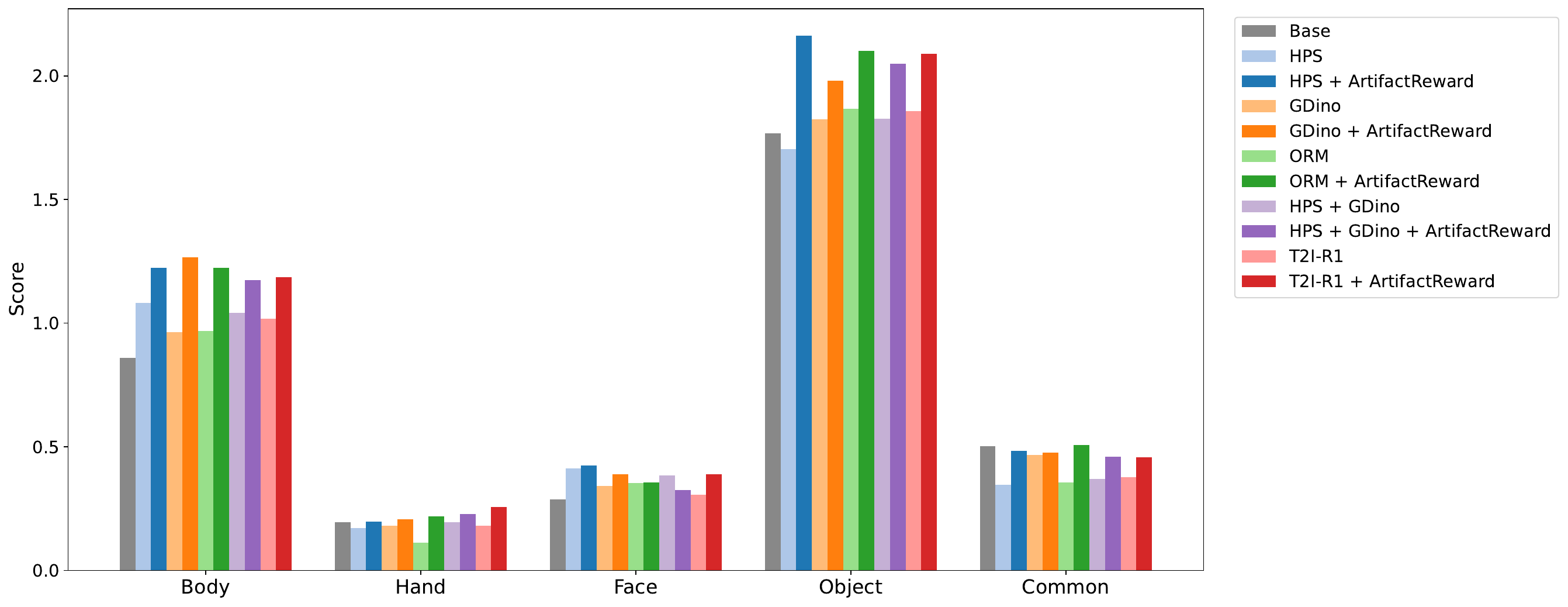}
        \label{fig:evalalign_1b_faithfulness}
    \end{subfigure}

    \caption{\small 
    Performance on EvalAlign \cite{tan2024evalalign} benchmark across different categories trained on Janus-Pro-1B~\cite{chen2025janus}.
    }
    \label{fig:evalalign_1b_res}

\end{figure*}

\begin{figure*}[htbp]
    \centering

    \begin{subfigure}[b]{1.0\textwidth}
        \centering
        \includegraphics[width=\textwidth]{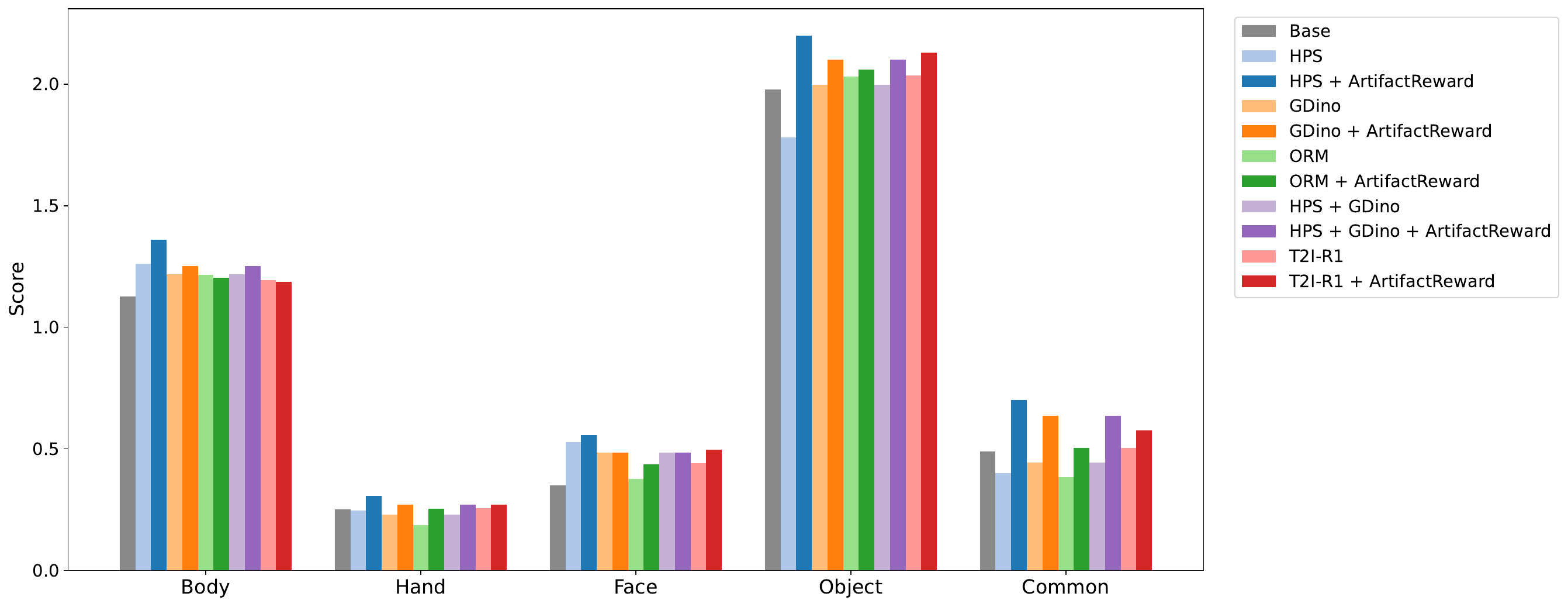}
        \label{fig:evalalign_7b_faithfulness}
    \end{subfigure}

    \caption{\small 
    Performance on EvalAlign \cite{tan2024evalalign} benchmark across different categories trained on Janus-Pro-7B~\cite{chen2025janus}.
    }
    \label{fig:evalalign_7b_res}

\end{figure*}

As shown in Table \ref{tab:evalalign_1b}, Figure \ref{fig:evalalign_1b_res} and \ref{fig:evalalign_7b_res}, incorporating our ArtifactReward consistently leads to substantial improvements, achieving higher scores across nearly all subcategories. These results further confirm that ArtifactReward can enhance semantic plausibility and structural correctness beyond what existing reward models capture.




\end{document}